\documentclass{article}

\usepackage{microtype}
\usepackage{graphicx}
\usepackage{booktabs}
\usepackage{hyperref}
\RequirePackage{snapshot}

\usepackage[accepted]{mlsys2021}
\mlsystitlerunning{Bit Error Robustness for Energy-Efficient DNN Accelerators}

\usepackage{amsmath}
\usepackage{xspace}
\usepackage{subcaption}
\usepackage{bbm}
\usepackage{nicefrac}
\usepackage{tabularx}
\usepackage{multirow}
\usepackage{tikz}
\usepackage{pgfplots}
\usepackage{amsmath}
\usepackage{amsthm}
\usepackage{amsfonts}
\usepackage{xspace}
\usepackage{mathtools}
\usepackage{algorithm}
\usepackage{algorithmicx}
\usepackage{algpseudocode}
\usepackage{listings}
\usepackage{placeins}
\DeclareFixedFont{\ttb}{T1}{txtt}{bx}{n}{12} 
\DeclareFixedFont{\ttm}{T1}{txtt}{m}{n}{12}  
\definecolor{deepblue}{rgb}{0,0,0.5}
\definecolor{deepred}{rgb}{0.6,0,0}
\definecolor{deepgreen}{rgb}{0,0.5,0}
\algrenewcommand\algorithmicindent{0.75em}%
\algtext*{EndWhile}
\algtext*{EndIf}
\algtext*{EndFor}
\algtext*{EndProcedure}
\usepackage{caption}
\usepackage{wrapfig}
\usepackage[breakable, theorems, skins]{tcolorbox}
\usepackage{xspace}
\usepackage{array}

\newcolumntype{L}[1]{>{\raggedright\let\newline\\\arraybackslash\hspace{0pt}}m{#1}}
\newcolumntype{C}[1]{>{\centering\let\newline\\\arraybackslash\hspace{0pt}}m{#1}}
\newcolumntype{R}[1]{>{\raggedleft\let\newline\\\arraybackslash\hspace{0pt}}m{#1}}

\def\Vhrulefill{\leavevmode\hrule height 0.7ex depth \dimexpr0.4pt-0.7ex with 1cm\kern0pt}

\usepackage{xcolor}
\definecolor{colorbrewer0}{RGB}{45,45,45}
\definecolor{colorbrewer1}{RGB}{228,26,28}
\definecolor{colorbrewer2}{RGB}{55,126,184}
\definecolor{colorbrewer3}{RGB}{77,175,74}
\definecolor{colorbrewer4}{RGB}{152,78,163}
\definecolor{colorbrewer5}{RGB}{255,127,0}
\definecolor{colorbrewer6}{RGB}{255,255,51}
\definecolor{colorbrewer7}{RGB}{166,86,40}
\definecolor{colorbrewer8}{RGB}{247,129,191}
\definecolor{colorbrewer9}{RGB}{153,153,153}
\definecolor{colorbrewer10}{RGB}{24,167,181}

\newtheorem{proposition}{Proposition}

\definecolor{gray}{RGB}{65, 65, 65}
\definecolor{lightgray}{RGB}{135, 135, 135}
\definecolor{darkred}{RGB}{255, 0, 0}
\definecolor{darkgreen}{RGB}{25, 100, 25}
\definecolor{darkmagenta}{RGB}{255, 0, 255}
\definecolor{darkblue}{RGB}{0, 0, 255}

\definecolor{colorbrewer0}{RGB}{45,45,45}
\definecolor{colorbrewer1}{RGB}{228,26,28}
\definecolor{colorbrewer2}{RGB}{55,126,184}
\definecolor{colorbrewer3}{RGB}{77,175,74}
\definecolor{colorbrewer4}{RGB}{152,78,163}
\definecolor{colorbrewer5}{RGB}{255,127,0}
\definecolor{colorbrewer6}{RGB}{255,255,51}
\definecolor{colorbrewer7}{RGB}{166,86,40}
\definecolor{colorbrewer8}{RGB}{247,129,191}
\definecolor{colorbrewer9}{RGB}{153,153,153}
\definecolor{colorbrewer10}{RGB}{24,167,181}

\makeatletter
\DeclareRobustCommand\onedot{\futurelet\@let@token\@onedot}
\def\@onedot{\ifx\@let@token.\else.\null\fi\xspace}

\def\eg{\emph{e.g}\onedot} 
\def\ie{\emph{i.e}\onedot} 
\def\cf{\emph{c.f}\onedot} 
\def\etc{\emph{etc}\onedot} 
\def\wrt{w.r.t\onedot} 

\makeatother

\newcommand{\figref}[1]{\Fig~\ref{#1}}
\newcommand{\secref}[1]{\Sec~\ref{#1}}
\newcommand{\appref}[1]{\App~\ref{#1}}

\renewcommand{\algref}[1]{\Alg~\ref{#1}}
\newcommand{\eqnref}[1]{\Eq~\eqref{#1}}
\newcommand{\tabref}[1]{\Tab~\ref{#1}}

\makeatletter
\DeclareRobustCommand\onedot{\futurelet\@let@token\@onedot}
\def\@onedot{\ifx\@let@token.\else.\null\fi\xspace}
\def\eg{e.g\onedot} 
\def\ie{i.e\onedot} 
 
\def\cf{cf\onedot} 
\def\etc{etc\onedot}

\def\wrt{wrt\onedot}

\def\Fig{Fig\onedot} \def\Eq{Eq\onedot}
\def\Sec{Sec\onedot} \def\Alg{Alg\onedot}
\def\Tab{Tab\onedot} 
\def\App{App\onedot} 
\makeatother

\DeclareRobustCommand{\RTE}{%
    \ifmmode
    \text{RErr}
    \else
    RErr\xspace
    \fi
}
\DeclareRobustCommand{\TE}{%
    \ifmmode
    \text{Err}
    \else
    Err\xspace
    \fi
}
\DeclareRobustCommand{\wmax}{%
    \ifmmode
    w_{\text{max}}
    \else
    $w_{\text{max}}$\xspace
    \fi
}
\DeclareRobustCommand{\qmax}{%
    \ifmmode
    q_{\text{max}}
    \else
    $q_{\text{max}}$\xspace
    \fi
}
\DeclareRobustCommand{\qmin}{%
    \ifmmode
    q_{\text{min}}
    \else
    $q_{\text{min}}$\xspace
    \fi
}
\DeclareRobustCommand{\Vmin}{%
    \ifmmode
    V_{\text{min}}
    \else
    $V_{\text{min}}$\xspace
    \fi
}
\DeclareRobustCommand{\Vt}{%
    \ifmmode
    V_{\text{th}}
    \else
    $V_{\text{th}}$\xspace
    \fi
}
\DeclareRobustCommand{\biterror}{%
    \ifmmode
    \text{BErr}
    \else
    BErr\xspace
    \fi
}
\DeclareRobustCommand{\pfault}{%
    \ifmmode
    p_{\text{flt}}
    \else
    $p_{\text{{flt}}}$\xspace
    \fi
}
\DeclareRobustCommand{\perror}{%
    \ifmmode
    p_{\text{err}}
    \else
    $p_{\text{err}}$\xspace
    \fi
}

\def\Pr{\mathrm{P}}

\def\Exp{\mathbb{E}}

\newcommand{\Id}{\mathbbm{1}}

\def\min{\mathop{\rm min}\nolimits}
\def\max{\mathop{\rm max}\nolimits}

\def\maxop{\mathop{\rm max}\limits}

\newcommand{\MNIST}{MNIST\xspace}

\newcommand{\Cifar}{CIFAR\xspace}
\newcommand{\CifarT}{CIFAR10\xspace}
\newcommand{\CifarH}{CIFAR100\xspace}

\newcommand{\Normal}{\textsc{Normal}\xspace}
\newcommand{\Quant}{\textsc{RQuant}\xspace}
\newcommand{\Clipping}[1][]{\textsc{Clipping}\textsubscript{#1}\xspace}
\newcommand{\Pattern}[1][]{\textsc{PattBET}\textsubscript{#1}\xspace}
\newcommand{\Random}[1][]{\textsc{RandBET}\textsubscript{#1}\xspace}

\newcommand{\red}[1]{\noindent{\color{darkred}{#1}}}

\newcommand{\magenta}[1]{\noindent{\color{darkmagenta}{#1}}}
\newcommand{\blue}[1]{\noindent{\color{darkblue}{#1}}}

\newcommand{\david}[1]{\noindent{\color{red}{David: #1}}}

\newcommand{\nandhini}[1]{\noindent{\color{violet}{Nandhini: #1}}}
\newcommand{\revision}[1]{\noindent{#1}}

\renewcommand{\david}[1]{}
\renewcommand{\nandhini}[1]{}
\begin{document} 

\twocolumn[
\mlsystitle{Bit Error Robustness for Energy-Efficient DNN Accelerators}
\mlsyssetsymbol{equal}{*} 

\begin{mlsysauthorlist}
\mlsysauthor{David Stutz}{mpii}
\mlsysauthor{Nandhini Chandramoorthy}{ibm}
\mlsysauthor{Matthias Hein}{tue}
\mlsysauthor{Bernt Schiele}{mpii}
\end{mlsysauthorlist}

\mlsysaffiliation{mpii}{Max Planck Institute for Informatics}
\mlsysaffiliation{ibm}{IBM T. J. Watson Research Center}
\mlsysaffiliation{tue}{University of T\"{u}bingen}

\mlsyscorrespondingauthor{David Stutz}{david.stutz@mpi-inf.mpg.de}
\mlsyskeywords{Robustness, Accelerators, Bit Errors, Quantization, Weight Clipping}
 
\vskip 0.3in

\begin{abstract}
	Deep neural network (DNN) accelerators
	received considerable attention in past years due to saved energy compared to mainstream hardware. Low-voltage operation of DNN accelerators allows to further reduce energy consumption significantly, however, causes bit-level failures in the memory storing the quantized DNN weights.
	In this paper, we show that a combination of \textbf{robust fixed-point quantization}, 
	\textbf{weight clipping}, and \textbf{random bit error training (\Random)}
    \textbf{improves robustness against random bit errors in (quantized) DNN weights} significantly.
    This leads to high energy savings from \emph{both} low-voltage operation \emph{as well as} low-precision quantization. Our approach generalizes across operating voltages and accelerators, as demonstrated on bit errors from profiled SRAM arrays. We also discuss why weight clipping alone is already a quite effective way to achieve robustness against bit errors. Moreover, we specifically discuss the involved trade-offs regarding accuracy, robustness and precision: Without losing more than $1\%$ in accuracy compared to a normally trained $8$-bit DNN, we can reduce energy consumption on \CifarT by $20\%$. Higher energy savings of, \eg, $30\%$, are possible at the cost of $2.5\%$ accuracy, even for $4$-bit DNNs.
\end{abstract}
]

\printAffiliationsAndNotice{}

\section{Introduction}

Energy-efficiency is an important goal to lower carbon-dioxide emissions of deep neural network (DNN) driven applications and is a critical prerequisite to enable applications in edge computing.
\emph{DNN accelerators}, \ie, specialized hardware for inference, are used to reduce and limit energy consumption alongside cost and space compared to mainstream hardware, \eg, GPUs.
These accelerators generally feature on-chip SRAM used as scratchpads, \eg, to store DNN weights. Data access/movement
constitutes a dominant component of accelerator energy consumption \cite{SzeIEEE2017}.
Reduced precision \cite{LinICML2016} is a widely used measure to reduce energy consumption at the cost of \emph{approximate computing} \cite{SampsonPLDI2011}.
Recently, DNN accelerators \cite{ReagenISCA2016, KimDATE2018,ChandramoorthyHPCA2019} further lower memory supply voltage to increase energy efficiency since dynamic power varies quadratically with voltage. However, aggressive SRAM supply voltage scaling, causes bit-level failures in SRAM on account of process variation \cite{GanapathyDAC2017,GuoJSSC2009} with direct impact on the stored DNN weights. The rate $p$ of these errors increases exponentially with lowered voltage
and causes devastating drops in DNN accuracy such that memory reliability becomes the bottleneck in realizing low power DNN accelerators. In this paper, we aim to enable very low-voltage operation of DNN accelerators by developing DNNs robust to such bit errors in their weights, allowing DNN inference on \emph{``approximate hardware''} \cite{KoppulaMICRO2019,SampsonPLDI2011}. This is also desirable to improve security against adversarial
manipulation of voltage settings \cite{TangUSENIX2017}. 
In general, robustness to bit errors in DNNs is a desirable goal in order to maintain safe operation and should become a standard performance metric in low power DNN design.

\begin{figure}[t]
    \centering
    \vspace*{-0.1cm}
    \hspace*{-0.4cm}
    \begin{subfigure}{0.475\textwidth}
    	\includegraphics[height=5.25cm]{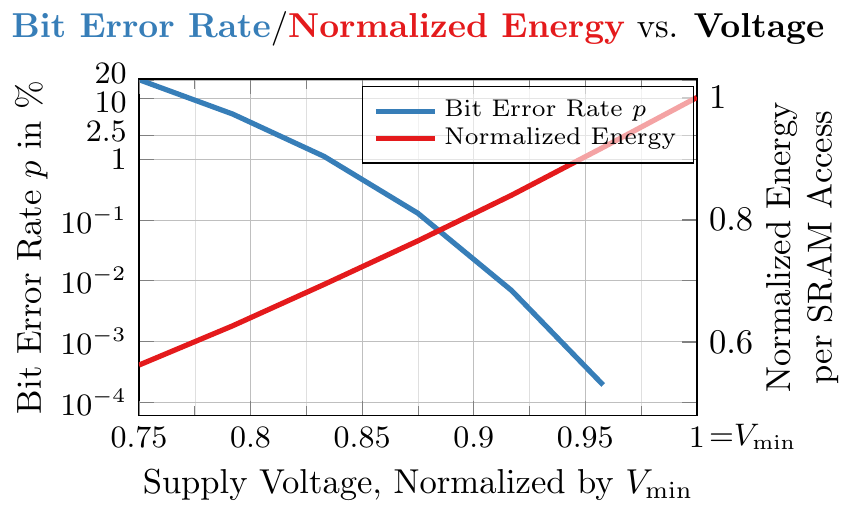}
    \end{subfigure}
    \vspace*{-8px}
    \caption{
    \textbf{Energy and Low-Voltage Operation.} Average bit error rate $p$ ({\color{colorbrewer2}blue}, left y-axis) from $32$ 14nm SRAM arrays of size $512{\times}64$ from \cite{ChandramoorthyHPCA2019} and energy ({\color{colorbrewer1}red}, right y-axis) vs. voltage (x-axis). Voltage is normalized by \Vmin, the minimal measured voltage for error-free operation, as well as the energy per SRAM access at \Vmin. SRAM accesses have significant impact on the DNN accelerator's energy \cite{ChenISCA2016}. Reducing voltage leads to exponentially increasing bit error rates.
    }
    \label{fig:introduction}
    \vspace*{-0.2cm}
\end{figure}

\begin{figure}[t]
	\centering
	\vspace*{-0.1cm}
	\hspace*{-0.25cm}
	\begin{subfigure}{0.425\textwidth}
         \includegraphics[height=5.25cm]{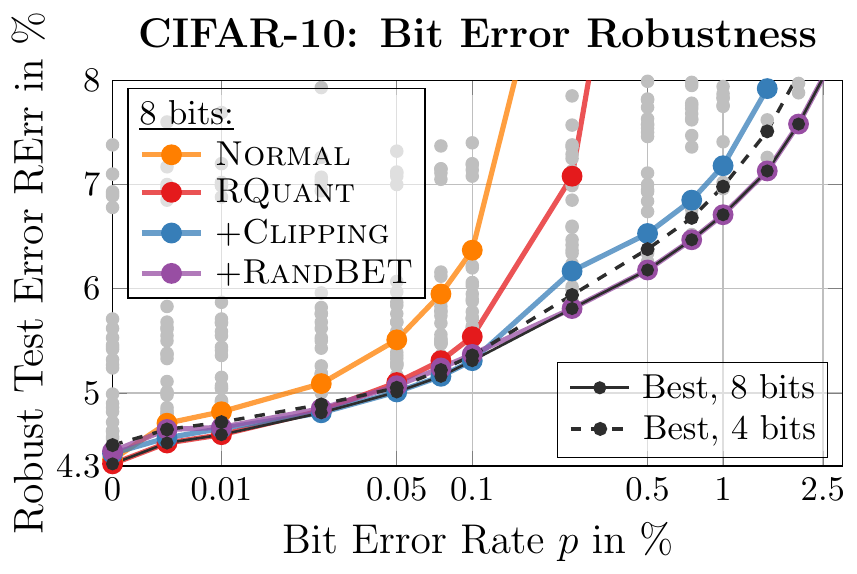}
    \end{subfigure}
    \vspace*{-8px}
    \caption{
    \textbf{Robustness to Random Bit Errors.} Robust test error (test error \emph{after} injecting bit errors, \RTE, lower is better $\downarrow$, y-axis) plotted against bit error rate $p$ (x-axis). Robustness to higher bit error rates allows more energy efficient operation, \cf \figref{fig:introduction}. For $8$ bit, through robust quantization (\Quant, {\color{colorbrewer1}red}), additionally weight clipping (\Clipping, {\color{colorbrewer2}blue}) and finally adding random bit error training (\Random, {\color{colorbrewer4}violet}) robustness improves significantly. 
    The Pareto optimal frontier is shown for $8$ bit (black solid) and $4$ bit (dashed) quantization.
    }
    \label{fig:contributions}
    \vspace*{-0.2cm}
\end{figure}

\figref{fig:introduction} shows the average bit error rates of SRAM arrays as supply voltage is aggressively scaled below \Vmin, \ie, the measured lowest voltage at which there are no bit errors. Voltage (x-axis) and energy ({\color{colorbrewer1}red}, right y-axis) are normalized \wrt \Vmin and the energy per access at \Vmin, respectively. DNNs robust to a bit error rate ({\color{colorbrewer2}blue}, left y-axis) of, \eg, $p = 1\%$ allow to reduce SRAM energy by roughly $30\%$.
To improve DNN robustness to bit errors, we first consider the impact of fixed-point quantization on robustness. While prior work \cite{MurthyARXIV2019,MerollaARXIV2016,SungARXIV2015} studies robustness \emph{to} quantization, the impact of random bit errors \emph{in} quantized weights has not been considered so far. We find that the choice of quantization scheme has tremendous impact on robustness, even though accuracy is not affected. In particular, we identify a particularly \textbf{robust quantization scheme}, \Quant in \figref{fig:contributions} ({\color{colorbrewer1}red}). 
Additionally, independent of the quantization scheme, we propose aggressive \textbf{weight clipping} during training. This acts as an explicit regularizer leading to spread out weight distributions, improving
robustness significantly,
\Clipping in \figref{fig:contributions} ({\color{colorbrewer2}blue}). This is in contrast to, \eg, \cite{ZhuangCVPR2018,SungARXIV2015} ignoring weight outliers to reduce quantization range, with sole focus of improving accuracy.

Common error correcting codes (ECCs such as SECDED), cannot correct \emph{multiple} bit errors per word (containing multiple DNN weights). However, for $p = 1\%$, the probability of two or more bit errors in a $64$-bit word is $13.5\%$.
Error detection via redundancy \cite{ReagenISCA2016} or supply voltage boosting \cite{ChandramoorthyHPCA2019} allow error-free low-voltage operation at the cost of additional energy or space. Therefore, \cite{KimDATE2018} proposes a co-design approach of training DNNs on \emph{profiled} SRAM bit errors. Similarly, for approximate DRAMs, \cite{KoppulaMICRO2019} combines profiled bit error training with a clever weight to DRAM mapping. These approaches work as the spatial bit error patterns can be assumed fixed for a \emph{fixed} accelerator \emph{and} voltage. The bit error pattern is obtained by post-silicon profiling and characterization of memories.
The random nature of variation-induced bit errors requires profiling to be carried out for each voltage, memory array and individual chip in order to obtain the corresponding bit error patterns. This makes training DNNs on profiled bit error patterns an expensive process. More importantly, we demonstrate that the obtained DNNs do \emph{not} generalize across voltages or to unseen bit error patterns, \eg, from other memory arrays. 
We propose \textbf{random bit error training (\Random)} which, in combination with weight clipping and robust quantization, obtains robustness against completely \emph{random} bit error patterns, see  \figref{fig:contributions} ({\color{colorbrewer4}violet}). Thereby, it generalizes across chips \emph{and} voltages, without any profiling, hardware-specific data mapping or other circuit-level mitigation strategies.

\textbf{Contributions:}
We combine our \textbf{robust fixed-point quantization \Quant}, \ie, reduced quantization range and robust implementation, with \textbf{weight clipping} and \textbf{random bit error training (\Random)} in order to obtain high robustness against low-voltage induced, random bit errors. We consider fixed-point quantization schemes in terms of robustness \emph{and} accuracy, instead of \emph{solely} focusing on accuracy as related work. Furthermore, we show that aggressive weight clipping, as regularization during training, is an effective strategy to improve robustness through redundancy. In contrast to \cite{KimDATE2018,KoppulaMICRO2019}, the robustness obtained through \Random generalizes across chips \emph{and} voltages, as evaluated on profiled SRAM bit error patterns from \cite{ChandramoorthyHPCA2019}. 
Finally, we discuss the involved trade-offs regarding robustness and accuracy and make our code publicly available to facilitate research in this highly applicable area of DNN robustness. \figref{fig:contributions} highlights key results on \CifarT: with $8$ bit and an increase in test error of less than $1\%$, roughly $20\%$ energy savings are possible. Combined with low-precision, \eg, for $4$ bit quantization, $30\%$ energy savings are possible at $p = 1\%$ with an increase in error rate of less than $2.5\%$. 

\begin{figure*}
	\centering
	\vspace*{-0.1cm}
	\begin{tikzpicture}
		\node[anchor=north west] at (0,0){\includegraphics[width=4cm]{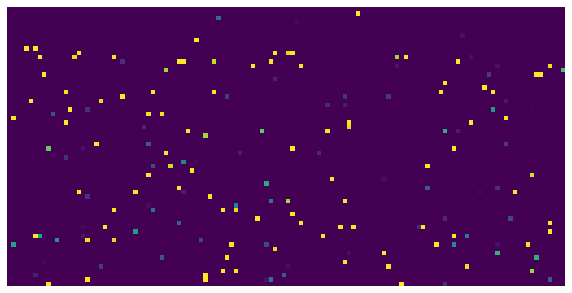}};
		\node[anchor=north west] at (4,0){\includegraphics[width=4cm]{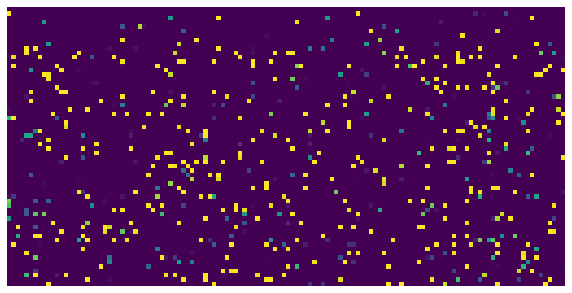}};
		\node[anchor=north west] at (8,0){\includegraphics[width=4cm]{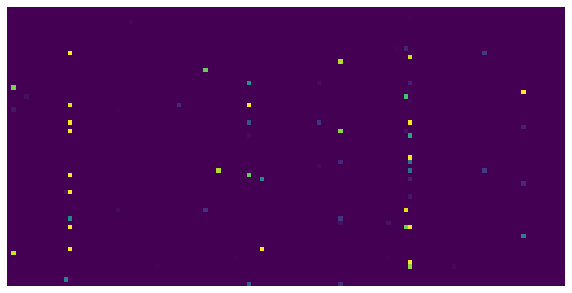}};
		\node[anchor=north west] at (12,0){\includegraphics[width=4cm]{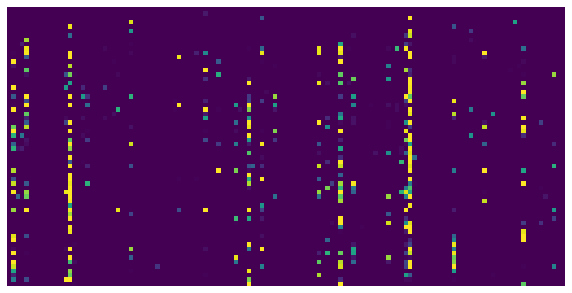}};
		
		\node[anchor=south east,fill opacity=0.75,fill=white] at (4, -2){$p{\approx}0.86\%$};
		\node[anchor=south east,fill opacity=0.75,fill=white] at (8, -2){$p{\approx}2.75\%$};
		\node[anchor=south east,fill opacity=0.75,fill=white] at (12, -2){$p{\approx}0.14\%$};
		\node[anchor=south east,fill opacity=0.75,fill=white] at (16, -2){$p{\approx}1.08\%$};
		
		\draw[white!30!black,-, line width=0.75pt] (8.125,-2.4) -- (8.125,0.4);
		\draw[thick,->] (0.05,0.05) -- (1,0.05);
		\draw[thick,->] (-0.05,-0.05) -- (-0.05,-1);
		\node[anchor=south] at (0.75,0.05){128 columns};
		\node[rotate=90,anchor=south] at (-0.05,-0.6){64 rows};
		
		\node[anchor=south] at (4,-0.15) {\bfseries Chip 1};
		\node[anchor=south] at (12,-0.15) {\bfseries Chip 2};
		
		\draw[thick,->] (2.5,-2.8) -- (4.5,-2.8) node[anchor=west] {bit error rate $p$ increases};
		\draw[thick,->] (2.5,-2.4) -- (4.5,-2.4) node[anchor=west] {voltage decreases};
		
		\draw[thick,-] (9.75, -2.2) -- (9.75,-2.4) -- (10.75,-2.4);
		\node[anchor=west] at (10.75, -2.4){bit errors subset of};
		\draw[thick,->] (13.75,-2.4) -- (14.75,-2.4) -- (14.75, -2.2);
	\end{tikzpicture}
	\vspace*{-8px}
	\caption{\textbf{Exemplary SRAM Bit Error Patterns.} Measured bit errors from two chips with on-chip SRAM (left and right), showing bit flip probability for a segment of size $64 \times 128$ bits: {\color{yellow!75!black}yellow} indicates a bit flip probability of one, {\color{violet}violet} indicates zero probability. We show measurements corresponding to two supply voltages. 
	With lower voltage, bit error rate increases. Also, the bit errors for higher voltage (= lower bit error rate) are a subset of those for lower voltage (= higher rate), \cf \secref{sec:errors}. Our error model randomly distributes bit errors across space. However, as example, we also show SRAM chip 2 which has a different spatial distribution with bit errors distributed along columns. We aim to obtain robustness across different memory arrays, voltages \emph{and} allowing arbitrary DNN weight to memory mappings.}
	\label{fig:errors}
\end{figure*}
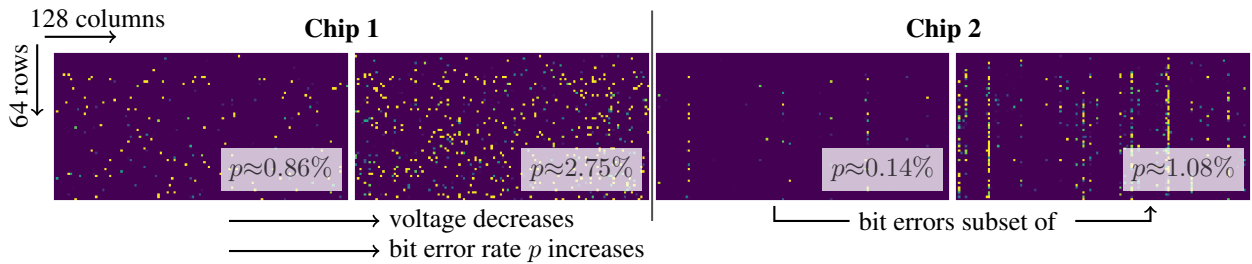

\textbf{Outline:} We review related work in \secref{sec:related-work} and provide a detailed description and discussion of the considered low-voltage bit error model in \secref{sec:errors}. In \secref{sec:robustness}, we discuss fixed-point quantization and its influence on bit error robustness and present weight clipping and \Random as effective strategies to improve robustness. Finally, \secref{sec:experiments} includes our experimental results. We conclude in \secref{sec:conclusion}.
\section{Related Work}
\label{sec:related-work}
 
We review prior work on quantization, low-voltage induced random bit errors and weight robustness; \revision{more in \appref{sec:supp-related-work}.}
 
\textbf{Quantization:} DNN Quantization \citep{GuoARXIV2018b} is usually motivated by faster 
DNN inference, \eg, through fixed-point quantization and arithmetic \citep{ShinICASSP2017,LinICML2016,LiNIPS2017}, and energy savings. To avoid reduced accuracy, quantization is considered during training \citep{JacobCVPR2018,KrishnamoorthiARXIV2018} instead of post-training or with fine-tuning \cite{GoncharenkoARXIV2018,BannerNIPS2019,nvtensorrt,nervana}, enabling low-bit quantization such as binary DNNs \citep{RastegariECCV2016,CourbariauxNIPS2015}. Some works also consider quantizing activations \citep{RastegariECCV2016,ChoiARXIV2018,HubaraJMLR2017} or gradients \citep{SeideINTERSPEECH2014,AlistarhARXIV2016,ZhouARXIV2016}. While works such as \citep{MurthyARXIV2019,MerollaARXIV2016,SungARXIV2015,AlizadehICLR2020} study the robustness of DNNs \emph{to} quantization, the robustness of various quantization schemes \emph{against} random bit errors has not been studied. This is in stark contrast to our findings that quantization impacts robustness significantly. Furthermore, works such as \citep{ZhuangCVPR2018,SungARXIV2015,ParkISCA2018} clip weight outliers to reduce approximation error of inliers, improving accuracy. In contrast, we consider \emph{weight clipping} independent of quantization \emph{as regularization during training} which spreads
out the weight distribution and improves robustness to bit errors.

\textbf{Bit Errors in DNN Accelerators:} Recent work \cite{GanapathyDAC2017,GanapathyHPCA2019} demonstrates that bit flips in SRAMs increase exponentially when reducing voltage below \Vmin. The authors of~\cite{ChandramoorthyHPCA2019} study the impact of bit flips in different layers of DNNs, showing severe accuracy degradation. Similar observations hold for DRAM \cite{ChangPOMACS2017}. To prevent accuracy drops at low voltages, \cite{ReagenISCA2016} combines SRAM fault detection with logic to set faulty data reads to zero. \cite{ChandramoorthyHPCA2019} uses supply voltage boosting for SRAMs to ensure error-free, low-voltage operation, while \cite{SrinivasanDATE2016} proposes storing critical bits in specifically robust SRAM cells. However, such methods incur power and area overhead. \revision{Thus, \cite{YangISQED2017} trains with a SRAM in the loop and} \cite{KimDATE2018,KoppulaMICRO2019} propose co-design approaches combining training on profiled SRAM/DRAM bit errors with hardware mitigation strategies and clever weight to memory mapping. Besides low-voltage operation for energy efficiency, recent work \cite{TangUSENIX2017} shows that an attacker can reduce voltage maliciously. Similarly, works such as \cite{KimISCA2014,MurdockSP2020} demonstrate software-based approaches to induce few, but targeted, bit flips in DRAM. In contrast to \cite{KimDATE2018,KoppulaMICRO2019}, our \emph{random bit error training} obtains robustness that generalizes across chips and voltages without expensive chip-specific profiling or hardware mitigation strategies. Furthermore, \cite{KimDATE2018,KoppulaMICRO2019} do not address the role of quantization and we demonstrate that these approaches can benefit from our weight clipping, as well. We show that energy savings from low-voltage operation and low-precision \cite{ParkISCA2018} can be combined.

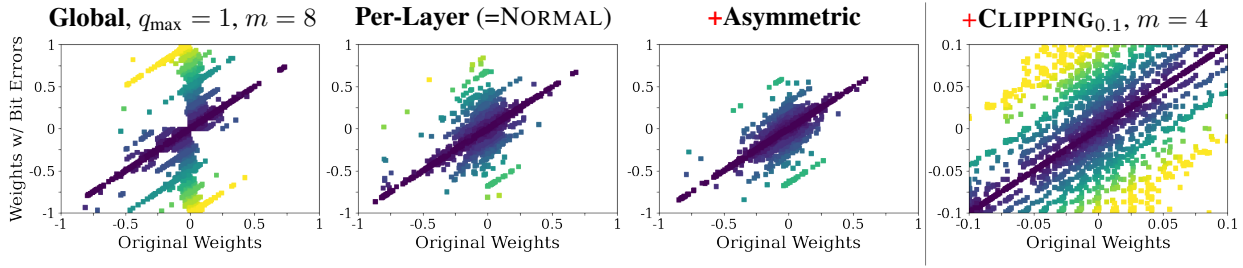
\begin{figure*}[t]
	\centering
	\vspace*{-0.1cm}
	\begin{tikzpicture}
		\node[anchor=north west] at (-0.25,0){\includegraphics[height=3cm]{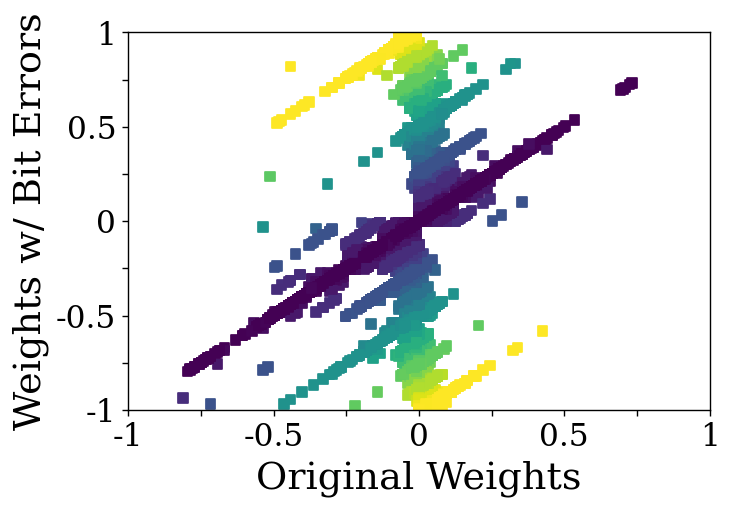}};
		\node[anchor=north west] at (4,0){\includegraphics[height=3cm,trim=1cm 0 0 0,clip]{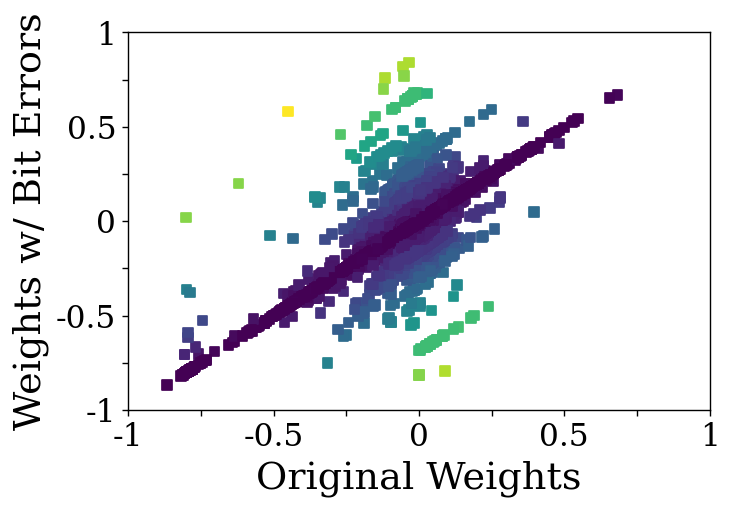}};
		\node[anchor=north west] at (8,0){\includegraphics[height=3cm,trim=1cm 0 0 0,clip]{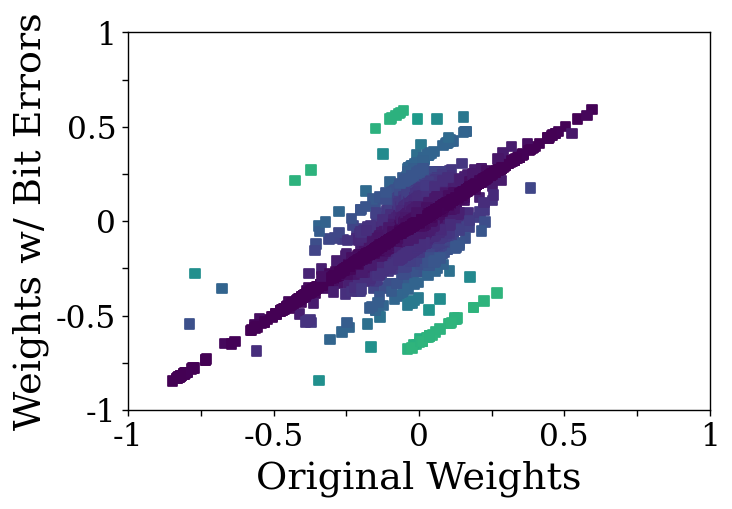}};
		\node[anchor=north west] at (12,0){\includegraphics[height=3cm,trim=1cm 0 0 0,clip]{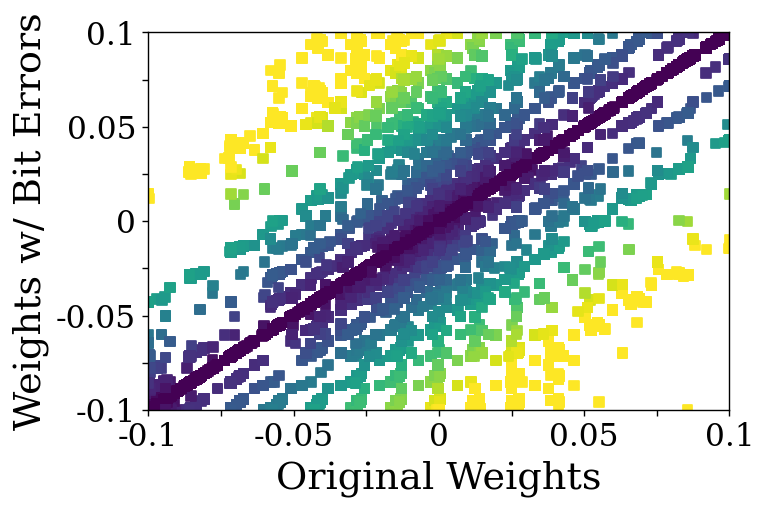}};
		
		\draw[-,white!50!black] (12.125, 0.25) -- (12.125, -3.25);
		\node[anchor=south,xshift=0.25cm,yshift=-0.25cm] at (2, 0){\textbf{Global}, $\qmax=1$, $m = 8$};
		\node[anchor=south,xshift=0.25cm,yshift=-0.25cm] at (6, 0){\textbf{Per-Layer} (=\Normal)};
		\node[anchor=south,xshift=0.25cm,yshift=-0.25cm] at (10, 0){\textbf{{\color{red}+}Asymmetric}};
		\node[anchor=south,xshift=0.25cm,yshift=-0.25cm] at (14, 0){{\color{red}+}\textbf{\Clipping[$0.1$]}, $m = 4$};
	\end{tikzpicture} 
	\vspace*{-6px}
	\caption{\textbf{Quantization and Random Bit Errors.} Original weights (x-axis) plotted against perturbed weights with bit errors (y-axis), for different 
	fixed-point quantization schemes with $m = 8$ bit (left) and $p=2.5\%$. We also show the $m = 4$ bit case with \Clipping at $\wmax = 0.1$, \cf \secref{subsec:robustness-clipping}. Color indicates absolute error: from zero {\color{violet}violet} to the maximal possible error {\color{yellow!75!black}yellow} of $1$ (left) and $0.1$ (right). Asymmetric per-layer quantization reduces the impact of bit errors compared to a the symmetric per-layer/global quantization. Clipping reduces absolute error, but the errors \emph{relative} to $\wmax$ increase.}
	\label{fig:quantization}
	\vspace*{-0.2cm}
\end{figure*}

\textbf{Weight Robustness:} Only few works consider weight robustness: \citep{WengAAAI2020} certify the robustness of weights with respect to $L_\infty$ perturbations and \citep{CheneyARXIV2017} study Gaussian noise on weights. \citep{RakinICCV2019,HeCVPR2020} consider identifying and (adversarially) flipping few vulnerable bits in quantized weights.
Fault tolerance, in contrast, describes structural changes such as removed units, and is rooted in early work such as \citep{NetiTNN1992,Chiu1994}.
Finally, \citep{JiCCS2018,DumfordARXIV2018} explicitly manipulate weights to integrate backdoors. We study robustness against \emph{random bit errors}, exhibiting a unique noise pattern, \cf \figref{fig:quantization}.

\section{Low-Voltage Induced Random Bit Errors in Quantized DNN Weights}
\label{sec:errors}

We assume the quantized DNN weights to be stored  on multiple memory banks, \eg, SRAM in the case of on-chip scratchpads or DRAM for off-chip memory. As shown in \cite{GanapathyDAC2017,KimDATE2018,ChandramoorthyHPCA2019}, the probability of memory bit cell failures increases exponentially as operating voltage is scaled below $\Vmin$, \ie, the minimal voltage required for reliable operation, see \figref{fig:introduction}. This is done intentionally to reduce energy consumption, \eg, \cite{ChandramoorthyHPCA2019,KimDATE2018,KoppulaMICRO2019}, or adversarially by an attacker, \eg, \cite{TangUSENIX2017}. Process variation during fabrication causes a variation in the vulnerability of individual bit cells. As shown in \figref{fig:errors} (left), for a specific memory array, bit cell failures are typically approximately random and independent of each other \cite{GanapathyDAC2017}. We also consider chips showing other error patterns as in \figref{fig:errors} (right). Nevertheless, there is generally an ``inherited'' distribution of bit cell failures across voltages \cite{GanapathyHPCA2019}, if a bit error occurred at a given voltage, it is likely to occur at lower voltages, as made explicit in \figref{fig:errors}. However, across different SRAM arrays in a chip or different chips, the patterns or spatial distribution of bit errors is usually different and can be assumed random \cite{ChandramoorthyHPCA2019}. Throughout the paper, we use the following bit error model:

\textbf{Random Bit Error Model:}
\textit{The probability of a bit error is $p$ (in \%) for all weight values and bits. For a fixed memory array, bit errors are persistent across supply voltages, \ie, bit errors at probability $p'{\leq}p$ also occur at probability $p$. A bit error flips the currently stored bit. We denote random bit error injection by $\biterror_p$.}

\begin{figure*}
    \vspace*{-0.15cm}
    \centering
    \includegraphics[width=0.85\textwidth]{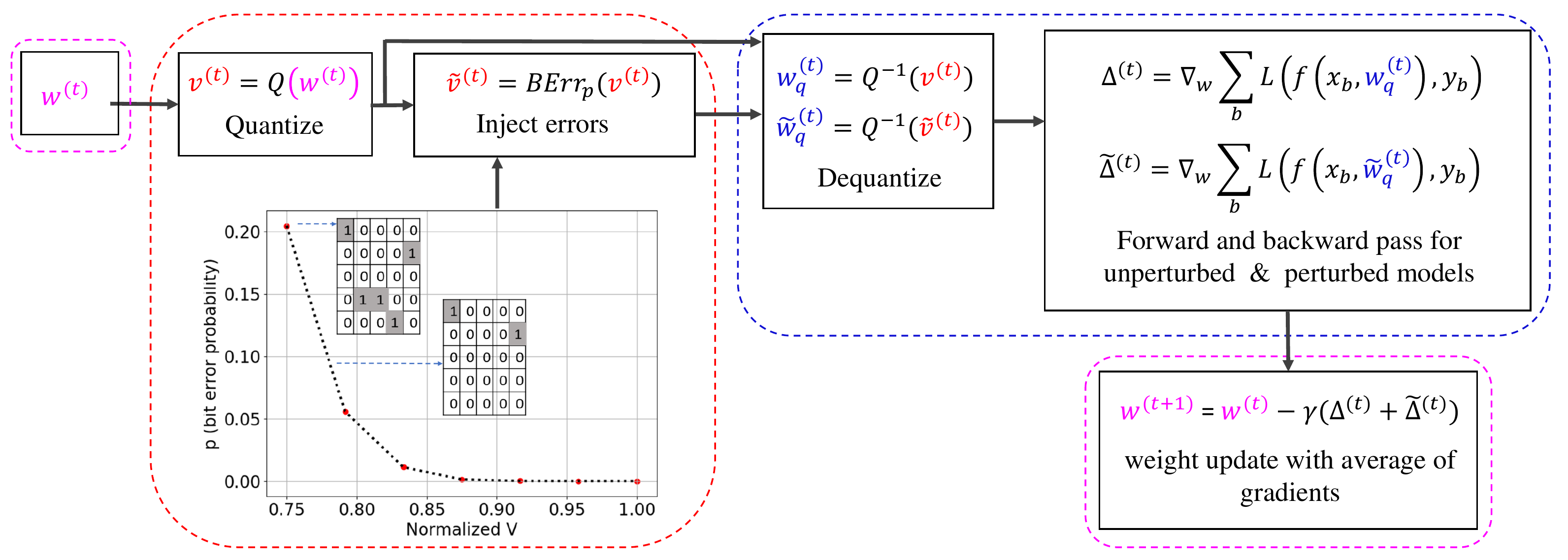}
    \vspace*{-6px}
    \caption{\textbf{Random Bit Error Training (\Random).} We illustrate the data-flow for \Random as in \algref{alg:training}. Here, $\biterror_p$ injects random bit errors in the \red{quantized weights} $\red{v^{(t)}} = Q(\magenta{w^{(t)}})$, resulting in $\red{\tilde{v}^{(t)}}$, while the forward pass is performed on the \blue{de-quantized perturbed weights} $\blue{\tilde{w}_q^{(t)}} = Q^{-1}(\red{\tilde{v}^{(t)}})$, \ie, fixed-point arithmetic is not emulated. The weight update during training is not affected by bit errors and computed in \magenta{floating point}.}
    \label{fig:flowchart}
    \vspace*{-0.2cm}
\end{figure*}

This error model realistically captures the nature of low-voltage induced bit errors, from both SRAM and DRAM as confirmed in \cite{ChandramoorthyHPCA2019,KimDATE2018,KoppulaMICRO2019}. However, our approach in \secref{sec:robustness} is model-agnostic: the error model can be refined if extensive memory characterization results are available for individual chips. For example, faulty bit cells with $1$-to-$0$ or $0$-to-$1$ flips might not be equally likely. Similarly, as in \cite{KoppulaMICRO2019}, bit errors might be biased towards alignment along rows or columns of the memory array. The latter case is illustrated in \figref{fig:errors} (right). However, estimating these specifics requires testing infrastructure and detailed characterization of individual chips. 
More importantly, it introduces the risk of overfitting to few specific memories/chips. 
Furthermore, we demonstrate that the robustness obtained using our uniform error model generalizes to bit error distributions with strong spatial biases as in \figref{fig:errors} (right).

We assume the quantized weights to be mapped linearly to the memory. This is the most direct approach and, in contrast to \cite{KoppulaMICRO2019}, does not require knowledge of the exact spatial distribution of bit errors. This also means that we do not map particularly vulnerable weights to more reliable memory cells, and therefore no changes to the hardware or the application are required. Thus, in practice, for $W$ weights and $m$ bits per weight value, we sample uniformly $u \sim U(0, 1)^{W \times m}$. Then, the $j$-th bit in the quantized weight $v_i = Q(w_i)$ is flipped iff $u_{ij} \leq p$.
Our model assumes that the flipped bits at lower probability $p' \leq p$ are a subset of the flipped bits at probability $p$ and that bit flips to $1$ and $0$ are equally likely. The noise pattern of random bit errors is illustrated in \figref{fig:quantization}: 
for example a single bit flip in the most-significant bit (MSB) of the signed integer $v_i$ can result in a change of roughly half of the quantized range (also \cf \secref{subsec:robustness-quantization}).

\section{Towards Robustness Against Random Bit Errors}
\label{sec:robustness}

We address robustness against random bit errors in three steps: First, we analyze the impact of fixed-point quantization schemes on bit error robustness. This has been neglected both in prior work on low-voltage DNN accelerators \cite{KimDATE2018,KoppulaMICRO2019} and in work on quantization robustness \cite{MurthyARXIV2019,MerollaARXIV2016,SungARXIV2015}. This yields our \textbf{robust quantization} (\secref{subsec:robustness-quantization}). On top, we propose aggressive \textbf{weight clipping} as regularization during training (\secref{subsec:robustness-clipping}). 
Weight clipping enforces a more uniformly distributed, \ie, redundant, weight distribution, improving robustness. We show that this is due to minimizing the cross-entropy loss, enforcing large logit differences.
Finally, in addition to robust quantization and weight clipping, we perform \textbf{random bit error training (\Random)} (\secref{subsec:robustness-training}): in contrast to the fixed bit error patterns in \cite{KimDATE2018,KoppulaMICRO2019}, we train on completely \emph{random} bit errors and, thus, generalize across chips and voltages.
Generalization is measured using \emph{average robust test error (\RTE)}, the test error after injecting bit errors, \wrt to our error model from \secref{sec:errors} as well as real, profiled bit error patterns. 
Robustness against bit error rate $p$ has to induce robustness for $p' \leq p$ (\ie, higher voltage), as well.

\subsection{Robust Fixed-Point Quantization}
\label{subsec:robustness-quantization}

We consider quantization-aware training \cite{JacobCVPR2018,KrishnamoorthiARXIV2018} using a generic, deterministic fixed-point quantization scheme commonly used in DNN accelerators \cite{ChandramoorthyHPCA2019}. However, we focus on the impact of quantization schemes on robustness against random bit errors, mostly neglected so far \cite{MurthyARXIV2019,MerollaARXIV2016,SungARXIV2015}. We find that quantization affects robustness significantly, even if accuracy is largely unaffected.

\textbf{Fixed-Point Quantization:} Let $f(x; w)$ be a DNN taking an example $x \in [0, 1]^D$, \eg, an image, and weights $w \in \mathbb{R}^W$ as input. Quantization determines how weights are represented in memory, \eg, on SRAM. In a \emph{fixed-point quantization} scheme, $m$ bits allow to represent $2^m$ distinct values. 
A weight $w_i \in [-\qmax, \qmax]$ 
is represented by a signed $m$-bit integer $v_i = Q(w_i)$ corresponding to the underlying bits. Here, $[-\qmax, \qmax]$ is the \emph{symmetric} quantization range and signed integers use two's complement representation. Then, $Q: [-\qmax, \qmax] \mapsto \{-2^{m - 1} - 1, \ldots, 2^{m - 1} - 1\}$ is defined as 
\begin{align}
    Q(w_i) = \left\lfloor \frac{w_i}{\Delta}\right\rfloor,\text{  }
    Q^{-1}(v_i) = \Delta v_i,\text{  }
    \Delta = \frac{\qmax}{2^{m - 1} - 1}
    \label{eq:quantization}
\end{align}
Flipping the most significant bit (MSB, \ie, sign bit) leads to an absolute error of half the quantization range, \ie, $\qmax$ ({\color{yellow!75!black!}yellow} in \figref{fig:quantization}).
Flipping the least significant bit (LSB) incurs an error of $\Delta$, \cf \eqnref{eq:quantization}. Thus, the impact of bit errors ``scales with'' $\qmax$.

\textbf{Global and Per-Layer Quantization:} $\qmax$ can be chosen to accommodate all weights, \ie, $\qmax = \max_i |w_i|$. This is called \emph{global} quantization. However, it has become standard to apply quantization \textit{per-layer} allowing to adapt $\qmax$ to each layer. As in PyTorch \cite{PaszkeNIPSWORK2017}, we consider weights and biases of each layer separately. By reducing the quantization range for each layer individually, the errors incurred by bit flips are automatically minimized, \cf \figref{fig:quantization}. The
\textbf{per-layer, symmetric quantization is our default reference}, referred to as \Normal. However, it turns out that it is further beneficial to consider arbitrary quantization ranges $[\qmin, \qmax]$ (allowing $\qmin > 0$). In practice, we
first map $[\qmin, \qmax]$ to $[-1,1]$ and then quantize $[-1,1]$ using \eqnref{eq:quantization}.
Overall, per-layer asymmetric quantization has the finest granularity, \ie, lowest $\Delta$ and approximation error. Nevertheless it is not the most robust
quantization.

\begin{figure}[t]
	\centering
	\vspace*{-0.1cm}
	\hspace*{-0.3cm}
	\begin{subfigure}{0.24\textwidth}
		\vspace*{3px}
		
		\includegraphics[height=1.35cm]{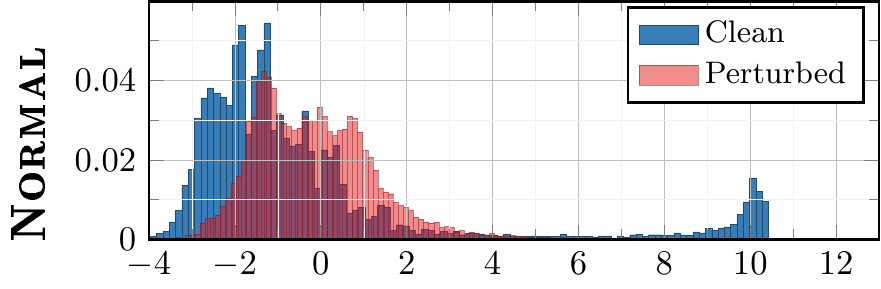}
	\end{subfigure}
	\begin{subfigure}{0.12\textwidth}
		\vspace*{0px}
		
		\includegraphics[height=1.425cm]{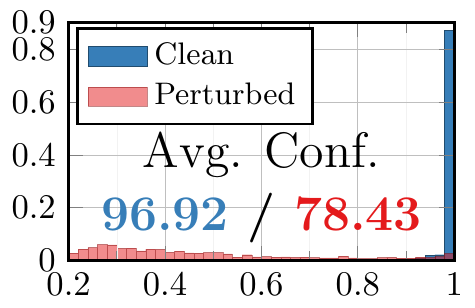}
	\end{subfigure}
	\begin{subfigure}{0.12\textwidth}
		\vspace*{0px}
		
		\includegraphics[height=1.425cm]{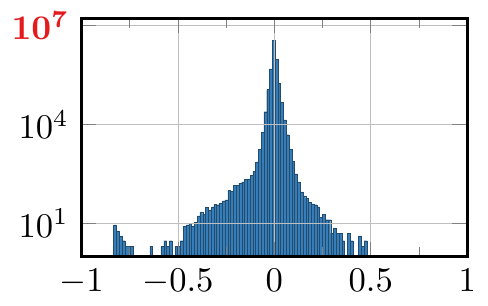}
	\end{subfigure}
	
	\hspace*{-0.4cm}
	\begin{subfigure}{0.24\textwidth}
		\vspace*{3px}
		
		\includegraphics[height=1.35cm]{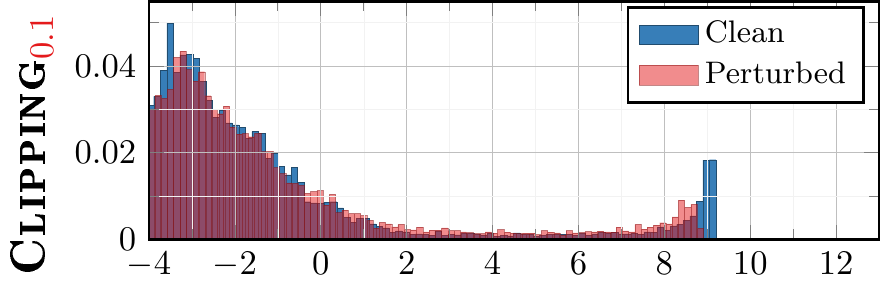}
	\end{subfigure}
	\begin{subfigure}{0.12\textwidth}
		\vspace*{0px}
		
		\includegraphics[height=1.425cm]{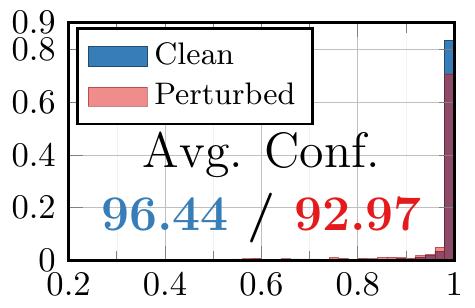}
	\end{subfigure}
	\begin{subfigure}{0.12\textwidth}
		\vspace*{3px}
		
		\includegraphics[height=1.425cm]{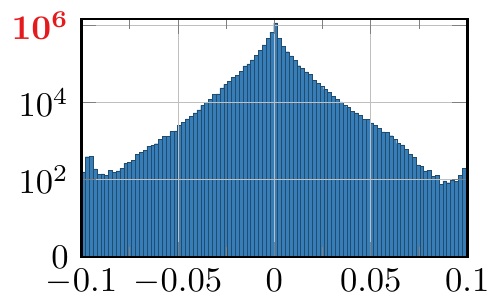}
	\end{subfigure}
	
	\hspace*{-0.3cm}
	{\color{black!25!white}\rule{0.5\textwidth}{0.5px}}
	
	\hspace*{-0.4cm}
	\begin{subfigure}{0.24\textwidth}
		\vspace*{0px}
		
		\includegraphics[height=1.825cm]{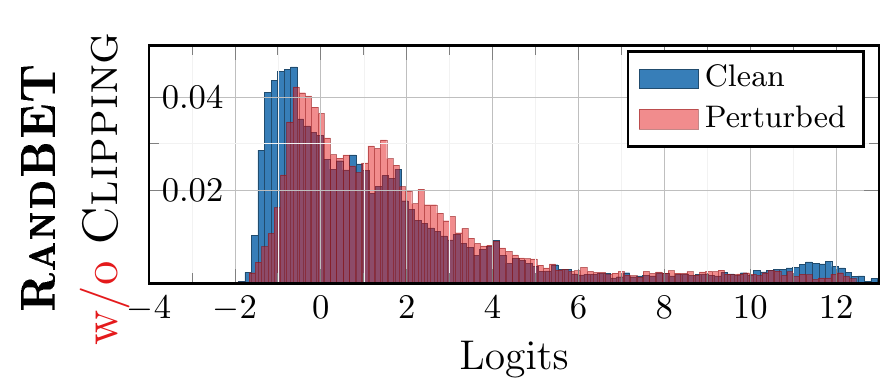}
	\end{subfigure}
	\begin{subfigure}{0.12\textwidth}
		\vspace*{0px}
		
		\includegraphics[height=1.7cm]{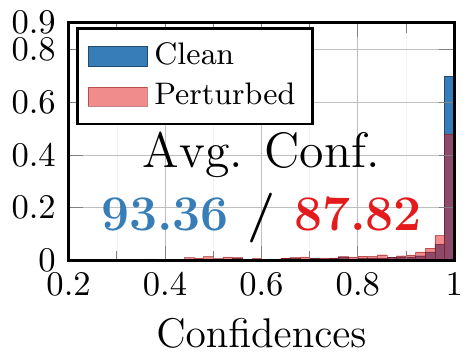}
	\end{subfigure}
	\begin{subfigure}{0.12\textwidth}
		\vspace*{3px}
		
		\includegraphics[height=1.7cm]{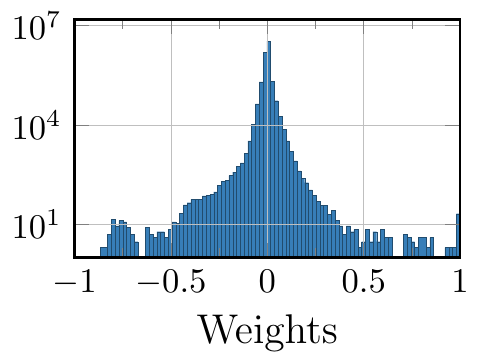}
	\end{subfigure}
	
	\vspace*{-8px}
	\caption{\textbf{Effect of Weight Clipping.} On \CifarT, weight clipping constraints the weights (right), thereby implicitly limiting the possible range for logits (left, {\color{colorbrewer2}blue}).
	However, even for $\wmax = 0.1$ the DNN is able to produce high confidences (middle, {\color{colorbrewer2}blue}), suggesting that more weights are used to obtain these logits. Furthermore, the impact of random bit errors, $p = 1\%$, on the logits/confidences ({\color{colorbrewer1}red}) is reduced significantly. \Random (trained with $p = 1\%$, w/o weight clipping), increases the range of weights and is less effective at preserving logit/confidence distribution. 
	}
	\label{fig:clipping}
	\vspace*{-0.2cm}
\end{figure}

\textbf{Robust Quantization:} Quantization as in \eqnref{eq:quantization} does \emph{not} provide optimal robustness against bit errors. First, the floor operation $\lfloor \nicefrac{w_i}{\Delta}\rfloor$ is commonly implemented as float-to-integer conversion. Using proper rounding $\lceil\nicefrac{w_i}{\Delta}\rfloor$ instead has negligible impact on accuracy, even though approximation error improves slightly. In stark contrast, bit error robustness is improved considerably. During training, DNNs can compensate the differences in approximation errors, even for small precision $m < 8$. However, at test time,
rounding decreases the impact of bit errors considerably. Second, \eqnref{eq:quantization} uses signed integers for symmetric quantization. For asymmetric quantization, with arbitrary $[\qmin, \qmax]$, we found quantization into \emph{unsigned} integers to improve robustness, \ie, $Q : [\qmin, \qmax] \mapsto \{{\color{red}0}, \ldots, {\color{red}2^m - 1}\}$. This is implemented using an additive term of $2^{m-1}-1$ in \eqnref{eq:quantization}. While accuracy is not affected, the effect of bit errors in the sign bit changes: in symmetric quantization, the sign bit mirrors the sign of the weight value. For asymmetric quantization, an unsigned integer representation is more meaningful. Overall, our \textbf{robust fixed-point quantization (\Quant)}  uses per-layer, asymmetric quantization into unsigned integers with rounding.
These seemingly \emph{small differences} have little to no impact on accuracy, while having tremendous impact on robustness against bit errors, see \secref{subsec:experiments-quantization} and \appref{sec:supp-implementation}. 
\revision{They are simple to implement, do not add training complexity or hyper-parameters and demonstrate the importance of robustness in developing DNN quantization schemes.}

\begin{algorithm}[t]
\caption{\textbf{Random Bit Error Training (\Random).} The forward passes are performed using \blue{de-quantized weights (blue)}. Perturbed weights are obtained by injecting bit errors in the \red{quantized weights (in red)}. The update, averaging gradients from both forward passes, is performed in \magenta{floating-point (magenta)}. Also see \figref{fig:flowchart}.}
\label{alg:training}
\begin{algorithmic}[1]
\small
\Procedure{RandBET}{$p$}
    \State initialize \magenta{$w^{(0)}$}
	\For{$t = 0, \ldots, T - 1$}
    	\State sample batch $\{(x_b, y_b)\}_{b = 1}^B$
    	\State \{element-wise clipping:\}
    	\State $\magenta{w^{(t)}} = \min(\wmax, \max(-\wmax, \magenta{w^{(t)}}))$
    	\State \{quantization:\}
    	\State $\red{v^{(t)}} = Q(\magenta{w^{(t)}})$
        \State $\blue{w_q^{(t)}} = Q^{-1}(\red{v^{(t)}})$
        \State \{clean forward and backward pass:\}
        \State $\Delta^{(t)} = \nabla_w \sum_{b = 1}^B \mathcal{L}(f(x_b; \blue{w_q^{(t)}}), y_b)$
        \State \{\emph{perturbed} forward and backward pass:\}
        \State $\blue{\tilde{w}_q^{(t)}}{\hskip 1px=\hskip 1px}Q^{-1}(\biterror_p(\red{v^{(t)}}))$\label{line:attack} \{inject random bit errors\}
        \State $\tilde{\Delta}^{(t)} = \nabla_w \sum_{b = 1}^B \mathcal{L}(f(x_b; \blue{\tilde{w}_q^{(t)}}), y_b)$
        \State \{average gradients and weight update:\}
        \State $\magenta{w^{(t + 1)}} = \magenta{w^{(t)}} - \gamma(\Delta^{(t)} + \tilde{\Delta}^{(t)})$
    \EndFor
    \State \textbf{return} $\blue{w_q^{(T)}} = Q^{-1}(Q(\magenta{w^{(T)}}))$
\EndProcedure
\end{algorithmic}
\end{algorithm}

\subsection{Training with Weight Clipping as Regularization}
\label{subsec:robustness-clipping}

\textbf{Weight clipping} refers to constraining the weights to $[-\wmax, \wmax]$ \emph{during training}, where $\wmax$ is a hyper-parameter. Generally, $\wmax$ is independent of the quantization range(s) which always adapt(s) to the weight range(s) at hand. However, weight clipping limits the maximum possible quantization range (\cf \secref{subsec:robustness-quantization}), \ie, $\qmax \leq \wmax$.
It might seem that weight clipping with small $\wmax$ automatically improves robustness against bit errors as the absolute errors are reduced. However, the \emph{relative} errors are not influenced by rescaling. As the DNN's decision is usually invariant to rescaling, reducing the scale of the weights does not impact robustness. In fact, the mean relative error of the weights in \figref{fig:quantization} (right) increased with clipping at $\wmax = 0.1$. Thus, weight clipping does \emph{not} ``trivially'' improve robustness by reducing the scale of weights. Nevertheless, we found that weight clipping actually improves robustness considerably on top of our robust quantization.

The interplay of weight clipping and minimizing the the cross-entropy loss during training is the key. High confidences can only be achieved by large differences in the logits. Because the weights are limited to $[-\wmax, \wmax]$, large logits can only be achieved using more weights in each layer to produce larger outputs. This is illustrated in \figref{fig:clipping} (right): using $\wmax = 0.1$, the weights are (depending on the layer) up to $5$ times smaller. Considering deep NNs, the ``effective'' scale factor for the logits is significantly larger, scaling exponentially with the number of layers. Thus, using $\wmax = 0.1$ is a significant constraint on the DNNs ability to produce large logits. As result, weight clipping produces a much more uniform weight distribution.
\figref{fig:clipping} (left and middle) shows that a DNN constrained at $\wmax = 0.1$ can produce similar logit and confidence distributions (in {\color{colorbrewer2}blue}) as the unclipped DNN. At the same time, random bit errors, have a significantly smaller impact on the logits and confidences (in {\color{colorbrewer1}red}). \figref{fig:clipping} (right column) also shows the induced redundancy in the weight distribution. Weight clipping leads to more weights being utilized, \ie, less weights are zero (note log-scale, marked in {\color{colorbrewer1}red}, on the y-axis). Also, more weights reach large values, relative to the maximum absolute weight. Overall, we found weight clipping to be an easy-to-use but effective measure to improve weight robustness. We use \Clipping[$\wmax{=}0.1$] to refer to, \eg, weight clipping with $\wmax = 0.1$. For more evidence supporting our argumentation, see \tabref{tab:clipping-robustness}. For example, we show that DNNs loose robustness when using label smoothing, \ie, not enforcing high confidences/logits during training.
\revision{Finally, weight clipping is straight-forward to implement (\cf \algref{alg:training}, line 6) and adds negligible training cost. The additional hyper-parameter, \ie, $\wmax$, can easily be tuned based on constraints on clean (or robust) performance, \cf \secref{subsec:experiments-clipping}.}

\subsection{Random Bit Error Training (\Random)}
\label{subsec:robustness-training}

In \emph{addition to} weight clipping and robust quantization, we inject random bit errors with probability $p$ during training to further improve robustness. This results in the following learning problem, which we optimize as illustrated in \figref{fig:flowchart}:
\begin{align}
	\begin{split}
    	&\min_w \mathbb{E}[\mathcal{L}(f(x; \tilde{w}), y) + \mathcal{L}(f(x; w), y)]\\
    	\text{s.t.}&\quad v = Q(w),\, \tilde{v} = \biterror_p(v),\, \tilde{w} = Q^{-1}(\tilde{v}).
   	\end{split}\label{eq:random-training-average}
\end{align}
where $(x, y)$ are labeled examples, $\mathcal{L}$ is the cross-entropy loss and $v = Q(w)$ denotes the (element-wise) quantized weights $w$ which are to be learned. $\biterror_p(v)$ injects random bit errors with rate $p$ in $v$. Note that we consider both the loss on clean weights and weights with bit errors. This is desirable to avoid an increase in (clean) test error and stabilizes training compared to training only on bit errors in the weights. Note that bit error rate $p$ implies, in expectation, $pmW$ bit errors. Following \algref{alg:training}, we use stochastic gradient descent to optimize \eqnref{eq:random-training-average}, by performing the gradient computation using the perturbed weights $\tilde{w} = Q^{-1}(\tilde{v})$ with $\tilde{v} = \biterror_p(v)$, while applying the gradient update on the (floating-point) clean weights $w$. In spirit, this is similar to data augmentation, however, the perturbation is applied on the weights instead of the inputs. As we found that introducing bit errors right from the start may prevent the DNN from converging, we apply bit errors as soon as the (clean) cross-entropy loss is below $1.75$. Interestingly, weight clipping and \Random have somewhat orthogonal effects, which allows to combine them easily in practice: While weight clipping encourages redundancy in weights by constraining them to $[-\wmax,\wmax]$, 
\Random (w/o weight clipping) causes the DNN to have larger tails in the weight distribution, as shown in \figref{fig:clipping} (bottom). However, considering logits and confidences, especially with random bit errors (in {\color{colorbrewer1}red}), \Random alone performs slightly worse than \Clipping[$0.1$]. Thus, \Random becomes particularly effective when combined with weight clipping, as we make explicit using the notation \Random[$\wmax$] in \algref{alg:training}.
\revision{While \Random increases training complexity (\cf \algref{alg:training}), inference is \emph{not} affected. This is in stark contrast to hardware- or redundancy-based bit error mitigation strategies which usually impact inference time and energy consumption. We also note that the additional hyper-parameter, \ie, $p$, is easily chosen according to the target bit error rate.}
\section{Experiments}
\label{sec:experiments}

We present experiments on \MNIST \citep{LecunIEEE1998} and \Cifar \citep{Krizhevsky2009}. We first analyze the impact of fixed-point quantization schemes on robustness (\secref{subsec:experiments-quantization}). Subsequently, we discuss weight clipping (\Clipping, \secref{subsec:experiments-clipping}), showing that improved robustness originates from increased redundancy in the weight distribution. Then, we focus on random bit error training (\Random, \secref{subsec:experiments-randbet}). We show that related work \cite{KimDATE2018,KoppulaMICRO2019} does not generalize, while \Random generalizes across chips and voltages, as demonstrated on profiled bit error patterns from different chips. \secref{subsec:experiments-discussion} summarizes our results for various precisions $m$.

\begin{table}
	\centering
	\small 
	\caption{\textbf{Robust Quantization}. \RTE for random bit errors at $p = 0.05\%$ and $p = 0.5\%$
	for different quantization schemes, \cf \secref{subsec:robustness-quantization}. Minor differences
	can have large impact on \RTE while clean test error is unaffected. For $8$ bit the second row shows \Normal quantization (symmetric/per-layer) whereas the last row is our \Quant.
	*\Clipping[$0.1$]+\Quant with and without rounding.
	}
	\label{tab:quantization-robustness}
	\vspace*{-0.25cm}
	\hspace*{-0.2cm}
	\begin{tabular}{| c | l | c | c | c |}
		\hline
		\multicolumn{2}{|c|}{Quantization Schemes} & \multirow{2}{*}{\begin{tabular}{@{}c@{}}\TE\\in \%\end{tabular}}& \multicolumn{2}{c|}{\RTE in \%}\\
		\cline{4-5} 
		\multicolumn{2}{|c|}{(
		\CifarT)} && $p{=}0.05$ & $p{=}0.5$\\
		\hline
		\hline
		\multirow{5}{*}{\rotatebox{90}{$8$ bit}} & \eqnref{eq:quantization}, global & 4.63 & 86.01 {\color{gray}\scriptsize ${\pm}$3.65} & 90.71 {\color{gray}\scriptsize ${\pm}$0.49}\\
		& \eqnref{eq:quantization}, per-layer & 4.36 & 5.51 {\color{gray}\scriptsize ${\pm}$0.19} & 24.76 {\color{gray}\scriptsize ${\pm}$4.71}\\
		& +asymmetric & 4.36 & 6.47 {\color{gray}\scriptsize ${\pm}$0.22} & {\color{colorbrewer1}40.78} {\color{gray}\scriptsize ${\pm}$7.56}\\
		& +unsigned & 4.42 & 6.97 {\color{gray}\scriptsize ${\pm}$0.28} & 17.00 {\color{gray}\scriptsize ${\pm}$2.77}\\
		& +rounding (=\Quant) & 4.32 & \bfseries 5.10 {\color{gray}\scriptsize ${\pm}$0.13} & \bfseries 11.28 {\color{gray}\scriptsize ${\pm}$1.47}\\
		\hline
		\hline
		\multirow{2}{*}{\rotatebox{90}{$4$ bit}} & w/o rounding* & 5.81 & 90.40 {\color{gray}\tiny ${\pm}$0.21} & 90.36 {\color{gray}\tiny ${\pm}$0.2}\\
		& w/ rounding* & \bfseries 5.29 & \bfseries 5.75 {\color{gray}\tiny ${\pm}$0.06} & \bfseries 7.71 {\color{gray}\tiny ${\pm}$0.36}\\
		\hline
	\end{tabular}
	\vspace*{-0.2cm}
\end{table}

\textbf{Metrics:} We report (clean) test error \TE (lower is better, $\downarrow$), corresponding to \emph{clean} weights, and \textbf{robust test error \RTE} ($\downarrow$) which is the 
\textbf{test error after injecting bit errors into the weights}. As the
bit errors are random we report the \emph{average} \RTE and its standard deviation for $50$ samples of random bit errors with rate $p$ as detailed in \secref{sec:errors}.

\textbf{Architecture:} We use SimpleNet \citep{HasanpourARXIV2016}, providing comparable performance to ResNets \cite{HeCVPR2016} with only $W{=}5.5\text{Mio}$ weights on \CifarT. On \MNIST, we halve all channel widths, resulting in roughly $1\text{Mio}$ weights. On \CifarH, we use a Wide ResNet (WRN) \cite{ZagoruykoBMVC2016}. As batch normalization (BN) \cite{IoffeICML2015} yields
consistently worse robustness against bit errors we use group normalization (GN) \cite{WuECCV2018}, see 
\appref{subsec:supp-experiments-bn}.

\textbf{Training:} We use stochastic gradient descent with an initial learning rate of $0.05$, multiplied by $0.1$ after $\nicefrac{2}{5}$, $\nicefrac{3}{5}$ and $\nicefrac{4}{5}$ of $100$/$250$ epochs on \MNIST/\Cifar. On \Cifar, we whiten the input images and use AutoAugment \cite{CubukARXIV2018} with Cutout \cite{DevriesARXIV2017}. For \Random, random bit error injection starts when the loss is below 1.75 on \MNIST/\CifarT or 3.5 on \CifarH. Normal training with the standard and our robust quantization are denoted \Normal and \Quant, respectively. Weight clipping with $\wmax$ is referred to as \Clipping[\wmax] or together with \Random as \Random[\wmax]. 
For \Quant, $m = 8$, we obtain $4.3\%$ on \CifarT and $18.5\%$ \TE on \CifarH. On \MNIST, $0.47\%$ are possible even for $m = 2$. 

\begin{table}
	\centering
	\caption{\textbf{Weight Clipping Robustness.} Clean \TE and \RTE as well as clean confidence and confidence at $p{=}1\%$ bit errors (in \%, higher is better, $\uparrow$) for \Clipping and \Clipping with label smoothing (+LS). \TE increases for $\wmax = 0.025$ where the DNN is not able to produce large (clean) confidences. LS consistently reduces robustness, indicating that robustness is due to enforcing high confidence during training \emph{and} weight clipping.}
	\label{tab:clipping-robustness}
	\vspace*{-0.25cm} 
	\small 
	\hspace*{-0.2cm}
	\begin{tabular}{| l | c | c | c | c | c |}
		\hline
		Model & \multirow{2}{*}{\begin{tabular}{@{}c@{}}\TE\\in \%\end{tabular}} & \multirow{2}{*}{\begin{tabular}{@{}c@{}}Conf\\in \%\end{tabular}} & \multirow{2}{*}{\begin{tabular}{@{}c@{}}Conf\\$p{=}1$\end{tabular}} & \multicolumn{2}{c|}{\RTE in \%}\\
		\cline{5-6}
		(\CifarT) & & & & $p{=}0.1$ & $p{=}1$\\
		\hline 
		\hline
		\Quant & \bfseries 4.32 & \bfseries 97.42 & 78.43 & 5.54 & 32.05\\
		\hline
		\Clipping[$0.15$] & 4.42 & 96.90 & 88.41 & \bfseries  5.31 & 13.08\\
		\Clipping[$0.1$] & 4.82 & 96.66 & 92.97 & 5.58 & 8.93\\
		\Clipping[$0.05$] & 5.44 & 95.90 & \bfseries 94.73 & 5.90 & \bfseries 7.18\\
		\Clipping[$0.025$] & 7.10 & {\color{colorbrewer1}84.69} & 83.28 & 7.40 & 8.18\\
		\hline
		\Clipping[$0.15$]+LS & 4.67 & 88.22 & 47.55 & 5.83 & {\color{colorbrewer2}29.40}\\
		\Clipping[$0.1$]+LS & 4.82 & 87.90 & 78.89 & 6.10 & 10.59\\
		\Clipping[$0.05$]+LS & 5.30 & 87.41 & 85.04 & 6.43 & 7.30\\
		\hline
	\end{tabular}
	\vspace*{-0.2cm}
\end{table}

Our \textbf{appendix} includes implementation details (\appref{sec:supp-implementation}), more information on our experimental setup (\appref{subsec:supp-experiments-setup}), and complementary experiments (\appref{sec:supp-experiments}). Among others, we discuss the robustness of BN (\appref{subsec:supp-experiments-bn}), other architectures such as ResNet-50 (\appref{subsec:supp-experiments-bn}), qualitative results for \Clipping (\appref{subsec:supp-experiments-clipping}) and complete results for $m = 4,3,2$ bits precision (\appref{subsec:supp-experiments-summary}). Also, we discuss a simple guarantee how the average \RTE relates
to the true expected robust error (\appref{subsec:supp-bound}). Our \textbf{code} will be made publicly available.

\subsection{Quantization Choice Impacts Robustness}
\label{subsec:experiments-quantization}

Quantization schemes affect robustness significantly, even when not affecting accuracy. 
\tabref{tab:quantization-robustness} shows that per-layer quantization reduces \RTE significantly for small bit error rates, \eg, $p = 0.05\%$. While asymmetric quantization further reduces the quantization range, \RTE increases, especially for large bit error rates, \eg, $p = 0.5\%$ (marked in {\color{colorbrewer1}red}). This is despite \figref{fig:quantization} showing a slightly smaller impact of bit errors. This is caused by an asymmetric quantization into \emph{signed} integers: Bit flips in the most significant bit (MSB, \ie, sign bit) are not meaningful if the quantized range is not symmetric as the sign bit does not reflect the sign of the represented weight value, see \appref{subsec:supp-experiments-quantization}. Similarly, replacing integer conversion of $\nicefrac{w_i}{\Delta}$ by proper rounding, $\lceil\nicefrac{w_i}{\Delta}\rfloor$, reduces \RTE significantly (resulting in our \Quant).
This becomes particularly important for $m = 4$. Here, rounding also improves clean \TE slightly, but the effect is significantly less pronounced. Proper rounding generally reduces the quantization error. However, it is striking that this has little impact on \TE but tremendous effect on \RTE.
For $m = 4$ or lower, we also found weight clipping to help training, obtaining lower \TE.
Overall, random bit errors induce unique error distributions, \cf \figref{fig:quantization}, heavily dependent on quantization details.

\begin{table}[t]
	\centering
	\caption{\textbf{Fixed Pattern Bit Error Training.} \RTE for training on an entirely fixed bit error pattern (\Pattern). \emph{Top:} Evaluation on the same pattern; \Pattern trained on $p = 2.5\%$ does not generalize to $p = 1\%$ even though the bit errors for $p = 1\%$ are a subset of those seen during training for $p = 2.5\%$ (in {\color{colorbrewer1}red}). \emph{Bottom:} \Pattern also fails to generalize to completely random bit errors. This can be confirmed on profiled bit errors in \appref{subsec:supp-randbet-baselines}.}
	\label{tab:randbet-baselines}
	\vspace*{-0.25cm}
	\small
	\begin{tabular}{| l | c | c |}
		\hline
		Model (\CifarT) & \multicolumn{2}{c|}{\RTE in \%, $p$ in \%}\\
		\hline
		\hline
		\textbf{Evaluation on Fixed Pattern} & $p{=}1$ & $p{=}2.5$\\
		\hline
		\Pattern $p{=}2.5$ & {\color{colorbrewer1}14.14} & 7.87\\
		\Pattern[$0.15$] $p{=}2.5$ & {\color{colorbrewer1}8.50} & 7.41\\
		\hline\hline
		\textbf{Evaluation on \emph{Random} Patterns} & $p{=}1$ & $p{=}2.5$\\
		\hline
		\Pattern[$0.15$] $p{=}2.5$ & 12.09 & 61.59\\
		\hline
	\end{tabular}
	\vspace*{-0.2cm}
\end{table}

\subsection{Weight Clipping Improves Robustness}
\label{subsec:experiments-clipping}
While the quantization range adapts to the weight range
after every update during training, weight clipping explicitly constraints the weights to $[-\wmax, \wmax]$.
\tabref{tab:clipping-robustness} shows the effect of
different $\wmax$ for \CifarT with 8 bit precision. The clean test error is not affected for \Clipping[$\mathbf{\wmax{=}0.15}$] but
one has already strong robustness improvements for $p=1\%$
compared to \Quant (\RTE of 13.18\% vs 32.05\%). Further reducing $\wmax$ leads to a slow increase in clean \TE and decrease in average clean confidence, while significantly
improving \RTE to $7.18\%$ for $p=1\%$ at $\wmax=0.05$. For $\wmax=0.025$
the DNN is no longer able to achieve high confidence (marked in {\color{colorbrewer1}red}) which leads to stronger loss of clean \TE. Interestingly, the gap between clean and perturbed confidences under bit errors for $p=1\%$ is (almost) monotonically decreasing. These findings generalize to other datasets and precisions, see \appref{subsec:supp-experiments-summary}. However, for low precision $m\leq 4$ the effects are stronger 
as \Quant alone does not yield any robust models and weight clipping is essential for achieving robustness. 

As discussed in \secref{subsec:robustness-clipping} the robustness of the DNN originates in the cross-entropy loss enforcing high confidences on the training set and, thus, large logits while weight clipping works against having large logits. Therefore, the network has to utilize more weights with larger absolute values (compared to $\wmax$).
In order to test this hypothesis, we limit the confidences that need to be achieved via label smoothing \cite{SzegedyCVPR2016}, targeting $0.9$ for the true class and $\nicefrac{0.1}{9}$ for the other classes. According to \secref{subsec:robustness-clipping}, this should lead to less robustness, as the DNN has to use ``fewer'' weights. Indeed, in \tabref{tab:clipping-robustness}, \RTE at $p=1\%$ increases from $13.08\%$ for \Clipping[$0.15$] to $29.4\%$ when using label smoothing (marked in {\color{colorbrewer2}blue}). Moreover, the difference between average clean and perturbed confidence is significantly larger for DNNs trained with label smoothing. 

In \appref{subsec:supp-experiments-clipping} we show that robustness against bit errors also leads to robustness against $L_\infty$ perturbations which generally affect all weights in contrast to random bit errors, and provide more qualitative results about the change of the weight distribution induced by clipping in \figref{fig:supp-clipping}.

\begin{table}[t]
	\centering
	\small
	\caption{\textbf{Random Bit Error Training (\Random).} Average \RTE (and standard deviation) of \Random evaluated at various bit error rates $p$ and using $m = 8$ or $4$ bit precision. For low $p$, weight clipping provides sufficient robustness. However for $p \geq 0.5$, \Random increases robustness significantly. This is pronounced for lower precisions.}
	\label{tab:randbet-robustness}
	\vspace*{-0.25cm}
	\hspace*{-0.25cm}
	\begin{tabular}{|@{\hskip 3px}c@{\hskip 3px}|@{\hskip 3px}l@{\hskip 3px}|@{\hskip 3px}c@{\hskip 3px}|@{\hskip 3px}c@{\hskip 3px}|@{\hskip 3px}c@{\hskip 3px}|@{\hskip 3px}c@{\hskip 3px}|}
		\hline
		& Model (\CifarT) & \multirow{2}{*}{\begin{tabular}{@{}c@{}}\TE\\in \%\end{tabular}} &\multicolumn{3}{c|}{\RTE in \%}\\
		\cline{4-6}
		& $\mathbf{\wmax{=}0.1}$, $p$ in \% && $p{=}0.5$ & $p{=}1$ & $p{=}1.5$\\
		\hline
		\hline
		\multirow{5}{*}{\rotatebox{90}{$8$bit}} & \Quant & \bfseries 4.32 & 11.28 {\color{gray}\tiny ${\pm}$1.47} & 32.05 {\color{gray}\tiny ${\pm}$6} & 68.65 {\color{gray}\tiny ${\pm}$9.23}\\
		& \Clipping & 4.82 & 6.95 {\color{gray}\tiny ${\pm}$0.24} & 8.93 {\color{gray}\tiny ${\pm}$0.46} & 12.22 {\color{gray}\tiny ${\pm}$1.29}\\
		& \Random $p{=}1$ & 4.90 & \bfseries 6.36 {\color{gray}\tiny ${\pm}$0.17} & \bfseries 7.41 {\color{gray}\tiny ${\pm}$0.29} & \bfseries 8.65 {\color{gray}\tiny ${\pm}$0.37}\\
		\hline
		\multirow{2}{*}{\rotatebox{90}{$4$bit}} & \Clipping & \bfseries 5.29 & 7.71 {\color{gray}\tiny ${\pm}$0.36} & 10.62 {\color{gray}\tiny ${\pm}$1.08} & 15.79 {\color{gray}\tiny ${\pm}$2.54}\\
		& \Random $p{=}1$ & 5.39 & \bfseries 7.04 {\color{gray}\tiny ${\pm}$0.21} & \bfseries 8.34 {\color{gray}\tiny ${\pm}$0.42} & \bfseries 9.77 {\color{gray}\tiny ${\pm}$0.81}\\
		\hline
	\end{tabular}
	\vspace*{-0.2cm}
\end{table}

\subsection{\Random Yields Generalizable Robustness}
\label{subsec:experiments-randbet}

\textbf{Training on Profiled Errors Does Not Generalize:}
%
Co-design approaches such as \cite{KimDATE2018,KoppulaMICRO2019} combine training DNNs on profiled SRAM or DRAM bit errors with hardware-approaches to limit the errors' impact.
However, profiling SRAM or DRAM requires expensive infrastructure, expert knowledge and time. 
More importantly, training on profiled bit errors does not generalize to previously unseen bit error distributions (\eg, other chips or voltages): \tabref{tab:randbet-baselines} (top) shows \RTE of \Pattern, \ie, pattern-specific bit error training. The main problem is that \Pattern does \emph{not} even generalize to lower bit error rates (\ie, higher voltages) of the same pattern as trained on (marked in {\color{colorbrewer1}red}). This is striking as, following \figref{fig:errors}, the bit errors form a subset of the bit errors seen during training: training with $p = 2.5\%$ bit errors does not provide robustness for $p = 1\%$, \RTE increases $7.9\%$ to $14.1\%$. It is not surprising, that \tabref{tab:randbet-baselines} (bottom) also demonstrates that \Pattern does not generalize to random bit error patterns: \RTE increases from $7.4\%$ to $61.6\%$ at $p = 2.5\%$. The same observations can be made when training on real, profiled bit errors corresponding to the chips in \figref{fig:errors}, see \appref{subsec:supp-randbet-baselines}.
Overall, obtaining robustness that generalizes across voltages \emph{and} chips is crucial for low-voltage operation to become practical.

\begin{table}[t]
	\centering
	\small
	\caption{\textbf{Generalization to Profiled Bit Errors.} \RTE for \Random on two different profiled chips. The bit error rates differ across chips due to measurements at different voltages, also see \figref{fig:errors}. Chip 2 exhibits a bit error distribution significantly different from uniform random bit errors: bit errors are strongly aligned along columns and biased towards $0$-to-$1$ flips, \cf \figref{fig:errors}. Nevertheless, \Random generalizes surprisingly well.}
	\label{tab:randbet-generalization}
	\vspace*{-0.25cm}
	\hspace*{-0.15cm}
	\begin{tabular}{| l | l | c | c |}
		\hline
		Chip (\figref{fig:errors}) & Model (\CifarT)& \multicolumn{2}{c|}{\RTE in \%}\\
		\hline
		\hline
		\bfseries Chip 1 && $p{\approx}0.86$ & {\color{colorbrewer1}$p{\approx}2.75$}\\
		\hline
		& \Random[$0.05$] $p{=}1.5$ & 7.04 & 9.37\\
		\hline
		\hline
		\bfseries Chip 2 && $p{\approx}0.14$ & {\color{colorbrewer1}$p{\approx}1.08$}\\
		\hline
		& \Random[$0.05$] $p{=}1.5$ & 6.00 & 9.00\\
		\hline
	\end{tabular}
	\vspace*{-0.2cm}
\end{table}
\begin{figure*}[t]
	\centering
	\vspace*{-0.1cm}
	\hspace*{-0.3cm}
	\begin{subfigure}{0.32\textwidth}
		\centering
		\includegraphics[width=5.75cm]{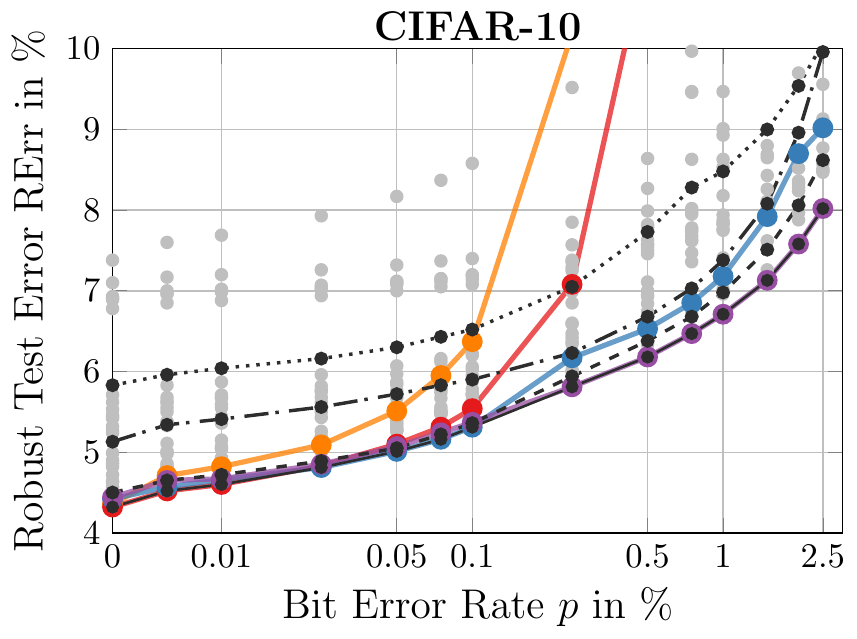}
	\end{subfigure}
	\hfill
	\begin{subfigure}{0.32\textwidth}
		\centering
		\includegraphics[width=5.75cm]{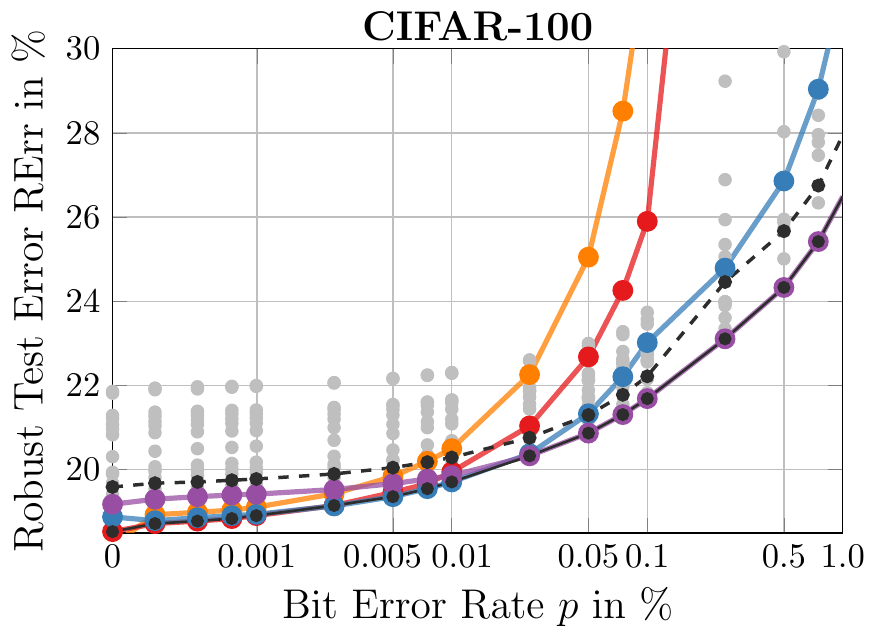}
	\end{subfigure}
	\hfill
	\begin{subfigure}{0.32\textwidth}
		\centering
		\includegraphics[width=5.75cm]{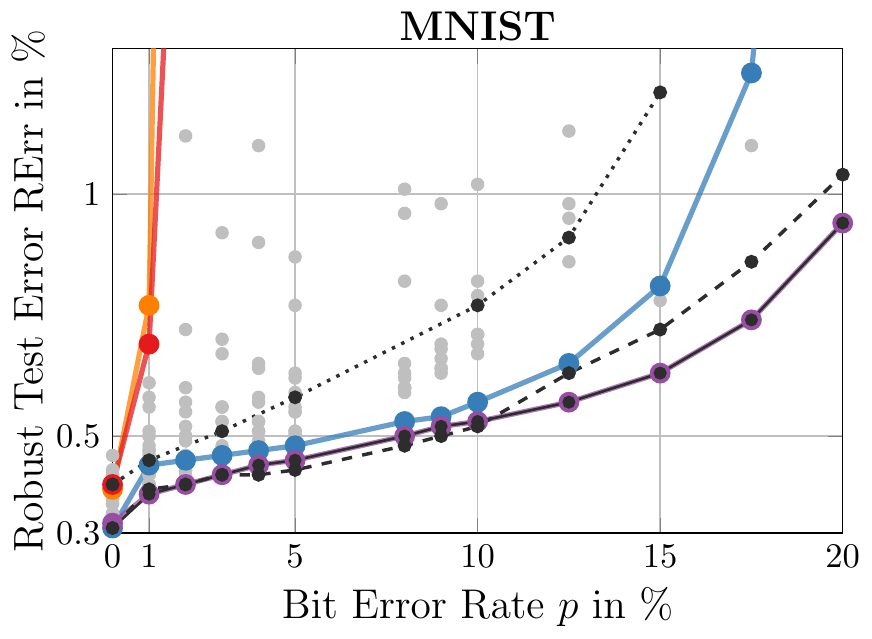}
	\end{subfigure}
	
	\hspace*{-0.1cm}
	\fbox{
	\begin{subfigure}{0.98\textwidth}
		\centering
		\includegraphics[width=1\textwidth]{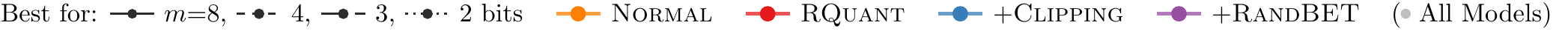}
	\end{subfigure}
	}
	\vspace*{-14px}
	\caption{\textbf{Bit Error Robustness on \CifarT, \CifarH and \MNIST.} Average \RTE plotted against bit error rate $p$, both in \%. We considered various models (in {\color{gray}$\bullet$ gray}), corresponding to different $\wmax$ and $p$ during training. We explicitly plot the best model for each bit error rate: for \Normal ({\color{colorbrewer5}orange}), \Quant ({\color{colorbrewer1}red}), \Clipping ({\color{colorbrewer2}blue}) and \Random ({\color{colorbrewer4}violet}). Note that these might correspond to different $\wmax$ and $p$ (also across datasets). Across all approaches, we plot the per-error-rate best model in black: for $m = 8,4,3,2$ bits, depending on dataset. For $8$ bit and low bit error rates, \Clipping is often sufficient. However, for $4$ bit or higher bit error rates, \Random is crucial to keep \RTE low.}
	\label{fig:summary}
	\vspace*{-0.2cm}
\end{figure*}

\textbf{\Random Improves Robustness:}
%
\Random, with weight clipping, further improves robustness and additionally generalizes across chips and voltages. \tabref{tab:randbet-robustness} shows results for weight clipping and \Random with $\wmax = 0.1$ and $m = 8,4$ bits precision. \Random is particularly effective against large bit error rates, \eg, $p = 1.5\%$, reducing \RTE from $12.22\%$ to $8.65\%$ ($m = 8$ bits). The effect is pronounced for $4$ bits or even lower precision, where models are generally less robust. The optimal combination of weight clipping and \Random depends on the bit error rate. For example, in \tabref{tab:clipping-robustness}, lowering $\wmax$ to $0.05$ reduces \RTE below \Random[$0.1$] with $p{=}1\%$ for some bit error rates.
We emphasize that \Random generalizes to lower bit errors than trained on, in stark contrast to the fixed-pattern training \Pattern. In \appref{subsec:supp-experiments-architectures}, we also show that \Random works on other architectures such as ResNet-50. On other datasets, \eg,\MNIST, \Random allows to operate at $p = 12.5\%$ bit error rate with $0.9\%$ \RTE and only $m = 2$ bits. At this point, weight clipping alone yields $90\%$ \RTE.

\textbf{\Random Generalizes to Profiled Bit Errors:}
%
\Random also generalizes to profiled bit errors from real chips. \tabref{tab:randbet-generalization} shows results on the two profiled chips of \figref{fig:errors}. Profiling was done at various voltage levels, resulting in different bit error rates. To simulate various weights to memory mappings, we apply various offsets before linearly mapping weights to the profiled SRAM arrays. \tabref{tab:randbet-generalization} reports average \RTE, showing that \Random generalizes quite well to these profiled bit errors. Regarding chip 1, \Random performs very well, even for large $p\approx 2.75$, as the bit error distribution of chip 1 largely matches our error model in \secref{sec:errors}, \cf \figref{fig:errors} (left). In contrast, with chip 2 we picked a more difficult bit error distribution which is strongly aligned along columns, potentially hitting many MSBs simultaneously. Thus, \RTE increases for chip 2 even for a lower bit error rate $p \approx 1.08$ (marked in {\color{colorbrewer1}red})
but energy savings are still possible without degrading prediction performance. 

\subsection{Summary and Discussion}
\label{subsec:experiments-discussion}
 
Our experiments are summarized in \figref{fig:summary}. We consider \Normal quantization vs. our robust quantization \Quant, various \Clipping and \Random models with different $\wmax$ and $p$ during training (indicated in {\color{gray}$\bullet$ gray}) and plot \RTE against bit error rate $p$ at test time. On all datasets \Quant outperforms \Normal. On \CifarT (left), \RTE increases significantly for \Quant ({\color{colorbrewer1}red}) starting at $p \approx 0.25\%$ bit error rate. While \Clipping ({\color{colorbrewer2}blue}) generally reduces \RTE, only \Random ({\color{colorbrewer4}violet}) can keep \RTE around $6\%$ or lower for a bit error rate of $p \approx 0.5\%$. The best model for each bit error rate $p$ (black and solid for $m = 8$) might vary. \CifarH is generally more difficult, while significantly higher bit error rates are possible on \MNIST. On \CifarT, \RTE increases slightly for $m = 4$. However, for $m = 3,2$ \RTE increases more significantly as clean \TE increases by $1-2\%$. 
Nevertheless, \RTE only increases slightly for larger bit error rates $p$. It remains future work whether \Random with a more sophisticated (but robust) quantization scheme can enable low-voltage operation even for $m = 2$ bits. In all cases, \RTE increases monotonically, ensuring safe operation at higher voltages. The best trade-off between robustness and accuracy depends on the application: higher energy savings require a larger ``sacrifice'' in terms of \RTE.
Finally, \appref{subsec:supp-bound} provides a
confidence-interval based guarantee on how strongly \RTE is expected to deviate from the empirical results in \figref{fig:summary}.

Overall, the results in \figref{fig:summary} enable robust low-voltage operation \emph{without} requiring expensive error correcting codes (ECCs) or other circuit techniques \cite{ReagenISCA2016,ChandramoorthyHPCA2019}. Furthermore, our analysis applies both to DRAM, commonly off-chip, and SRAM, usually used as scratchpads on-chip of DNN accelerators. Compared to co-design \cite{KimDATE2018,KoppulaMICRO2019}, we do not require expensive expert knowledge or profiling infrastructure. Moreover, \Random improves over these approaches by generalizing across chips and voltages.
Besides \Random, we show that robust fixed-point quantization \emph{only with} weight clipping can provide reasonable robustness, \eg, for $p = 0.1\%$ on \CifarT. This is without sophisticated quantization scheme, \eg, with special treatment for outliers \cite{ZhuangCVPR2018,SungARXIV2015,ParkISCA2018}, and complementary to  \cite{MurthyARXIV2019,MerollaARXIV2016,SungARXIV2015,AlizadehICLR2020}, focusing merely on robustness \emph{to} quantization.

\section{Conclusion}
\label{sec:conclusion}

We propose a combination of \textbf{robust quantization}, \textbf{weight clipping} and \textbf{random bit error training (\Random)} to get DNNs robustness against random bit errors in their (quantized) weights, enabling low-voltage operation of DNN accelerators to save energy. Here, the accelerator memory is operated far below its rated voltage \cite{ChandramoorthyHPCA2019,KoppulaMICRO2019,KimDATE2018}, inducing exponentially increasing rates of bit errors, directly affecting stored DNN weights. Weight clipping regularizes the weights to a small $[-\wmax, \wmax]$ during training, encouraging redundancy and increasing robustness. \Random further generalizes across chips, with different bit error patterns, and voltages without requiring expensive memory profiling or hardware mitigation strategies. These are important criteria for low-voltage operation in practice. Besides, we also discuss the impact of fixed-point quantization schemes on robustness, which has been neglected in prior work. We are able to train low-precision DNNs robust to significant rates of random bit errors which allow a reduction in energy consumption of roughly $20\%$ or more on \MNIST and \Cifar.
 
\section*{Acknowledgements}
MH acknowledges support from the German Federal Ministry of Education and Research
(BMBF) through the Tübingen AI Center (FKZ: 01IS18039A) and from the Deutsche Forschungsgemeinschaft
(DFG, German Research Foundation) under Germany’s Excellence Strategy (EXC number
2064/1, project number 390727645). 
NC acknowledges that this research was developed in part with funding from the U.S. Defense Advanced Research Projects Agency (DARPA). The views, opinions and/or other findings expressed are those of the authors and should not be interpreted as representing the official views or policies of the Department of Defense or the U.S. Government. DISTRIBUTION STATEMENT A. Approved for public release: distribution unlimited.

\bibliography{bibliography}
\bibliographystyle{mlsys2021}

\renewcommand{\topfraction}{0.9}
\renewcommand{\bottomfraction}{0.9}
\setcounter{topnumber}{2}
\setcounter{bottomnumber}{2}
\setcounter{totalnumber}{4}
\setcounter{dbltopnumber}{2}
\renewcommand{\dbltopfraction}{0.9}
\renewcommand{\textfraction}{0.05}
\renewcommand{\floatpagefraction}{0.8}
\renewcommand{\dblfloatpagefraction}{0.8}

\vfill
\pagebreak
\begin{appendix}
	\twocolumn[{\begin{figure}[H]
	\setlength{\linewidth}{\textwidth}
	\setlength{\hsize}{\textwidth}
	\centering
	\centering
	\begin{tikzpicture}
		\node[anchor=north west] at (0,0){\includegraphics[width=4cm]{errors_18_2}};
		\node[anchor=north west] at (4,0){\includegraphics[width=4cm]{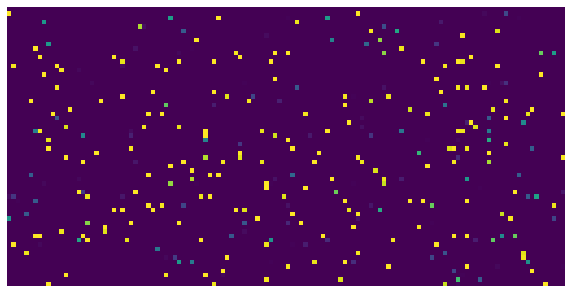}};
		\node[anchor=north west] at (8,0){\includegraphics[width=4cm]{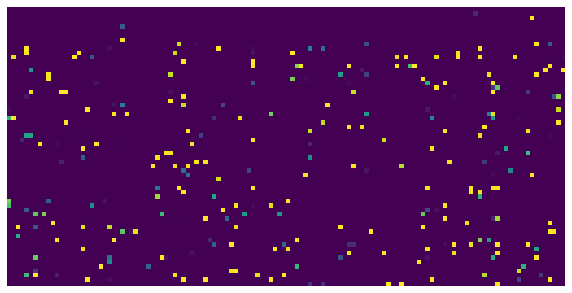}};
		\node[anchor=north west] at (12,0){\includegraphics[width=4cm]{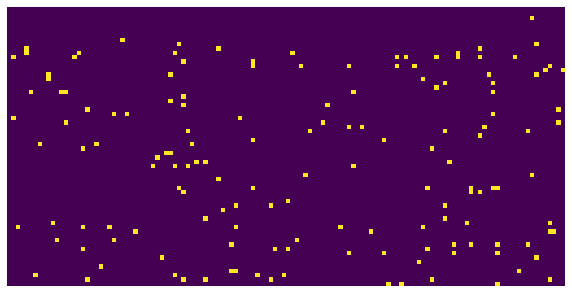}};
		
		\node[anchor=north west] at (0,-2.5){\includegraphics[width=4cm]{errors_n_2}};
		\node[anchor=north west] at (4,-2.5){\includegraphics[width=4cm]{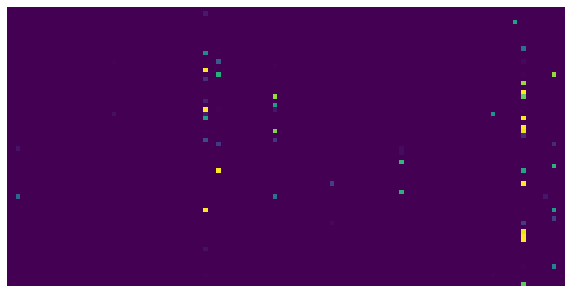}};
		\node[anchor=north west] at (8,-2.5){\includegraphics[width=4cm]{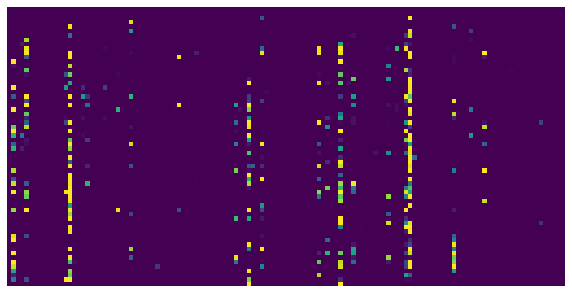}};
		\node[anchor=north west] at (12,-2.5){\includegraphics[width=4cm]{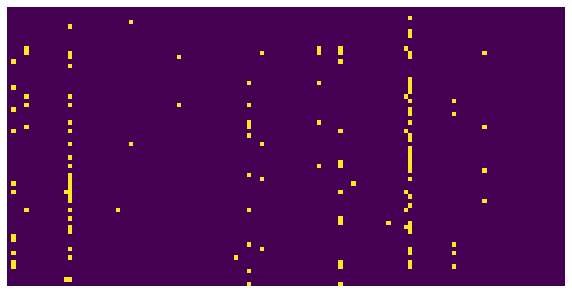}};
		
		\node[anchor=south east,fill opacity=0.75,fill=white] at (4, -2){$p{\approx}2.75\%$};
		\node[anchor=south east,fill opacity=0.75,fill=white] at (4, -4.5){$p{\approx}1.08\%$};
		
		\draw[white!30!black,-, line width=0.75pt] (12.125,-5.1) -- (12.125,0.4);
		\draw[thick,->] (0.05,0.05) -- (1,0.05);
		\draw[thick,->] (-0.05,-0.05) -- (-0.05,-1);
		\node[anchor=south] at (0.75,0.05){128 columns};
		\node[rotate=90,anchor=south] at (-0.05,-0.6){64 rows};
		
		\node[anchor=south] at (6,-0.2) {\bfseries Chip 1};
		\node[anchor=north] at (6,-2.2) {\bfseries Chip 2};
		
		\node at (2, -5){Overall bit flips};
		\node at (4, -5){=};
		\node at (6, -5){$1$-to-$0$ flips};
		\node at (8, -5){+};
		\node at (10, -5){$0$-to-$1$ flips};
		\node at (14, -5){persistent errors};
	\end{tikzpicture}
	\vspace*{-6px}
	\caption{\textbf{Low-Voltage Induced Bit Errors on Profiled Chips.} Complementary to \figref{fig:errors}, we break the the bit error distribution of chips 1 and 2 down into $1$-to-$0$ and $0$-to-$1$ bit flips. Additionally, most of the bit errors are actually persistent across accesses at a given supply voltage. As before, we show a sub-array of size $64 \times 128$ from all profiled bit cells (\ie, across all SRAM arrays). \secref{sec:supp-introduction} includes details on profiling.}
	\label{fig:supp-errors}
	\vspace*{-0.1cm}
\end{figure}}]

\section{Energy Savings in \figref{fig:introduction}}
\label{sec:supp-introduction}

\figref{fig:introduction} shows bit error rate characterization results of SRAMs in the DNN accelerator chip described in \cite{ChandramoorthyHPCA2019}, fabricated using 14nm FinFET technology. The average bit error rate is measured from 32 SRAMs, each SRAM array of size 4KB (512 $\times$ 64 bit), as supply voltage is scaled down. \textit{Bit error rate} $p$ (in \%) at a given supply voltage is measured as the count of read or write bit cell failures averaged over the total number of bit cells in the SRAM. A bit cell failure refers to reading 1 on writing 0 or reading 0 on writing 1.  For a more comprehensive characterization of SRAMs in 14nm technology, the reader is referred to \cite{GanapathyDAC2017}. \figref{fig:introduction} also shows the energy per write and read access of a 4KB (512 $\times$ 64 bit) SRAM, obtained from Cadence Spectre simulations. Energy is obtained at the same constant clock frequency at all supply voltages. The voltage (x-axis) shown is normalized over \Vmin which is the lowest measured voltage at which there are no bit cell failures.  Energy shown in the graph (secondary axis on the right) is also normalized over the energy per access at \Vmin.

Accelerators such as \cite{ChenISCA2016,ChenASPLOS2014,ChandramoorthyHPCA2019, ReagenISCA2016, nvdla, DuISCA2015,SharmaISCA2018} have a large amount of on-chip SRAM to store weights and intermediate computations. Total dynamic energy of accelerator SRAMs can be obtained as the total number of SRAM accesses times the energy of a single SRAM access. Optimized dataflow in accelerators leads to better re-use of weights read from memories in computation, reducing the number of such memory accesses ~\cite{ChenISCA2016,ChenASPLOS2014, nvdla}. Low voltage operation focuses on reducing the memory access energy, leading to significant energy savings as shown. 
	\section{Related Work}
\label{sec:supp-related-work}

\begin{table*}[t]
    \centering
   	\caption{\textbf{Architectures, Number of Weights $\mathbf{W}$, Expected Number of Bit Errors.} \textit{Left and Middle:} SimpleNet architectures used for \MNIST and \CifarT with the corresponding output sizes, channels $N_C$, height $N_H$ and width $N_W$, and the total number of weights $W$. We use group normalization \emph{with} learnable scale/bias, but reparameterized as outlined in \appref{sec:supp-clipping}. \textit{Right:} The number of expected bit errors for random bit errors, \ie, $pmW$.}
    \label{tab:supp-architectures}
    \vspace*{-0.25cm}
    \begin{subfigure}[t]{.30\textwidth}
        \vspace*{0px}
    
        \footnotesize
\begin{tabular}{|l|c|}
    \hline
    \multicolumn{2}{|c|}{\textbf{SimpleNet} on \textbf{MNIST}}\\
    \hline
    Layer & Output Size\\
    & $N_C, N_H, N_W$\\
    \hline
    \hline
    Conv+GN+ReLU & $32, 28, 28$\\
    Conv+GN+ReLU & $64, 28, 28$\\
    Conv+GN+ReLU & $64, 28, 28$\\
    Conv+GN+ReLU & $64, 28, 28$\\
    Pool & $64, 14, 14$\\
    Conv+GN+ReLU & $64, 14, 14$\\
    Conv+GN+ReLU & $64, 14, 14$\\
    Conv+GN+ReLU & $128, 14, 14$\\
    Pool & $128, 7, 7$\\
    Conv+GN+ReLU & $256, 7, 7$\\
    Conv+GN+ReLU & $1024, 7, 7$\\
    Conv+GN+ReLU & $128, 7, 7$\\
    Pool & $128, 3, 3$\\
    Conv+GN+ReLU & $128, 3, 3$\\
    Avg Pool & $128, 1, 1$\\
    FC & $10$\\
    \hline
    \hline
    $W$ & 1,082,826\\
    \hline
\end{tabular}
    \end{subfigure}
    \begin{subfigure}[t]{.30\textwidth}
        \vspace*{0px} 
        
        \footnotesize
\begin{tabular}{|l|c|}
    \hline
    \multicolumn{2}{|c|}{\textbf{SimpleNet} on \textbf{\CifarT}}\\
    \hline
    Layer & Output Size\\
    & $N_C, N_H, N_W$\\
    \hline
    \hline
    Conv+GN+ReLU & $64, 32, 32$\\
    Conv+GN+ReLU & $128, 32, 32$\\
    Conv+GN+ReLU & $128, 32, 32$\\
    Conv+GN+ReLU & $128, 32, 32$\\
    Pool & $128, 16, 16$\\
    Conv+GN+ReLU & $128, 16, 16$\\
    Conv+GN+ReLU & $128, 16, 16$\\
    Conv+GN+ReLU & $256, 16, 16$\\
    Pool & $256, 8, 8$\\
    Conv+GN+ReLU & $256, 8, 8$\\
    Conv+GN+ReLU & $256, 8, 8$\\
    Pool & $256, 4, 4$\\
    Conv+GN+ReLU & $512, 4, 4$\\
    Pool & $512, 2, 2$\\
    Conv+GN+ReLU & $2048, 2, 2$\\
    Conv+GN+ReLU & $256, 2, 2$\\
    Pool & $256, 1, 1$\\
    Conv+GN+ReLU & $256, 1, 1$\\
    Avg Pool & $256, 1, 1$\\
    FC & $10$\\
    \hline
    \hline
    $W$ & 5,498,378\\
    \hline
\end{tabular}
    \end{subfigure}
    \begin{subfigure}[t]{.20\textwidth}
        \vspace*{0px}
        
        \footnotesize
\begin{tabular}{|l|c|}
    \hline
    \multicolumn{2}{|c|}{$\mathbf{p}$ on \textbf{MNIST}}\\
    \hline
    $p$ in \% & $pmW$, $m = 8$\\
    \hline
    \hline
    \multicolumn{2}{|c|}{\textit{Random} Bit Errors}\\
    \hline
    $10$ & 866260\\
    $5$ & 433130\\
    $1.5$ & 129939\\
    $1$ & 86626\\
    $0.5$ & 43313\\
    \hline
\end{tabular}
        
        \footnotesize
\begin{tabular}{|l|c|}
    \hline
    \multicolumn{2}{|c|}{$\mathbf{p}$ on \textbf{\Cifar}}\\
    \hline
    $p$ in \% & $pmW$, $m = 8$\\
    \hline
    \hline
    \multicolumn{2}{|c|}{\textit{Random} Bit Errors}\\
    \hline
    $1$ & 439870\\
    $0.5$ & 219935\\
    $0.01$ & 43987\\
    \hline
\end{tabular}
    \end{subfigure}
    \vspace*{-0.1cm}
\end{table*}

In the following, we briefly review work on adversarial robustness, fault tolerance, backdooring and quantization. These areas are broadly related to the topic of the main paper.

\textbf{Adversarial and Corruption Robustness:} Robustness of DNNs against adversarially perturbed or randomly corrupted inputs received considerable attention in recent years, see, \eg, relevant surveys \citep{BiggioARXIV2018,XuARXIV2019}. Adversarial examples \citep{SzegedyARXIV2013}, \ie, nearly imperceptibly perturbed inputs causing misclassification, consider an adversarial environment where potential attackers can actively manipulate inputs. This has been shown to be possible in the white-box setting, with full access to the DNN, \eg, \citep{MadryICLR2018,CarliniSP2017,DongARXIV2017,ChiangARXIV2019,CroceARXIV2020}, as well as in the black-box setting, without access to DNN weights and gradients, \eg, \citep{ChenAISEC2017,IlyasICML2018,CroceARXIV2019,AndriushchenkoARXIV2019}. Such attacks are also transferable between models \citep{LiuARXIV2016} and can be applied in the physical world \citep{LuARXIV2017,KurakinARXIV2016b}. Obtaining robustness against adversarial inputs is challenging, recent work focuses on achieving certified/provable robustness \citep{PeckNIPS2017,ZhangNIPS2018,WongICML2018,GorwalARXIV2019} and variants of adversarial training \citep{MiyatoARXIV2015,HuangARXIV2015,MadryICLR2018}, \ie, training on adversarial inputs generated on-the-fly. Adversarial training has been shown to work well empirically, and flaws such as reduced accuracy \citep{StutzCVPR2019,TsiprasICLR2019} or generalization to attacks not seen during training has been addressed repeatedly \citep{CarmonARXIV2019,UesatoARXIV2019,StutzICML2020,TramerARXIV2019,MainiARXIV2019}. Adversarial inputs have also been considered for quantized DNNs \citep{KhalilICLR2019}. Corrupted inputs, in contrast, consider ``naturally'' occurring corruptions to which robustness/invariance is desirable for practical applications. Popular benchmarks such as MNIST-C \citep{MuICMLWORK2019}, Cifar10-C or ImageNet-C \citep{HendrycksARXIV2019} promote research on corruption robustness by extending standard datasets with common corruptions, \eg, blur, noise, saturation changes \etc. It is argued that adversarial robustness, and robustness to random corruptions is related. Approaches are often similar, \eg, based on adversarial training \citep{StutzICML2020,LopesICMLWORK2019,KangARXIV2019}. In contrast, we consider random bit errors in the weights, not the inputs.

\textbf{Fault Tolerance:} Fault tolerance, describes structural changes such as removed units, and has been studied in early works such as \citep{AlippiISCAS1994,NetiTNN1992,Chiu1994}. These approaches obtain fault tolerant NNs using approaches similar to adversarial training \citep{DeodhareTNN1998,LeeICASSP2014}.
\revision{Recently, hardware mitigation strategies \cite{MarquesSIPS2017}, weight regularization \citep{RahmanICIP2018,DeyTSMCS2018,LeungTNN2010}, fault detection \cite{XiaDAC2017} or GAN-based training \citep{DudduARXIV2019} has been explored. Generally, a wide range of different faults/errors are considered, including node faults \cite{LeeICASSP2014,DeodhareTNN1998}, hardware soft errors \cite{AziziMazreahNAS2018}, timing errors \cite{DengDATE2015} or transient errors in general \cite{SalamiSBACPAD2018}. However, to the best of our knowledge, large rates of non-transient bit errors provoked through low-voltage operation has not been considered. Nevertheless, some of these approaches are related to ours in spirit: \cite{DuASPDAC2014} consider inexact computation for energy-efficiency and \cite{CavalieriOJINN1999,KlachkoIJCNN2019,HoangDATE2020} constrain weights and/or activations to limit the impact of various errors -- similar to our weight clipping.}
Additionally, fault tolerance of adversarially robust models has been considered in \citep{DudduARXIV2019b}. We refer to \citep{TorreshuitzilIEEEACCESS2017} for a comprehensive survey. In contrast, we do \emph{not} consider structural changes/errors in DNNs.

\textbf{Backdooring:} The goal of backdooring is to introduce a backdoor into a DNN, allowing to control the classification result by fixed input perturbations at test time. This is usually achieved through data poisoning  \citep{LiuNDSS2018,LiaoARXIV2018,ZhangASIACCS2018}. However, some works also consider directly manipulating the weights \citep{JiCCS2018,DumfordARXIV2018}. However, such weight perturbations are explicitly constructed not to affect accuracy on test examples without backdoor. In contrast, we consider random bit errors (\ie, weight perturbations) that degrade accuracy significantly.
	    \section{Low-Voltage Induced Random Bit Errors in Quantized DNN Weights}
\label{sec:supp-main}

We provide a more detailed discussion of the considered error model: random bit errors, induced through low-voltage operation of memories commonly used on DNN accelerators \cite{KimDATE2018,KoppulaMICRO2019}. Work such as \cite{ChandramoorthyHPCA2019,KoppulaMICRO2019} model the effect of low-voltage induced bit errors using two parameters: the probability $\pfault$ of bit cells in accelerator memory being faulty at a given low voltage and the probability $\perror$ that a faulty bit cell results in a bit error on access.
Following measurements in works such as \cite{GanapathyHPCA2019,KimDATE2018}, we assume that these errors are \emph{not} transient errors by setting $\perror = 100\%$ such that the overall probability of bit errors is $p := \pfault \cdot \perror = \pfault$. In doing so, we consider the worst-case where faulty bit cells \emph{always} induce bit errors. However, the noise model from the main paper remains valid for any arbitrary but fixed $\perror \neq 100\%$. For the reminder of this document, we assume the probability of bit error $p = \pfault$, with $\perror = 100\%$, as in the main paper. In the following section, we describe the two parameters, $\pfault$ and $\perror$, in more details.

\begin{table}[t]
	\centering
	\caption{\textbf{Quantization-Aware Training Accuracies.} Clean \TE for $m = 8$ bits or lower using our robust fixed-point quantization. We obtain competitive performance for $m = 8$ and $m = 4$ bits. On \CifarH, a Wide ResNet (WRN) clearly outperforms our standard SimpleNet model. Batch normalization (BN), improving \TE slightly on \CifarT, is significantly less robust than group normalization (GN), \cf \tabref{tab:supp-bn}. * For $m \leq 4$, we report results with weight clipping, \Clipping[$0.1$].}
	\label{tab:supp-accuracy}
	\vspace*{-0.25cm}
	\hspace*{-0.25cm}
	\begin{subfigure}[t]{0.19\textwidth}
		\vspace*{0px}
		\small
		\begin{tabular}{|@{\hskip 4px}l@{\hskip 4px}|@{\hskip 4px}c@{\hskip 4px}|}
			\multicolumn{2}{c}{\bfseries \CifarT}\\
			\multicolumn{2}{c}{\bfseries SimpleNet+GN}\\
			\hline
			Quant. $m$ & \TE in \%\\
			\hline
			-- & 4.34\\
			8 & \bfseries 4.32\\
			4* & 5.29\\
			3* & 5.71\\
			\hline
		\end{tabular}
	\end{subfigure}
	\begin{subfigure}[t]{0.28\textwidth}
		\vspace*{0px}
		\small
		\begin{tabular}{|@{\hskip 4px}l@{\hskip 4px}|@{\hskip 4px}c@{\hskip 4px}|@{\hskip 4px}c@{\hskip 4px}|}
			\multicolumn{3}{c}{\bfseries \CifarT}\\
			\multicolumn{3}{c}{\bfseries Arch. Comparison}\\
			\hline
			Model & no Quant. & $m = 8$\\
			\hline
			SimpleNet+GN & 4.34 & 4.32\\
			SimpleBet+BN & 4.04 & 3.83\\
			ResNet-50+GN & 5.88 & 6.81\\
			ResNet-50+BN & \bfseries 3.91 & \bfseries 3.67\\
			\hline
		\end{tabular}
	\end{subfigure}\\[4px]
	
	\hspace*{-0.5cm} 
	\begin{subfigure}[t]{0.19\textwidth}
		\vspace*{0px}
		\small
		\begin{tabular}{| l | c |}
			\multicolumn{2}{c}{\bfseries \MNIST}\\
			\hline
			Quant. $m$ & \TE in \%\\
			\hline
			4 & 0.4\\
			2* & 0.47\\
			\hline
		\end{tabular}
	\end{subfigure}
	\begin{subfigure}[t]{0.2\textwidth}
		\vspace*{0px}
		\small
		\begin{tabular}{| l | c |}
			\multicolumn{2}{c}{\bfseries \CifarH}\\
			\hline
			Quant. $m$, Model & \TE in \%\\
			\hline
			8, SimpleNet & 23.68\\
			8, WRN & 18.53\\
			\hline
		\end{tabular}
	\end{subfigure}
	\vspace*{-0.1cm}
\end{table}

\textbf{Faulty Bit Cells.} Due to variations in the fabrication process, SRAM bit cells become more or less vulnerable to low-voltage operation. For a specific voltage, the resulting bit cell failures can be assumed to be random and independent of each other. We assume a bit to be faulty with probability $\pfault$ increasing exponentially with decreased voltage \cite{GanapathyDAC2017,GanapathyHPCA2019,KimDATE2018,ChandramoorthyHPCA2019}. Furthermore, the faulty bits for $\pfault' \leq \pfault$ can be assumed to be a subset of those for $\pfault$. For a fixed chip, consisting of multiple memory arrays, the pattern (spatial distribution) of faulty cells is fixed for a specific supply voltage. Across chips/memory arrays, however, faulty cells are assumed to be random and independent of each other.

\textbf{Bit Errors in Faulty Bit Cells:} Faulty cells may cause bit errors with probability $\perror$ upon read/write access.
We note that bit errors read from memory affect \emph{all} computations performed on the read weight value. 
We assume that a bit error flips the currently stored bit, where flips $0$-to-$1$ and $1$-to-$0$ are assumed equally likely.

\subsection{Profiled Bit Errors}
\label{subsec:supp-errors-profiled}

\begin{figure}[t]
	\centering
	\includegraphics[width=0.4\textwidth]{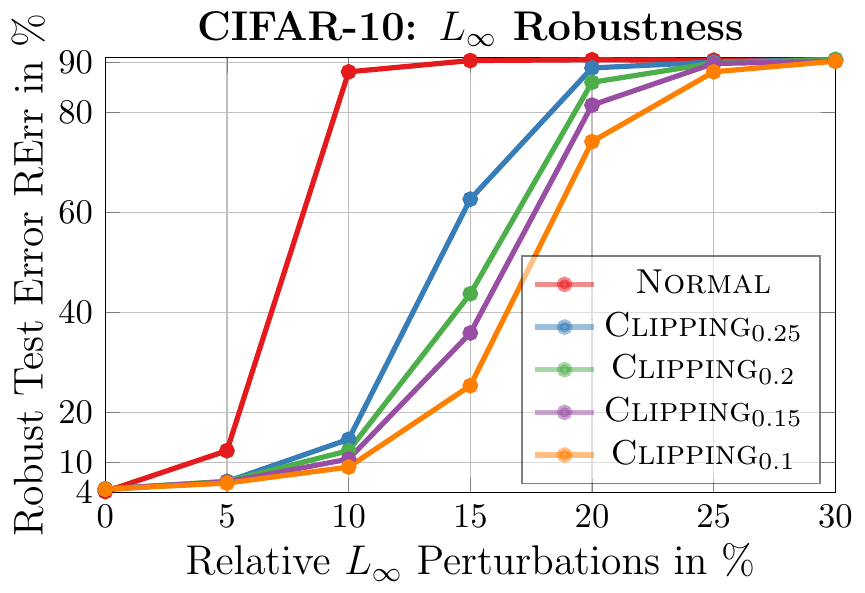}
	\vspace*{-8px}
	\caption{\textbf{Weight Clipping Improves $L_\infty$ Robustness.} On \CifarT, we plot \RTE for \emph{relative} $L_\infty$ perturbations on weights: Random noise with $L_\infty$-norm smaller than or equal to $x\%$ of the weight range is applied. \Clipping clearly improves robustness. Again, the relative magnitude of noise is not affected by weight clipping. Note that $L_\infty$ noise usually affects all weights, while random bit errors affect only a portion of the weights.}
	\label{fig:supp-clipping-inf}
	\vspace*{-0.2cm}
\end{figure}

\figref{fig:supp-errors} splits the bit error distributions of \figref{fig:errors} into a $0$-to-$1$ flip and a $1$-to-$0$ bit flip map. The obtained maps, $p_{\text{1t0}}$ and $p_{\text{0t1}}$, contain per-bit flip probabilities for $1$-to-$0$ and $0$-to-$1$ bit flips at a given low voltage. In this particular profiled chip, \figref{fig:supp-errors} (bottom), $0$-to-$1$ flips are more likely. Similarly, \figref{fig:supp-errors} (right) shows that most  $0$-to-$1$ flips are actually persistent across time at that voltage \ie, not random transient errors. 
The following table summarizing the key statistics of the profiled chips: the overall bit error rate $p$, the rate of $1$-to-$0$ and $0$-to-$1$ flips $p_{\text{1t0}}$ and $p_{\text{0t1}}$, and the rate of persistent errors $p_{\text{sa}}$, all in \% at a specific supply voltage:

\begin{center}\small
\begin{tabular}{| l | c | c | c | c |}
	\hline
	Chip & $p$ & $p_{\text{0t1}}$ & $p_{\text{1t0}}$ & $p_{\text{sa}}$ \\
	\hline
	\multirow{2}{*}{1} & 2.744 & 1.27 & 1.47 & 1.223\\
	& 0.866 & 0.38 & 0.49 & 0.393\\
	\hline
	\multirow{3}{*}{2} & 4.707 &  3.443 & 1.091 & 0.627\\
	& 1.01 &  0.82 & 0.19 & 0.105\\
	& 0.136 & 0.115 & 0.021 & 0.01 \\
	\hline
	\multirow{2}{*}{3} & 2.297 & 1.81 & 0.48 & 0.204 \\
	& 0.597 & 0.496 & 0.0995 & 0.206 \\
	\hline
\end{tabular}
\end{center}

\begin{table*}[t]
	\centering
	\small
	\caption{\textbf{Impact of Quantization Scheme on Robustness.} Complementary to \tabref{tab:quantization-robustness}, we report \TE and \RTE for various bit error rates $p$ for the quantization scheme in \eqnref{eq:quantization} with global, per-layer and asymmetric quantization, $m = 8$ bits. Instead of quantizing into signed integer, using unsigned integers works better for asymmetric quantization. Furthermore, proper rounding instead of integer conversion also improves robustness. Note that influence on clean \TE is neglegible, \ie, the DNN can ``learn around'' these difference in quantization-aware training. Especially for $m = 4$ bit, the latter makes a significant difference in terms of robustness.}
	\label{tab:supp-quantization}
	\vspace*{-0.25cm}
	\begin{tabular}{| c | l | c | c | c | c | c | c | c |}
		\hline
		\multicolumn{9}{|c|}{\bfseries \CifarT: quantization robustness}\\
		\hline
		& Model & \multirow{2}{*}{\begin{tabular}{c}\TE\\in \%\end{tabular}} & \multicolumn{6}{c|}{\RTE in \%, $p$ in \% p=0.01}\\
		\cline{4-9}
		& (see text) && $0.01$ & $0.05$ & $0.1$ & $0.5$ & $1$ & $1.5$\\
		\hline
		\hline
		\multirow{5}{*}{\rotatebox{90}{$m = 8$ bit}} & \eqnref{eq:quantization}, global & 4.63 & 10.70 {\color{gray}\scriptsize ${\pm}$1.37} & 86.01 {\color{gray}\scriptsize ${\pm}$3.65} & 90.36 {\color{gray}\scriptsize ${\pm}$0.66} & 90.71 {\color{gray}\scriptsize ${\pm}$0.49} & 90.57 {\color{gray}\scriptsize ${\pm}$0.43} & --\\
		& \eqnref{eq:quantization}, per-layer (= \Normal) & 4.36 & 4.82 {\color{gray}\scriptsize ${\pm}$0.07} & 5.51 {\color{gray}\scriptsize ${\pm}$0.19} & 6.37 {\color{gray}\scriptsize ${\pm}$0.32} & 24.76 {\color{gray}\scriptsize ${\pm}$4.71} & 72.65 {\color{gray}\scriptsize ${\pm}$6.35} & 87.40 {\color{gray}\scriptsize ${\pm}$2.47}\\
		& +asymmetric & 4.36 & 5.76 {\color{gray}\scriptsize ${\pm}$0.09} & 6.47 {\color{gray}\scriptsize ${\pm}$0.22} & 7.85 {\color{gray}\scriptsize ${\pm}$0.46} & 40.78 {\color{gray}\scriptsize ${\pm}$7.56} & 76.72 {\color{gray}\scriptsize ${\pm}$7.01} & 85.83 {\color{gray}\scriptsize ${\pm}$2.58}\\
		& +unsigned & 4.42 & 6.58 {\color{gray}\scriptsize ${\pm}$0.13} & 6.97 {\color{gray}\scriptsize ${\pm}$0.28} & 7.49 {\color{gray}\scriptsize ${\pm}$0.41} & 17.00 {\color{gray}\scriptsize ${\pm}$2.77} & 54.57 {\color{gray}\scriptsize ${\pm}$8.58} & 83.18 {\color{gray}\scriptsize ${\pm}$3.94}\\
		& +rounded (= \Quant) & 4.32 & 4.60 {\color{gray}\scriptsize ${\pm}$0.08} & 5.10 {\color{gray}\scriptsize ${\pm}$0.13} & 5.54 {\color{gray}\scriptsize ${\pm}$0.2} & 11.28 {\color{gray}\scriptsize ${\pm}$1.47} & 32.05 {\color{gray}\scriptsize ${\pm}$6} & 68.65 {\color{gray}\scriptsize ${\pm}$9.23}\\
		\hline
		\hline
		\multirow{2}{*}{\rotatebox{90}{$4$ bit}} & integer conversion & 5.81 & 90.46 {\color{gray}\scriptsize ${\pm}$0.2} & 90.40 {\color{gray}\scriptsize ${\pm}$0.21} & 90.39 {\color{gray}\scriptsize ${\pm}$0.22} & 90.36 {\color{gray}\scriptsize ${\pm}$0.2} & 90.36 {\color{gray}\scriptsize ${\pm}$0.22} & 90.39 {\color{gray}\scriptsize ${\pm}$0.22}\\
		& proper rounding & 5.29 & 5.49 {\color{gray}\scriptsize ${\pm}$0.04} & 5.75 {\color{gray}\scriptsize ${\pm}$0.06} & 5.99 {\color{gray}\scriptsize ${\pm}$0.09} & 7.71 {\color{gray}\scriptsize ${\pm}$0.36} & 10.62 {\color{gray}\scriptsize ${\pm}$1.08} & 15.79 {\color{gray}\scriptsize ${\pm}$2.54}\\
		\hline
	\end{tabular}
	\vspace*{-0.1cm} 
\end{table*}

For evaluation, we assume that the DNN weights are mapped linearly onto the memory of these chips. The bit error maps are of size $8192 \times 128$ bits for chips 2 and 3 and $2048 \times 128$ bits for chip 1. Furthermore, to simulate various different mappings, we repeat this procedure with various offsets and compute average \RTE across all mappings. For results, we refer to \appref{subsec:supp-randbet-baselines}.

\subsection{Bounding Generalization to Random Bit Errors}
\label{subsec:supp-bound}

Let $w$ denote the final weights of a trained DNN $f$. We test $f$ using $n$ i.i.d. test examples, \ie, $(x_i,y_i)_{i=1}^n$. We denote by $w'$ the weights where each bit of $w$ is flipped with probability $p$ uniformly at random, corresponding to the error model from \secref{sec:errors}.
The expected \emph{clean} error  of $f$ is given by
\begin{align*}
	\Exp[\Id_{f(x;w)\neq y}] = \Pr(f(x;w)\neq y).
\end{align*}
The expected \emph{robust} error (regarding i.i.d. test examples drawn from the data distribution) with random bit errors in the (quantized) weights is
\begin{align*}
	\Exp[ \Id_{f(x;w')\neq y}] = \Pr( f(x;w')\neq y).
\end{align*}
Here, the weights of the neural network are themselves random variables. Therefore, with $x, y, w$, and $w'$ we denote the random variables corresponding to test example, test label, weights and weights bit random bit errors. With $x_j, y_j, w_i$ and $w'_i$ we denote actual examples. Then, the following proposition derives a simple, probabilistic bound on the deviation of expected robust error from the empirically measured one (\ie, \RTE in our experiments):

\begin{table*}[t]
	\centering
	\small
	\caption{\textbf{Weight Clipping Improves Robustness.} We report \TE and \RTE for various experiments on the robustness of weight clipping with $\wmax$, \ie, \Clipping[$\wmax$]. First, we show that the robustness benefit of \Clipping is independent of quantization-aware training, robustness also improves when applying post-training quantization. Then, we show results for both symmetric and asymmetric quantization. For the latter we demonstrate that label smoothing \cite{SzegedyCVPR2016} reduces the obtained robustness. This supports our hypothesis that weight clipping, driven by minimizing cross-entropy loss during training, improves robustness through redundancy.}
	\label{tab:supp-clipping}
	\vspace*{-0.25cm}
	\begin{tabular}{| c | l | c | c | c | c | c | c | c |}
		\hline
		\multicolumn{9}{|c|}{\bfseries \CifarT ($\mathbf{m = 8}$ bit): clipping robustness for post- and during-training quantization}\\
		\hline
		&Model & \multirow{2}{*}{\begin{tabular}{c}\TE\\in \%\end{tabular}} & \multicolumn{6}{c|}{\RTE in \%, $p$ in \% p=0.01}\\
		\cline{4-9}
		&&& $0.01$ & $0.05$ & $0.1$ & $0.5$ & $1$ & $1.5$\\
		\hline
		\hline
		\multirow{6}{*}{\rotatebox{90}{\begin{tabular}{@{}c@{}}Post-Training\\Asymmetric\end{tabular}}} & \Normal & 4.37 & 4.95 {\color{gray}\scriptsize ${\pm}$0.11} & 5.47 {\color{gray}\scriptsize ${\pm}$0.17} & 6.03 {\color{gray}\scriptsize ${\pm}$0.22} & 15.42 {\color{gray}\scriptsize ${\pm}$3.4} & 51.83 {\color{gray}\scriptsize ${\pm}$9.92} & 81.74 {\color{gray}\scriptsize ${\pm}$5.14}\\
		& \Quant & 4.27 & 4.59 {\color{gray}\scriptsize ${\pm}$0.08} & 5.10 {\color{gray}\scriptsize ${\pm}$0.13} & 5.54 {\color{gray}\scriptsize ${\pm}$0.15} & 10.59 {\color{gray}\scriptsize ${\pm}$1.11} & 30.58 {\color{gray}\scriptsize ${\pm}$6.05} & 63.72 {\color{gray}\scriptsize ${\pm}$6.89}\\
		& \Clipping[$0.25$] & 4.96 & 5.24 {\color{gray}\scriptsize ${\pm}$0.07} & 5.73 {\color{gray}\scriptsize ${\pm}$0.14} & 6.16 {\color{gray}\scriptsize ${\pm}$0.21} & 10.51 {\color{gray}\scriptsize ${\pm}$0.91} & 26.27 {\color{gray}\scriptsize ${\pm}$5.65} & 61.49 {\color{gray}\scriptsize ${\pm}$9.03}\\
		& \Clipping[$0.2$] & 5.24 & 5.48 {\color{gray}\scriptsize ${\pm}$0.05} & 5.87 {\color{gray}\scriptsize ${\pm}$0.09} & 6.23 {\color{gray}\scriptsize ${\pm}$0.13} & 9.47 {\color{gray}\scriptsize ${\pm}$0.7} & 19.78 {\color{gray}\scriptsize ${\pm}$3.58} & 43.64 {\color{gray}\scriptsize ${\pm}$8.2}\\
		& \Clipping[$0.15$] & 5.38 & 5.63 {\color{gray}\scriptsize ${\pm}$0.05} & 6.03 {\color{gray}\scriptsize ${\pm}$0.09} & 6.38 {\color{gray}\scriptsize ${\pm}$0.13} & 8.80 {\color{gray}\scriptsize ${\pm}$0.41} & 15.74 {\color{gray}\scriptsize ${\pm}$2.24} & 36.29 {\color{gray}\scriptsize ${\pm}$7.34}\\
		& \Clipping[$0.1$] & 5.32 & 5.52 {\color{gray}\scriptsize ${\pm}$0.04} & 5.82 {\color{gray}\scriptsize ${\pm}$0.06} & 6.05 {\color{gray}\scriptsize ${\pm}$0.07} & 7.45 {\color{gray}\scriptsize ${\pm}$0.26} & 9.80 {\color{gray}\scriptsize ${\pm}$0.62} & 17.56 {\color{gray}\scriptsize ${\pm}$3.08}\\
		\hline
		\hline
		\multirow{7}{*}{\rotatebox{90}{\begin{tabular}{@{}c@{}}Symmetric\\(during training)\end{tabular}}} & \Normal & 4.36 & 4.82 {\color{gray}\scriptsize ${\pm}$0.07} & 5.51 {\color{gray}\scriptsize ${\pm}$0.19} & 6.37 {\color{gray}\scriptsize ${\pm}$0.32} & 24.76 {\color{gray}\scriptsize ${\pm}$4.71} & 72.65 {\color{gray}\scriptsize ${\pm}$6.35} & 87.40 {\color{gray}\scriptsize ${\pm}$2.47}\\
		& \Quant & 4.39 & 4.77 {\color{gray}\scriptsize ${\pm}$0.08} & 5.43 {\color{gray}\scriptsize ${\pm}$0.21} & 6.10 {\color{gray}\scriptsize ${\pm}$0.32} & 17.11 {\color{gray}\scriptsize ${\pm}$3.07} & 55.35 {\color{gray}\scriptsize ${\pm}$9.4} & 82.84 {\color{gray}\scriptsize ${\pm}$4.52}\\
		& \Clipping[$0.25$] & 4.63 & 4.99 {\color{gray}\scriptsize ${\pm}$0.07} & 5.53 {\color{gray}\scriptsize ${\pm}$0.1} & 6.06 {\color{gray}\scriptsize ${\pm}$0.16} & 13.55 {\color{gray}\scriptsize ${\pm}$1.42} & 41.64 {\color{gray}\scriptsize ${\pm}$7.35} & 73.39 {\color{gray}\scriptsize ${\pm}$7.15}\\
		& \Clipping[$0.2$] & 4.50 & 4.79 {\color{gray}\scriptsize ${\pm}$0.06} & 5.25 {\color{gray}\scriptsize ${\pm}$0.09} & 5.65 {\color{gray}\scriptsize ${\pm}$0.16} & 9.64 {\color{gray}\scriptsize ${\pm}$0.99} & 21.37 {\color{gray}\scriptsize ${\pm}$4.23} & 45.68 {\color{gray}\scriptsize ${\pm}$7.9}\\
		& \Clipping[$0.15$] & 5.18 & 5.42 {\color{gray}\scriptsize ${\pm}$0.05} & 5.76 {\color{gray}\scriptsize ${\pm}$0.08} & 6.07 {\color{gray}\scriptsize ${\pm}$0.09} & 8.36 {\color{gray}\scriptsize ${\pm}$0.43} & 13.80 {\color{gray}\scriptsize ${\pm}$1.45} & 24.70 {\color{gray}\scriptsize ${\pm}$3.77}\\
		& \Clipping[$0.1$] & 4.86 & 5.07 {\color{gray}\scriptsize ${\pm}$0.04} & 5.34 {\color{gray}\scriptsize ${\pm}$0.06} & 5.59 {\color{gray}\scriptsize ${\pm}$0.1} & 7.12 {\color{gray}\scriptsize ${\pm}$0.3} & 9.44 {\color{gray}\scriptsize ${\pm}$0.7} & 13.14 {\color{gray}\scriptsize ${\pm}$1.79}\\
		& \Clipping[$0.05$] & 5.56 & 5.70 {\color{gray}\scriptsize ${\pm}$0.03} & 5.89 {\color{gray}\scriptsize ${\pm}$0.06} & 6.03 {\color{gray}\scriptsize ${\pm}$0.08} & 6.68 {\color{gray}\scriptsize ${\pm}$0.14} & 7.31 {\color{gray}\scriptsize ${\pm}$0.2} & 8.06 {\color{gray}\scriptsize ${\pm}$0.36}\\
		\hline
		\hline
		\multirow{11}{*}{\rotatebox{90}{\begin{tabular}{@{}c@{}}{\color{red}\textbf{A}}symmetric (default) quant.\\(during training)\end{tabular}}} & \Normal & 4.36 & 4.82 {\color{gray}\scriptsize ${\pm}$0.07} & 5.51 {\color{gray}\scriptsize ${\pm}$0.19} & 6.37 {\color{gray}\scriptsize ${\pm}$0.32} & 24.76 {\color{gray}\scriptsize ${\pm}$4.71} & 72.65 {\color{gray}\scriptsize ${\pm}$6.35} & 87.40 {\color{gray}\scriptsize ${\pm}$2.47}\\
		& \Quant & 4.32 & 4.60 {\color{gray}\scriptsize ${\pm}$0.08} & 5.10 {\color{gray}\scriptsize ${\pm}$0.13} & 5.54 {\color{gray}\scriptsize ${\pm}$0.2} & 11.28 {\color{gray}\scriptsize ${\pm}$1.47} & 32.05 {\color{gray}\scriptsize ${\pm}$6} & 68.65 {\color{gray}\scriptsize ${\pm}$9.23}\\
		& \Clipping[$0.25$] & 4.58 & 4.84 {\color{gray}\scriptsize ${\pm}$0.05} & 5.29 {\color{gray}\scriptsize ${\pm}$0.12} & 5.71 {\color{gray}\scriptsize ${\pm}$0.16} & 10.52 {\color{gray}\scriptsize ${\pm}$1.14} & 27.95 {\color{gray}\scriptsize ${\pm}$4.16} & 62.46 {\color{gray}\scriptsize ${\pm}$8.89}\\
		& \Clipping[$0.2$] & 4.63 & 4.91 {\color{gray}\scriptsize ${\pm}$0.05} & 5.28 {\color{gray}\scriptsize ${\pm}$0.08} & 5.62 {\color{gray}\scriptsize ${\pm}$0.11} & 8.27 {\color{gray}\scriptsize ${\pm}$0.35} & 18.00 {\color{gray}\scriptsize ${\pm}$2.84} & 53.74 {\color{gray}\scriptsize ${\pm}$8.89}\\
		& \Clipping[$0.15$] & 4.42 & 4.66 {\color{gray}\scriptsize ${\pm}$0.05} & 5.01 {\color{gray}\scriptsize ${\pm}$0.09} & 5.31 {\color{gray}\scriptsize ${\pm}$0.12} & 7.81 {\color{gray}\scriptsize ${\pm}$0.6} & 13.08 {\color{gray}\scriptsize ${\pm}$2.21} & 23.85 {\color{gray}\scriptsize ${\pm}$5.07}\\
		& \Clipping[$0.1$] & 4.82 & 5.04 {\color{gray}\scriptsize ${\pm}$0.04} & 5.33 {\color{gray}\scriptsize ${\pm}$0.07} & 5.58 {\color{gray}\scriptsize ${\pm}$0.1} & 6.95 {\color{gray}\scriptsize ${\pm}$0.24} & 8.93 {\color{gray}\scriptsize ${\pm}$0.46} & 12.22 {\color{gray}\scriptsize ${\pm}$1.29}\\
		& \Clipping[$0.05$] & 5.44 & 5.59 {\color{gray}\scriptsize ${\pm}$0.04} & 5.76 {\color{gray}\scriptsize ${\pm}$0.07} & 5.90 {\color{gray}\scriptsize ${\pm}$0.07} & 6.53 {\color{gray}\scriptsize ${\pm}$0.13} & 7.18 {\color{gray}\scriptsize ${\pm}$0.16} & 7.92 {\color{gray}\scriptsize ${\pm}$0.25}\\
		\cline{2-9} 
		& \Clipping[$0.2$]+LS & 4.48 & 4.77 {\color{gray}\scriptsize ${\pm}$0.05} & 5.19 {\color{gray}\scriptsize ${\pm}$0.1} & 5.55 {\color{gray}\scriptsize ${\pm}$0.12} & 9.46 {\color{gray}\scriptsize ${\pm}$0.82} & 32.49 {\color{gray}\scriptsize ${\pm}$5.07} & 68.60 {\color{gray}\scriptsize ${\pm}$7.33}\\
		& \Clipping[$0.15$]+LS & 4.67 & 4.86 {\color{gray}\scriptsize ${\pm}$0.05} & 5.23 {\color{gray}\scriptsize ${\pm}$0.08} & 5.83 {\color{gray}\scriptsize ${\pm}$0.12} & 7.99 {\color{gray}\scriptsize ${\pm}$0.43} & 29.40 {\color{gray}\scriptsize ${\pm}$6.99} & 68.99 {\color{gray}\scriptsize ${\pm}$8.48}\\
		& \Clipping[$0.1$]+LS & 4.82 & 5.05 {\color{gray}\scriptsize ${\pm}$0.04} & 5.37 {\color{gray}\scriptsize ${\pm}$0.08} & 6.10 {\color{gray}\scriptsize ${\pm}$0.11} & 7.36 {\color{gray}\scriptsize ${\pm}$0.4} & 10.59 {\color{gray}\scriptsize ${\pm}$1.01} & 18.31 {\color{gray}\scriptsize ${\pm}$2.84}\\
		& \Clipping[$0.05$]+LS & 5.30 & 5.43 {\color{gray}\scriptsize ${\pm}$0.03} & 5.63 {\color{gray}\scriptsize ${\pm}$0.06} & 6.43 {\color{gray}\scriptsize ${\pm}$0.07} & 6.51 {\color{gray}\scriptsize ${\pm}$0.15} & 7.30 {\color{gray}\scriptsize ${\pm}$0.23} & 8.06 {\color{gray}\scriptsize ${\pm}$0.38}\\
		\hline
	\end{tabular}
	\vspace*{-0.1cm}
\end{table*}

\begin{proposition}
	\label{prop:bound}
	Let $w'_i$, $i = 1,\ldots,l$ be $l$ examples of weights bit random bit errors (each bit flipped with probability $p$). Then it holds
	\begin{align*}
		\Pr\Big( \frac{1}{nl}&\sum_{j=1}^n \sum_{i=1}^l \Id_{f(x_j;w'_i)\neq y_j} - \Pr(f(x;w')\neq y)\geq \epsilon\Big)\\
		&\quad\leq (n+1) e^{-n\epsilon^2 \frac{l}{(\sqrt{l}+\sqrt{n})^2}}.
	\end{align*}
	As alternative formulation, with probability $1-\delta$ it holds
	\begin{align*}
		\Pr( f(x; w'_i)\neq y) < & \frac{1}{nl}\sum_{j=1}^n \sum_{i=1}^l \Id_{f(x_j; w'_i)\neq y_j}\\ &+ \quad \leq\sqrt{\frac{\log\Big(\frac{n+1}{\delta}\Big)}{n}} \frac{\sqrt{l}+\sqrt{n}}{\sqrt{l}}.
	\end{align*} 
\end{proposition}
{\small
\begin{proof}
Let $0<\alpha<1$. Using the Hoeffding inequality and union bound, we have:
\begin{align*}
	&\Pr\Big(\maxop_{j=1,\ldots,n} \frac{1}{l}\sum_{i=1}^l \Id_{f(x_j; w'_i)\neq y_j} - \Exp_{w'}[\Id_{f(x_j; w')\neq y_j}] > \alpha\epsilon\Big) \\
	=& \Pr\Big(\bigcup_{j=1,\ldots,n}\big\{ \frac{1}{l}\sum_{i=1}^l \Id_{f(x_j; w'_i)\neq y_j} - \Exp_{w'}[\Id_{f(x_j; w')\neq y_j}] > \alpha\epsilon\big\}\Big)\\
	&\leq \; n\, e^{-l\alpha^2 \epsilon^2}.
\end{align*}
Then, again by Hoeffding's inequality, it holds:
\begin{align*}
	&\Pr\Big( \frac{1}{n}\sum_{j=1}^n \Exp_{w'}[\Id_{f(x_j; w')\neq y_j}]
	- \Exp_{x,y}[\Exp_{w'}[\Id_{f(x; w')\neq y}]] > (1-\alpha) \epsilon \Big)\\
	&\leq \; e^{-n\epsilon^2 (1-\alpha)^2}.
\end{align*}
Thus, using
\begin{align*}
	a + b>\epsilon \Longrightarrow \{a > \alpha \epsilon\} \cup \{b > (1-\alpha)\epsilon\}
\end{align*}
gives us:
\begin{align*}
	&\Pr\Big( \frac{1}{nl}\sum_{j=1}^n \sum_{i=1}^l \Id_{f(x_j; w'_i)\neq y_j} - \Pr( f(x; w')\neq y)\geq \epsilon\Big)\\
	=& \Pr\Big( \frac{1}{n}\sum_{j=1}^n \big(\frac{1}{l}\sum_{i=1}^l \Id_{f_{w'_i}(x_j)\neq y_j} - \Exp_{w'}[\Id_{f(x_j; w'_i)\neq y_j}]\big)\\
	 +& \frac{1}{n}\sum_{j=1}^n \Exp_{w'}[\Id_{f(x_j; w'_i)\neq y_j}] - \Pr( f(x; w')\neq y)\geq \epsilon\Big)\\
	\leq & \Pr\Big( \frac{1}{n}\sum_{j=1}^n \big(\frac{1}{l}\sum_{i=1}^l \Id_{f(x_j; w'_i)\neq y_j} - \Exp_{w'}[\Id_{f(x_j; w')\neq y_j}]\big)>\alpha \epsilon\Big)\\
	+& \Pr\Big( \frac{1}{n}\sum_{j=1}^n \Exp_{w'}[\Id_{f(x_j; w')\neq y_j}] - \Pr( f(x; w')\neq y)\geq (1-\alpha)\epsilon\Big)\\
	\leq &  n\, e^{-l\alpha^2 \epsilon^2} + e^{-n\epsilon^2 (1-\alpha)^2} 
\end{align*}
Having both exponential terms have the same exponent yields $\alpha=\frac{\sqrt{n}}{\sqrt{l}+\sqrt{n}}$ and we get the upper bound of the proposition.
\end{proof}
}

\textbf{Remarks:}
The samples of bit error injected weights $\{w'_i\}_{i=1}^l$ can actually be different for any test example $(x_j, y_j)$, even though this is not the case in our evaluation. Thus, the above bound involves a stronger result: for any test example, the empirical test error with random bit errors (\ie, robust test error \RTE) and the expected one have to be similar with the same margin. Note also that this bound holds for any fixed bit error distribution as the only requirement is that the bit error patterns we draw are i.i.d. but not the bit errors on the pattern.
In \appref{subsec:experiments-stress}, we consider results with $l = 10^6$, \ie, $l \gg n$ with $n = 10^4$ on \CifarT such that $\nicefrac{l}{(\sqrt{l}+\sqrt{n})^2}$ tends towards one. With $\delta=0.99$ the excess term $\sqrt{\frac{\log\Big(\frac{n+1}{\delta}\Big)}{n}} \frac{\sqrt{l}+\sqrt{n}}{\sqrt{l}}$ in the Proposition is equal to $4.1\%$. Thus larger test sets would be required to get stronger guarantees e.g. for $n=10^5$ one would get $1.7\%$.

\section{Quantization and Bit Manipulation in PyTorch}
\label{sec:supp-implementation}

Our fixed-point quantization $Q$ in \eqnref{eq:quantization} quantizes weights $w_i \in [-\qmax, \qmax] \subset \mathbb{R}$ into signed integers $\{-2^{m-1}-1,\ldots,2^{m-1}-1\}$. Here, the quantization range $[-\qmax,\qmax]$ is symmetric around zero. Note that zero is represented exactly. To implement asymmetric quantization, as outlined in \secref{subsec:robustness-quantization}, the same scheme can be used to quantize weights $w_i \in [\qmin, \qmax]$ within any arbitrary, potentially asymmetric, interval. To this end, \eqnref{eq:quantization} with $\qmax = 1$ is used and the weights in $[\qmin,\qmax]$ are mapped linearly to $[-1, 1]$ using the transformation $N$:
\begin{align}
	N(w_i) = \left(\frac{w_i - \qmin}{\qmax - \qmin}\right)\cdot 2 - 1.\label{eq:asymmetric-quantization}
\end{align}
Generally, $\qmin$ and $\qmax$ are chosen to reflect minimum and maximum weight value -- either from all weights (global quantization) or per-layer.
Furthermore, we argue that asymmetric quantization becomes more robust when using \emph{unsigned} integers as representation. In this case, \eqnref{eq:quantization} can be adapted using a simple additive term:
\begin{align}
	\begin{split}
		Q(w_i) &= \left\lceil \frac{w_i}{\Delta}\right\rfloor + (2^{m - 1} - 1)\\
		Q^{-1}(v_i) &= \Delta (v_i - (2^{m - 1} - 1))
	\end{split}\label{eq:unsigned-quantization}
\end{align}
We use asymmetric quantization using $N$ in \eqnref{eq:asymmetric-quantization} with \eqnref{eq:unsigned-quantization} as our \emph{robust} fixed-point quantization.

\begin{table}[t]
	\centering
	\small
	\caption{\textbf{Batch Normalization not Robust.} \RTE with group normalization (GN) or batch normalization (BN). \RTE increases when using BN even though clean \TE improves slightly compared GN. However, using batch statistics at test time (\ie, ``training mode'' in PyTorch) improves \RTE significantly indicating that the statistics accumulated throughout training do not account for random bit errors. We use \textbf{group normalization as default.}}
	\label{tab:supp-bn}
	\vspace*{-0.25cm}
	\hspace*{-0.25cm}
	\begin{tabular}{| c | l | c | c | c |}
		\hline
		\multicolumn{5}{|c|}{\bfseries \CifarT ($\mathbf{m = 8}$ bit): robustness of BN}\\
		\hline
		&& \multirow{2}{*}{\begin{tabular}{@{}c@{}}\TE\\in \%\end{tabular}} & \multicolumn{2}{c|}{\RTE in \%}\\
		\hline
		&& & $p{=}0.1$ & $p{=}0.5$\\
		\hline
		\hline
		\multirow{2}{*}{GN} & \Normal & 4.32 & 5.54 & 11.28\\
		& \Clipping[$0.1$] & 4.82 & 5.58 & 6.95\\
		\hline
		\hline
		\multicolumn{5}{|c|}{\textbf{BN w/ \emph{Accumulated} Statistics}}\\
		\hline
		\multirow{2}{*}{BN} & \Normal & 3.83 & 6.36 & 52.52\\
		& \Clipping[$0.1$] & 4.46 & 5.32 & 8.25\\
		\hline
		\hline
		\multicolumn{5}{|c|}{\textbf{BN w/ Batch Statistics at Test Time}}\\
		\hline
		\multirow{2}{*}{BN} & \Normal & 3.83 & 6.65 & 9.63\\
		& \Clipping[$0.1$] & 4.46 & 6.57 & 7.29\\
		\hline
	\end{tabular}
	\vspace*{-0.1cm}
\end{table}

Following \secref{subsec:robustness-quantization}, we implement ``fake'' fixed-point quantization for quantization-aware training and bit error injection directly in PyTorch \cite{PaszkeNIPSWORK2017}. Here, fake quantization means that computation is performed in floating point, but before doing a forward pass, the DNN is quantized and dequantized, \ie, $w_q = Q^{-1}(Q(w))$ in \algref{alg:training}. Note that we quantize into \emph{unsigned} $8$ bit integers, irrespective of the target precision $m \leq 8$. To later induce random bit errors, the $8 - m$ most significant bits (MSBs) are masked for $m < 8$. Bit manipulation of unsigned $8$ bit integers is then implemented in C/CUDA and interfaced to Python using CuPy \cite{cupy} or CFFI \cite{cffi}. These functions can directly operate on PyTorch tensors, allowing bit manipulation on the CPU as well as the GPU. We will make our code publicly available to faciliate research into DNN robustness against random bit errors.

\section{Weight Clipping with Group/Batch Normalization}
\label{sec:supp-clipping}

While weight clipping, \ie, globally constraining weights to $[-\wmax, \wmax]$ during training, is easy to implement, we make a simple adjustment to group and batch normalization layers: we reparameterize the scale parameter $\alpha$ of batch/group normalization, which usually defaults to $\alpha = 1$ and may cause problems when clipped, \eg, to $[-0.1, 0.1]$. In particular with aggressive weight clipping, $\alpha \leq \wmax < 1$, the normalization layers loose their ability to represent the identity function, considered important for batch normalization in \cite{IoffeICML2015}. Our reparameterization introduces a learnable, auxiliary parameter $\alpha'$ such that $\alpha$ as $\alpha = 1 + \alpha'$ to solve this problem.
	\section{Experimental Setup}
\label{subsec:supp-experiments-setup}

\textbf{Datasets:} We conduct experiments on \MNIST\footnote{\url{http://yann.lecun.com/exdb/mnist/}} \citep{LecunIEEE1998} and \Cifar\footnote{\url{https://www.cs.toronto.edu/~kriz/cifar.html}} \cite{Krizhevsky2009}. \MNIST consists of $60\text{k}$ training and $10\text{k}$ test images from $10$ classes. These are gray-scale and of size $28 \times 28$ pixels. \Cifar consists of $50\text{k}$ training and $10\text{k}$ test images of size $32\times 32\times 3$ (\ie, color images). \CifarT has images corresponding to $10$ classes, \CifarH contains images from $100$ classes.

\begin{table}[t]
	\centering
	\small
	\caption{\textbf{Weight Clipping with Weight Scaling.} For group normalization (GN) without the reparameterization in \secref{subsec:robustness-clipping}, using fixed scale/bias instead, our DNNs are scale-invariant. Scaling \Quant down to the weight range of \Clipping[$0.25$], however, does not improve robustness. Thus, the robustness benefit of \Clipping is \emph{not} due to reduced quantization range or smaller absolute errors.}
	\label{tab:supp-clipping-scaling}
	\vspace*{-0.25cm}
	\begin{tabular}{| l | c | c | c |}
		\hline
		\multicolumn{4}{| c |}{\bfseries \CifarT ($\mathbf{m = 8}$ bit): scaling w/o reparameterized GN}\\
		\hline
		Model & \multirow{2}{*}{\begin{tabular}{@{}c@{}}\TE\\in \%\end{tabular}} & \multicolumn{2}{c|}{\RTE in \%, $p$ in \%}\\
		\hline
		(see text) & & $p{=}0.1$ & $p{=}1$\\
		\hline
		\hline
		\Quant & 4.67 & 6.12 {\color{gray}\scriptsize ${\pm}$0.2} & 35.25 {\color{gray}\scriptsize ${\pm}$6.41}\\
		\Clipping[$0.25$] & 4.96 & 6.13 {\color{gray}\scriptsize ${\pm}$0.16} & 16.09 {\color{gray}\scriptsize ${\pm}$1.85}\\
		\hline
		\Quant $\rightarrow$ \Clipping[$0.25$] & 4.64 & 6.10 {\color{gray}\scriptsize ${\pm}$0.18} & 35.28 {\color{gray}\scriptsize ${\pm}$5.82}\\
		\hline
	\end{tabular}
	\vspace*{-0.1cm} 
\end{table}
\begin{figure*}[t]
	\centering
	\vspace*{-0.2cm}
	\begin{subfigure}{0.35\textwidth}
		\vspace*{0px}
		\centering
		\Quant
		
		\includegraphics[width=1\textwidth]{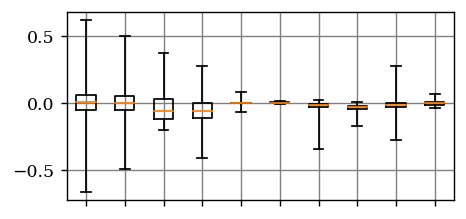}
	\end{subfigure}
	\begin{subfigure}{0.35\textwidth}
		\vspace*{0px}
		\centering
		\Random (w/o weight clipping)
		
		\includegraphics[width=1\textwidth]{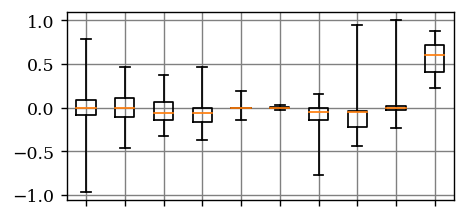}
	\end{subfigure}
	\\
	\begin{subfigure}{0.35\textwidth}
		\vspace*{0px}
		\centering
		\Clipping[$0.1$]
			
		\includegraphics[width=1\textwidth]{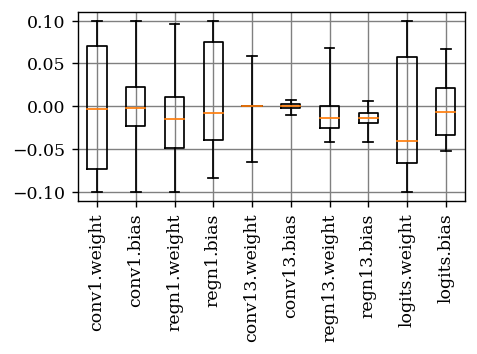}
	\end{subfigure}
	\begin{subfigure}{0.35\textwidth}
		\vspace*{0px}
		\centering
		\Clipping[$0.05$]
			
		\includegraphics[width=1\textwidth]{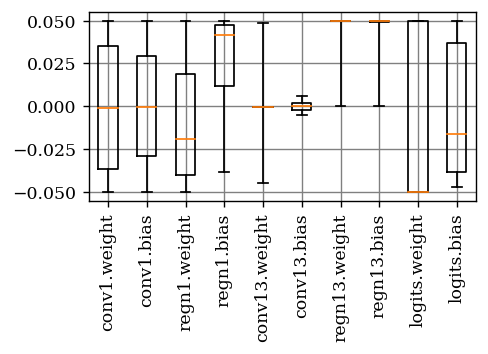}
	\end{subfigure}
	\\
	\begin{subfigure}{0.16\textwidth}
		\vspace*{0px}
		\centering
		\Quant
		\includegraphics[width=1\textwidth]{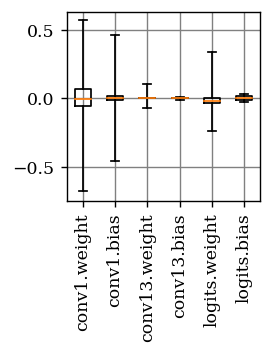}
	\end{subfigure}
	\begin{subfigure}{0.16\textwidth}
		\vspace*{0px}
		\centering
		\Clipping[$0.25$] $\uparrow$
			
		\includegraphics[width=1\textwidth]{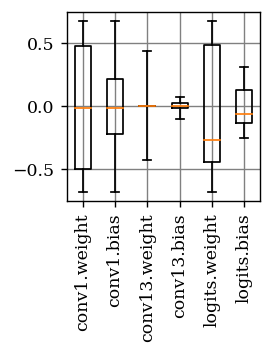}
	\end{subfigure}
	\begin{subfigure}{0.35\textwidth}
		\includegraphics[width=1\textwidth]{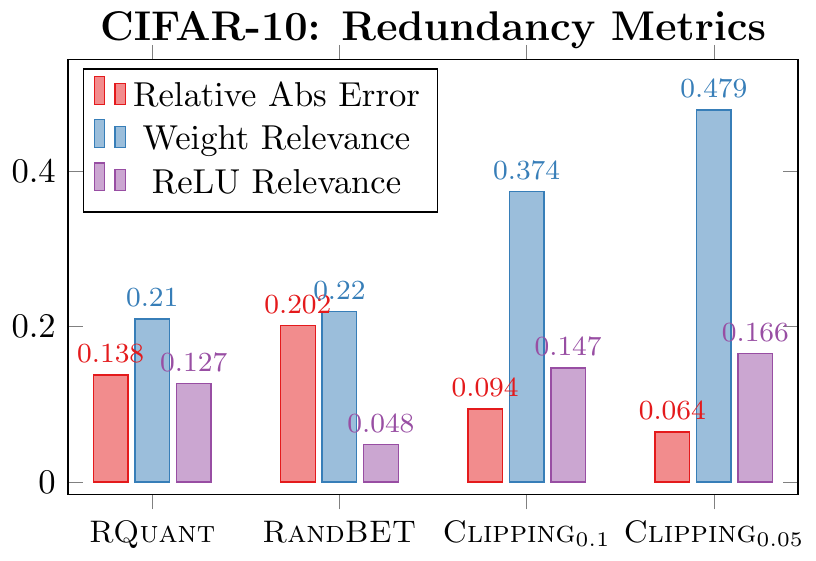}
	\end{subfigure}
	\vspace*{-6px}
	\caption{\textbf{Weight Clipping Increases Redundancy.} We show weight distributions of selected layers (top) for \Quant, \Random (without weight clipping) as well as \Clipping[$0.1$] and \Clipping[$0.05$]. We show weights and biases for the logit layer as well as the first and last (13th) convolutional layer. Scale/Bias parameters of GN are also included. Below (left), this is shown for the scaling experiment from \tabref{tab:supp-clipping-scaling} (GN parameters are fixed). Note that \Random only affects the logit layer, while \Clipping increases the used (relative) weight range significantly. On the bottom (right), we plot various measures of redundancy, see the text for discussion and details. The relative absolute error is computed considering random bit errors with probability $p = 1\%$.}
	\label{fig:supp-clipping}
	\vspace*{-0.1cm}
\end{figure*}

\textbf{Architecture:} The used SimpleNet architectures \cite{HasanpourARXIV2016} are summarized in \tabref{tab:supp-architectures}, including the total number of weights $W$. On \Cifar, this results in a total of roughly $W \approx 5.5\text{M}$ weights. Due to the lower resolution on \MNIST, channel width in each convolutional layer is halved, and one stage of convolutional layers including a pooling layer is skipped. This results in a total of roughly $W \approx 1\text{M}$ weights. In both cases, we replaced batch normalization (BN) \cite{IoffeICML2015} with group normalization (GN) \cite{WuECCV2018}. The GN layers are reparameterized as in \appref{sec:supp-clipping} to facilitate weight clipping. \tabref{tab:supp-architectures} also includes the expected number of bit errors given various rates $p$ for random bit errors. Regarding the number of weights $W$, SimpleNet compares favorably to, \eg, VGG \cite{SimonyanICLR2015}: VGG-16 has $14\text{M}$ weights on \Cifar. Additionally, we found SimpleNet to be easier to train without BN, which is desirable as BN reduces robustness to bit errors significantly, \cf \appref{subsec:supp-experiments-bn}. The ResNet-50 \cite{HeCVPR2016} used for experiments in \appref{subsec:supp-experiments-architectures} follows the official PyTorch \cite{PaszkeNIPSWORK2017} implementation. The Wide ResNet (WRN) \cite{ZagoruykoBMVC2016} used on \CifarH is adapted from\footnote{\url{https://github.com/meliketoy/wide-resnet.pytorch}}, but we use $12$ base channels, instead of $16$, reducing $W$ from roughly $36.5\text{Mio}$ to $20.5\text{Mio}$.

\begin{table*}[t]
	\centering
	\small
	\caption{\textbf{\Random Robustness with \emph{Symmetric} Quantization.} Average \RTE and standard deviation for \Clipping and \Random with $\wmax = 0.1$ and \textbf{s}ymmetric quantization, \ie, larger quantization range than {\color{red}a}symmetric quantization. Also \cf \tabref{tab:supp-clipping} and \tabref{tab:supp-summary-cifar10}. Robustness decreases slightly compared to asymmetric quantization, however, \Clipping and \Random are still effective in reducing \RTE against high bit error rates $p$.}
	\label{tab:supp-randbet-symmetric}
	\vspace*{-0.25cm}
	\begin{tabular}{|l | c | c | c | c | c | c | c |}
		\hline
		\multicolumn{8}{|c|}{\bfseries \CifarT ($\mathbf{m = 8}$ bit): \Random with symmetric quantization}\\
		\hline
		Model & \multirow{2}{*}{\begin{tabular}{c}\TE\\in \%\end{tabular}} & \multicolumn{6}{c|}{\RTE in \%, $p$ in \% p=0.01}\\
		\cline{3-8}
		&& $0.01$ & $0.05$ & $0.1$ & $0.5$ & $1$ & $1.5$\\
		\hline
		\hline
		\Normal & 4.36 & 4.82 {\color{gray}\scriptsize ${\pm}$0.07} & 5.51 {\color{gray}\scriptsize ${\pm}$0.19} & 6.37 {\color{gray}\scriptsize ${\pm}$0.32} & 24.76 {\color{gray}\scriptsize ${\pm}$4.71} & 72.65 {\color{gray}\scriptsize ${\pm}$6.35} & 87.40 {\color{gray}\scriptsize ${\pm}$2.47}\\
		\Quant & 4.39 & 4.77 {\color{gray}\scriptsize ${\pm}$0.08} & 5.43 {\color{gray}\scriptsize ${\pm}$0.21} & 6.10 {\color{gray}\scriptsize ${\pm}$0.32} & 17.11 {\color{gray}\scriptsize ${\pm}$3.07} & 55.35 {\color{gray}\scriptsize ${\pm}$9.4} & 82.84 {\color{gray}\scriptsize ${\pm}$4.52}\\
		\Clipping[$0.1$] & 4.86 & 5.07 {\color{gray}\scriptsize ${\pm}$0.04} & 5.34 {\color{gray}\scriptsize ${\pm}$0.06} & 5.59 {\color{gray}\scriptsize ${\pm}$0.1} & 7.12 {\color{gray}\scriptsize ${\pm}$0.3} & 9.44 {\color{gray}\scriptsize ${\pm}$0.7} & 13.14 {\color{gray}\scriptsize ${\pm}$1.79}\\
		\hline
		\Random[$0.1$] $p{=}0.01$ & 5.07 & 5.27 {\color{gray}\scriptsize ${\pm}$0.04} & 5.54 {\color{gray}\scriptsize ${\pm}$0.07} & 5.73 {\color{gray}\scriptsize ${\pm}$0.11} & 7.18 {\color{gray}\scriptsize ${\pm}$0.29} & 9.63 {\color{gray}\scriptsize ${\pm}$0.9} & 13.81 {\color{gray}\scriptsize ${\pm}$2.2}\\
		\Random[$0.1$] $p{=}0.1$ & 4.62 & 4.83 {\color{gray}\scriptsize ${\pm}$0.04} & 5.09 {\color{gray}\scriptsize ${\pm}$0.08} & 5.31 {\color{gray}\scriptsize ${\pm}$0.08} & 6.70 {\color{gray}\scriptsize ${\pm}$0.28} & 8.89 {\color{gray}\scriptsize ${\pm}$0.59} & 12.20 {\color{gray}\scriptsize ${\pm}$1.33}\\
		\Random[$0.1$] $p{=}1$ & 5.03 & 5.22 {\color{gray}\scriptsize ${\pm}$0.04} & 5.43 {\color{gray}\scriptsize ${\pm}$0.06} & 5.61 {\color{gray}\scriptsize ${\pm}$0.07} & 6.56 {\color{gray}\scriptsize ${\pm}$0.13} & 7.70 {\color{gray}\scriptsize ${\pm}$0.26} & 8.99 {\color{gray}\scriptsize ${\pm}$0.42}\\
		\Random[$0.1$] $p{=}1.5$ & 5.24 & 5.37 {\color{gray}\scriptsize ${\pm}$0.03} & 5.57 {\color{gray}\scriptsize ${\pm}$0.06} & 5.76 {\color{gray}\scriptsize ${\pm}$0.07} & 6.66 {\color{gray}\scriptsize ${\pm}$0.14} & 7.62 {\color{gray}\scriptsize ${\pm}$0.25} & 8.71 {\color{gray}\scriptsize ${\pm}$0.42}\\
		\Random[$0.1$] $p{=}2$ & 5.82 & 5.97 {\color{gray}\scriptsize ${\pm}$0.04} & 6.19 {\color{gray}\scriptsize ${\pm}$0.07} & 6.37 {\color{gray}\scriptsize ${\pm}$0.09} & 7.22 {\color{gray}\scriptsize ${\pm}$0.19} & 8.03 {\color{gray}\scriptsize ${\pm}$0.23} & 8.96 {\color{gray}\scriptsize ${\pm}$0.38}\\
		\hline
	\end{tabular}
	\vspace*{-0.1cm}
\end{table*}

\textbf{Training:} As outlined in \secref{sec:experiments}, we use stochastic gradient descent to minimize cross-entropy loss. We use an initial learning rate of $0.05$, multiplied by $0.1$ after $\nicefrac{2}{5}$, $\nicefrac{3}{5}$ and $\nicefrac{4}{5}$ of $100$/$250$ epochs on \MNIST/\Cifar. Our batch size is $128$ and momentum of $0.9$ is used together with weight decay of $5\cdot10^{-4}$. On \Cifar, we whiten the input images and use AutoAugment\footnote{\url{https://github.com/DeepVoltaire/AutoAugment}} \cite{CubukARXIV2018} with Cutout \cite{DevriesARXIV2017}. Cutout is applied with a window size of $16 \times 16$, and independent of AutoAugment, we apply random cropping with up to $4$ pixels. Created black spaces are filled using the mean image color (grayish). Initialization follows \cite{HeICCV2015}. The full training set is used for training, and we do \emph{not} rely on early stopping. For \Random, we use $\lambda = 1$ and start injecting bit errors when the loss is below 1.75 on \MNIST/\CifarT or 3.5 on \CifarH. \tabref{tab:supp-accuracy} highlights clean test error (\TE) obtained for various precision $m$ and compared to other architectures, \eg, ResNet-50, on \CifarT, which performs worse when using GN.

\textbf{Random Bit Errors:} We simulate $50$ different chips with enough memory arrays to accomodate all weights by drawing uniform samples $u^{(c)} \sim U(0, 1)^{W \times m}$ for each chip $c$ and all $m$ bits for a total of $W$ weights. Then, for chip $c$, bit $j$ in weight $w_i$ is flipped iff $u^{(c)}_{ij} \leq p$. This assumes a linear memory layout of all $W$ weights. The pattern, \ie, spatial distribution, of bit errors for chip $c$ is fixed by $u^{(c)}$, while across all $50$ chips, bit errors are uniformly distributed. We emphasize that we pre-determine $u^{(c)}$, $c = 1,\ldots,50$, once for all our experiments using fixed random seeds. Thus, our robustness results are entirely comparable across all models as well as bit error rates $p$. Also note that, as explained in \secref{sec:errors}, the bit errors for a fixed chip $c$ at probability $p' < p$ are a subset of those for bit error rate $p$. The expected number of bit errors for various rates $p$ is summarized in \tabref{tab:supp-architectures}.

\textbf{Implementation Details} are covered in \secref{sec:supp-implementation}.

\section{Experiments}
\label{sec:supp-experiments}

\subsection{Batch Normalization}
\label{subsec:supp-experiments-bn}

We deliberately replace batch normalization (BN) \cite{IoffeICML2015} by group normalization (GN) \cite{WuECCV2018} in our experiments. \tabref{tab:supp-bn} demonstrates that \RTE increases significantly when using BN compared to GN indicating that BN is more vulnerable to bit errors in DNN weights. For example, without clipping, \RTE increases from $11.28\%$ to staggering $52.52\%$ when replacing GN with BN. Note that, following \appref{sec:supp-clipping}, the BN/GN parameters (\ie, scale/bias) are reparameterized to account for weight clipping. The observations in \tabref{tab:supp-bn} can also be confirmed without quantization, \eg, considering random $L_\infty$ noise in the weights. We suspect that the running statistics accumulated during training do not account for the random bit errors at test time, even for \Random. This is confirmed in \tabref{tab:supp-bn} (bottom) showing that \RTE reduces significanlty when using the batch statistics at test time. Generally, BN improves accuracy, but might not be beneficial in terms of robustness, as also discussed for adversariale examples \cite{GallowayARXIV2019}. Using GN also motivates our use of SimpleNet instead of, \eg, ResNet-50, which generally performs worse with GN, \cf \tabref{tab:supp-accuracy}.

\subsection{Robust Quantization}
\label{subsec:supp-experiments-quantization}

\tabref{tab:supp-quantization} shows results complementary to the main paper, considering additional bit error rates $p$. Note that, for $m = 8$ bit, changes in the quantization has neglegible impact on clean \TE. Only the change from gloabl to per-layer quantization makes a difference. However, considering \RTE for larger bit error rates, reducing the quantization range, \eg, through per-layer and asymmetric quantization, improves robustness significantly. Oher aspects of the quantization scheme also play an important role, especially for low-precision such as $m = 4$ bit, \cf \tabref{tab:supp-quantization}, as outlined in the following.

For example, using asymmetric quantization into \emph{signed} integers actually increases \RTE for larger $p$ compared to ``just'' using symmetric per-layer quantization (rows 2 and 3). Using \emph{unsigned} integers instead reduces \RTE significantly. We belive this to be due to the two's complement representation of signed integers being used with an \textbf{a}symmetric quantization range. In symmetric quantization (around $0$, \ie, $[-\qmax, \qmax]$), bit errors in the sign bit incur not only a change of the integer's sign, but also the corresponding change in the weights sign\footnote{An \emph{un}signed integer of value $127$ is represented as $0111 1111$. Flipping the most (left-most) significant bit results in $1111 1111$ corresponding to $255$, \ie, the value increases. For a signed integer in two's complement representation, the same bit flip changes the value from $127$ to $-1$, while $0$-to-$1$ not affecting the sign bit generally increase value (also for negative integers).}. Assuming an asymmetric quantization of $[\qmin, \qmax]$ with $0 < \qmin < \qmax$, bit errors in sign bits are less meaningful. For example, flipping any bit $0$-to-$1$ usually increases the value of the integer. However, a $0$-to-$1$ flip in the sign bit actually decreases the value and produces a \emph{negative} integer. However, this change from positive to negative is not reflected in the corresponding weight value (as $\qmin > 0$). For high bit error rates $p\%$, this happens more and more frequently and these changes seem to have larger impact on DNN performance, \ie, \RTE.

Additionally, we considered the difference between using integer conversion for $\nicefrac{w_i}{\Delta}$ and using proper rounding, \ie, $\lceil\nicefrac{w_i}{\Delta}\rfloor$. We emphasize that, for $m = 8$ bit, there is \emph{no} significant difference in terms of clean \TE. However, using proper rounding reduces the approximation error slightly. For $m = 8$ bit, using $p = 2.5\%$ bit error rate, the average absolute error (in the weights) across $10$ random bit error patterns reduces by $2\%$. Nevertheless, it has significantly larger impact on \RTE. For $m = 4$, this is more pronounced: rounding reduces the average absolute error by roughly $67\%$. Surprisingly, this is not at all reflected in the clean \TE, which only decreases from $5.81\%$ to $5.29\%$. It seems that the DNN learns to compensate these errors during training. At test time, however, \RTE reflects this difference in terms of robustness.

\revision{Overall, we found that robust quantization plays a key role. While both weight clipping (\Clipping) and random bit error training (\Random) can improve robustness further, robust quantization lays the foundation for these improvements to be possible. Thus, we encourage authors to consider robustness in the design of future DNN quantization schemes. Even simple improvements over our basic fixed-point quantization scheme may have significant impact in terms of robustness. For example, proper handling of outliers \cite{ZhuangCVPR2018,SungARXIV2015}, learned quantization \cite{ZhangECCV2018}, or adaptive/non-uniform quantization \cite{ZhouAAAI2018,ParkECCV2018,NagelICCV2019} are promising directions to further improve robustness. Finally, we believe that this also poses new theoretical challenges, \ie, studying (fixed-point) quantization with respect to robustness \emph{and} quantization error.}

\subsection{Weight Clipping}
\label{subsec:supp-experiments-clipping}

\begin{table}[t]
	\centering
	\small
	\caption{\textbf{\Random Variants.} \TE and \RTE for \Random and two variants: curricular \Random, with $p$ being increased slowly from $\nicefrac{p}{20}$ to $p$ during the first half of training; and ``alternating'' \Random where weight updates increasing quantization range, \ie, increasing the maximum absolute weight per layer, are not possible based on gradients from perturbed weights, see \secref{subsec:supp-experiments-randbet} for details. Both variants decrease robustness slightly. This is in contrast to, \eg, \cite{KoppulaMICRO2019}, using curricular training on profiled bit errors with success.}
	\label{tab:supp-randbet-variants}
	\vspace*{-0.25cm}
	\begin{tabular}{| l | c | c | c |}
		\hline
		\multicolumn{4}{|c|}{\bfseries \CifarT ($\mathbf{m = 8}$ bit): \Random variants}\\
		\hline
		& \multirow{2}{*}{\begin{tabular}{@{}c@{}}\TE\\in \%\end{tabular}} & \multicolumn{2}{c|}{\RTE in \%}\\
		\hline
		& & $p{=}0.1$ & $p{=}1$\\
		\hline
		\hline
		\Random $p{=}0.1$, $\wmax = 0.1$ & 4.93 & 5.67 & 8.65\\
		\Random $p{=}1$, $\wmax = 0.1$ & 5.06 & 5.87 & 7.60\\
		Curr. \Random $p{=}1$, $\wmax = 0.1$ & 4.89 & 5.78 & 8.51\\
		Curr. \Random $p{=}1$, $\wmax = 0.1$ & 5.32 & 6.13 & 7.98\\
		Alt. \Random $p{=}1$, $\wmax = 0.1$ & 5.07 & 5.91 & 8.93\\
		Alt. \Random $p{=}1$, $\wmax = 0.1$ & 5.24 & 6.25 & 8.02\\
		\hline
	\end{tabular}
	\vspace*{-0.1cm}
\end{table}

In \tabref{tab:supp-clipping} we present robustness results, \ie, \RTE, for weight clipping. Note that weight clipping constraints the weights during training to $[-\wmax, \wmax]$ through projection. We demonstrate that weight clipping can also be used independent of quantization. To this end, we train DNNs with weight clipping, but without quantization. We apply post-training quantization and evaluate bit error robustness. While the robustness is reduced slightly compared to quantization-aware training \emph{and} weight clipping, the robustness benefits of weight clipping are clearly visible. For example, clipping at $\wmax = 0.1$ improves \RTE from $30.58\%$ to $9.8\%$ against $p = 1\%$ bit error rate when performing post-training quantization. With symmetric quantization-aware training, \Clipping[$0.1$] improves slightly to $7.31\%$. Below (middle), we show results for weight clipping and symmetric quantization. These results are complemented in \tabref{tab:supp-randbet-symmetric} with \Random. Symmetric quantization might be preferable due to reduced computation and energy cost compared to asymmetric quantization. However, this also increases \RTE slightly. Nevertheless, \Clipping consistently improves robustness, independent of the difference in quantization. Finally, on the bottom, we show complementary results to \tabref{tab:clipping-robustness}, confirming the adverse effect of label smoothing \cite{SzegedyCVPR2016} on \RTE, \cf \secref{subsec:experiments-clipping}. \figref{fig:supp-clipping-inf} also shows that the obtained robustness generalizes to other error models such as $L_\infty$ weight perturbations, see caption for details.

\revision{As \Clipping adds an additional hyper-parameter, \tabref{tab:supp-clipping} also illustrates that $\wmax$ can easily be tuned based on clean performance. Specifically, lower $\wmax$ will eventually increase \TE and reduce confidences (alongside increasing cross-entropy loss). This increase in \TE is usually not desirable except when optimizing for robust performance, \ie, considering \RTE. Also, we found that weight clipping does not (negatively) interact with any other hyper-parameters or regularizers. For example, as described in \secref{subsec:supp-experiments-setup}, we use weight clipping in combination with AutoAugment/Cutout and weight decay without problems. Furthermore, it was not necessary to adjust our training setup (\ie, optimizer, learning rate, epochs, \etc), even for low $\wmax$.} 

\begin{table}[t]
	\centering
	\small
	\caption{\textbf{\Random with ResNets.} We report \RTE for \Quant, \Clipping and \Random using ResNet-20 and ResNet-50. According to \tabref{tab:supp-accuracy}, \TE increases significantly when using group normalization for ResNet-50, explaining the generally higher \RTE. However, using ResNets, \Clipping and \Random continue to improve robustness significantly, despite a ResNet-50 having roughly $23.5\text{Mio}$ weights.}
	\label{tab:supp-randbet-resnet}
	\vspace*{-0.25cm}
	\begin{tabular}{| l | c | c | c |}
		\hline
		\multicolumn{4}{|c|}{\bfseries \CifarT ($\mathbf{m = 8}$ bit): ResNet architectures}\\
		\hline
		& \multirow{2}{*}{\begin{tabular}{@{}c@{}}\TE\\in \%\end{tabular}} & \multicolumn{2}{c|}{\RTE in \%}\\
		\hline
		& & $p{=}0.5$ & $p{=}1.5$\\
		\hline
		\hline
		\multicolumn{4}{|c|}{\bfseries ResNet-20}\\
		\hline
		\Quant & 4.34 & 13.89 {\color{gray}\scriptsize ${\pm}$2.45} & 81.25 {\color{gray}\scriptsize ${\pm}$5.08}\\
		\Clipping[$0.1$] & 4.83 & 6.76 {\color{gray}\scriptsize ${\pm}$0.16} & 11.23 {\color{gray}\scriptsize ${\pm}$0.97}\\
		\Random[$0.1$], $p{=}1$ & 5.28 & 6.72 {\color{gray}\scriptsize ${\pm}$0.19} & 8.96 {\color{gray}\scriptsize ${\pm}$0.49}\\
		\hline
		\hline
		\multicolumn{4}{|c|}{\bfseries ResNet-50}\\
		\hline
		\Quant & 6.81 & 32.94 {\color{gray}\scriptsize ${\pm}$5.51} & 90.98 {\color{gray}\scriptsize ${\pm}$0.67}\\
		\Clipping[$0.1$] & 5.99 & 9.27 {\color{gray}\scriptsize ${\pm}$0.44} & 36.39 {\color{gray}\scriptsize ${\pm}$7.03}\\
		\Random[$0.1$], $p{=}1$ & 6.04 & 7.87 {\color{gray}\scriptsize ${\pm}$0.22} & 11.27 {\color{gray}\scriptsize ${\pm}$0.6}\\
		\hline
	\end{tabular}
	\vspace*{-0.1cm}
\end{table}

We hypothesize that weight clipping improves robustness as it encourages redundancy in weights and activations during training. This is because cross-entropy loss encourages large logits and weight clipping forces the DNN to ``utilize'' many different weights to produce large logits. \tabref{tab:supp-clipping-scaling} presents a simple experiment in support of our hypothesis. We already emphasized that, relatively, weight clipping does \emph{not} reduce the impact of bit errors. Nevertheless, when using group normalization(GN) without our reparameterization, the trained DNNs are scale-invariant in their weights. Thus, we down-scale \Normal to have the same maximum absolute weight value as \Clipping[$0.25$] (\Normal $\rightarrow$ \Clipping[$0.25$]).
This scaling is applied globally, not per layer. \tabref{tab:supp-clipping-scaling} shows that ``just'' down-scaling does not induce robustness, as expected. Thus, the benefit of \Clipping in terms of robustness does not come from the reduced quantization range.

\figref{fig:supp-clipping} presents further supporting evidence for our hypothesis: While \Random mainly affects the logits layer, \Clipping clearly increases the weight range used by the DNN. Here, the weight range is understood relative to $\wmax$ (or the maxmimum absolute weight value for \Normal). This is pronounced in particular when up-scaling the clipped model (bottom left). Finally, \figref{fig:supp-clipping} (bootom right) also considers three attempts to measure redundancy in weights and activations. The relative absolute error is computed with respect to $p = 1\%$ bit error rate and decreases for \Clipping, meaning that random bit errors have less impact. \emph{Weight relevance} is computed as the sum of absolute weights, \ie, $\sum_i |w_i|$, normalized by the maximum absolute weight: $\nicefrac{\sum_i |w_i|}{\max_i |w_i|}$. This metric measures how many weights are, considering their absolute value, relevant. Finally, We also measure activation redundancy using ReLU relevance, computing the fraction of non-zero activations after the final ReLU activation. \Clipping increases redundancy in the final layer significantly. Finally, \figref{fig:supp-clipping} (bottom left) shows the difference in weight distributions by upscaling \Clipping[$0.25$] to the same weight range as \Normal. Clearly, \Clipping causes more non-zero weights be learned by the DNN. This can be observed across all types of parameters, \ie, weights or biases as well as convolutional or fully connected layers.

\begin{table}[t]
	\centering
	\small
	\caption{\textbf{Generalization to Profiled Bit Errors.} Complementary to \tabref{tab:randbet-generalization}, we show \RTE on profiled bit errors, chips 1-3, for \Random as well as \Clipping. Note that for chip 3, \Clipping[$0.05$] performs slightly better than \Random.}
	\label{tab:supp-randbet-generalization}
	\vspace*{-0.25cm}
	\hspace*{-0.15cm}
	\begin{tabular}{| c | l | c | c | c |}
		\hline
		\multicolumn{5}{|c|}{\bfseries\CifarT: Generalization to Profiled Bit Errors}\\
		\hline
		Chip & Model & \multirow{2}{*}{\begin{tabular}{@{}c@{}}\TE\\in \%\end{tabular}} & \multicolumn{2}{c|}{\RTE in \%}\\
		\cline{4-5}
		& (\CifarT) && $p{\approx}0.86$ & $p{\approx}2.7$\\
		\hline
		\hline
		\multirow{3}{*}{1} & \Quant & 4.32 & 23.57 & 89.84\\
		& \Clipping[$0.05$] & 5.44 & 7.17 & 10.50\\
		& \Random[$0.05$] $p{=}1.5$ & 5.62 & 7.04 & 9.37\\
		\hline
		\hline
		&&& $p{\approx}0.14$ & $p{\approx}1$\\
		\hline
		\multirow{3}{*}{2} & \Quant & 4.32 & 6.00 & 74.00\\
		& \Clipping[$0.05$] & 5.44 & 5.98 & 10.02\\
		& \Random[$0.05$] $p{=}1.5$ & 5.62 & 6.00 & 9.00\\
		\hline
		\hline
		&&& $p{\approx}0.03$ & $p{\approx}0.5$\\
		\hline
		\multirow{2}{*}{3} & \Quant & 4.32 & 5.47 & 80.49\\
		& \Clipping[$0.05$] & 5.44 & 5.78 & 11.88\\
		& \Random[$0.05$] $p{=}1.5$ & 5.62 & 5.85 & 12.44\\
		\hline
	\end{tabular}
	\vspace*{-0.1cm}
\end{table}

\subsection{Random Bit Error Training (\Random)}
\label{subsec:supp-experiments-randbet}

\tabref{tab:supp-randbet-symmetric} shows complementary results for \Random using \emph{symmetric} quantization. Symmetric quantization generally tends to reduce robustness, \ie, increase \RTE, across all bit error rates $p$, \cf \tabref{tab:randbet-robustness} in the main paper. Thus, the positive impact of \Random is pronounced, \ie, \Random becomes more important to obtain high robustness when less robust fixed-point quantization is used. These experiments also demonstrate the utility of \Random independent of the quantization scheme at hand. 

We consider two variants of \Random motivated by related work \cite{KoppulaMICRO2019}. Specifically, in \cite{KoppulaMICRO2019}, the bit error rate seen during training is increased slowly during training. Note that \cite{KoppulaMICRO2019} trains on fixed bit error patterns. Thus, increasing the bit error rate during training is essential to avoid the effect shown in \tabref{tab:randbet-baselines}: the DNN is supposed to be robustness to any bit error rate $p' < p$ smaller than the target bit error rate. While this is generally the case using our \Random, \tabref{tab:supp-randbet-variants} shows that slowly increasing the random bit error rate during training, called ``curricular'' \Random, has no significant benefit over standard \Random. In fact, \RTE increases slightly. Similarly, we found that \Random tends to increase the range of weights: the weights are ``spread out'', \cf \figref{fig:supp-clipping} (top right). This also increases the quantization range, which has negative impact on robustness as discussed in \secref{subsec:experiments-quantization}. Thus, we experimented with \Random using two weight updates per iteration: one using clean weights, one on weights with bit errors. This is in contrast to averaging both updates as in \algref{alg:training}. Updates computed from perturbed weights are limited to the current quantization ranges, \ie, the maximum absolute error cannot change. This is ensured through projection. This makes sure that \Random does not increase the quantization range during training as changes in the quantization range are limited to updates from clean weights. Again, \tabref{tab:supp-randbet-variants} shows this variant to perform slightly worse.

\begin{table}[t]
	\centering
	\caption{\textbf{Fixed Pattern Bit Error Training.} Complementary results for \tabref{tab:randbet-baselines}, reporting \RTE for training on fixed (\eg, profiled) bit error patterns (\Pattern). We show additional results for chip 2 from \tabref{tab:randbet-generalization}. Note that for \Pattern on chip 1/2 we used only the persistent errors shown in \figref{fig:supp-errors}, which is why the bit error rates deviate from \tabref{tab:randbet-generalization}.}
	\label{tab:supp-randbet-baselines}
	\vspace*{-0.25cm}
	\small
	\begin{tabular}{| L{4cm} | C{1cm} | C{1cm} |}
		\hline
		Model (\CifarT) & \multicolumn{2}{c|}{\RTE in \%, $p$ in \%}\\
		\hline
		\hline
		\textbf{\emph{Profiled} Bit Errors (Chip 1)} & $p{\approx}0.39$ & $p{\approx}1.22$\\
		\hline
		\Pattern, $p{\approx}1.22$ & 9.52 & 7.20\\
		\Pattern, $p{\approx}0.39$ & 5.77 & 67.87\\
		\Pattern[$0.15$], $p{\approx}1.22$ & 7.67 & 6.52\\
		\Pattern[$0.15$], $p{\approx}0.39$ & 5.94 & 30.96\\
		\hline
		\hline
		\textbf{\emph{Profiled} Bit Errors (Chip 2)} & $p{\approx}0.1$ & $p{\approx}0.63$\\
		\hline
		\Pattern $p{\approx}0.63$ &  85.84 & 10.76\\
		\Pattern, $p{\approx}0.1$ & 90.56 & 5.93\\
		\Pattern[$0.15$] $p{\approx}0.63$ & 12.02 & 8.70\\
		\Pattern[$0.15$] $p{\approx}0.1$ & 90.68 & 6.51\\
		\hline
	\end{tabular}
	\vspace*{-0.2cm}
\end{table}

\revision{Following \algref{alg:training}, \Random adds an additional forward and backward pass during training, increasing training complexity roughly by a factor of two. In practice, however, we found that training time for \Random (in comparison with \Clipping) roughly triples. This is due to our custom implementation of bit error injection, which was not optimized for speed. However, we believe that training time can be reduced significantly using an efficient CUDA implementation of bit error injection. We also note that inference time remains unchanged. In this respect, bit error mitigation strategies in hardware are clearly less desirable due to increased inference time, space and energy consumption.}

\subsection{Profiled Bit Errors}
\label{subsec:supp-randbet-baselines}

Following the evaluation on profiled bit errors outlined in \appref{subsec:supp-errors-profiled}, \tabref{tab:supp-randbet-generalization} shows complementary results for \Clipping[$0.05$] and \Random[$0.05$] trained with $p = 1.5\%$ on all profiled chips. Note that for particularly extreme cases, such as chip 3, \Clipping might perform slightly better than \Random. Overall, however, \Random generalizes reasonably well, with very good results on chip 1 which is closest to our bit error model. Results on chip 2 and 3, due to bit errors being strongly aligned along columns (\cf \figref{fig:supp-errors}), are slightly worse. However, \Random does not fail catastrophically. Instead, \RTE degrades slowly.

\begin{table}[t]
	\centering
	\small
	\caption{\textbf{Results for Probabilistic Guarantees.}. Average \RTE and standard deviation for $l = 1\text{Mio}$ random bit error patterns. In comparison with the results for $l = 50$ from the main paper, there are no significant changes in \RTE. However, standard deviation increases slightly, from $0.11$ to $0.15$ against \Random.}
	\label{tab:supp-stress}
	\vspace*{-0.25cm}
	\hspace*{-0.15cm}
	\begin{tabular}{|l | c | c | c |}
		\hline
		\multicolumn{4}{|c|}{\bfseries\CifarT: Stress Test for Guarantees}\\
		\hline
		Model & \multirow{2}{*}{\begin{tabular}{@{}c@{}}\TE\\in \%\end{tabular}} & \multicolumn{2}{c|}{\RTE in \%, $p = 1\%$}\\
		\hline
		(\CifarT) && $l=50$ & {\color{red}$\mathbf{l=1\text{Mio}}$}\\
		\hline
		\hline
		\Quant & 4.32 & 32.05 {\color{gray}\scriptsize ${\pm}$6} & 31.97 {\color{gray}\scriptsize ${\pm}$6.35}\\
		\Clipping[$0.05$] & 5.44 & 7.18 {\color{gray}\scriptsize ${\pm}$0.16} & 7.19 {\color{gray}\scriptsize ${\pm}$0.2}\\
		\Random[$0.05$] $p{=}2$ & 5.42 & 6.71 {\color{gray}\scriptsize ${\pm}$0.11} & \bfseries 6.73 {\color{gray}\scriptsize ${\pm}$0.15}\\
		\hline
	\end{tabular}
	\vspace*{-0.2cm}
\end{table}
\begin{figure*}[t]
	\centering
	\vspace*{-0.1cm}
	\begin{subfigure}{0.26\textwidth}
		\includegraphics[height=3.075cm]{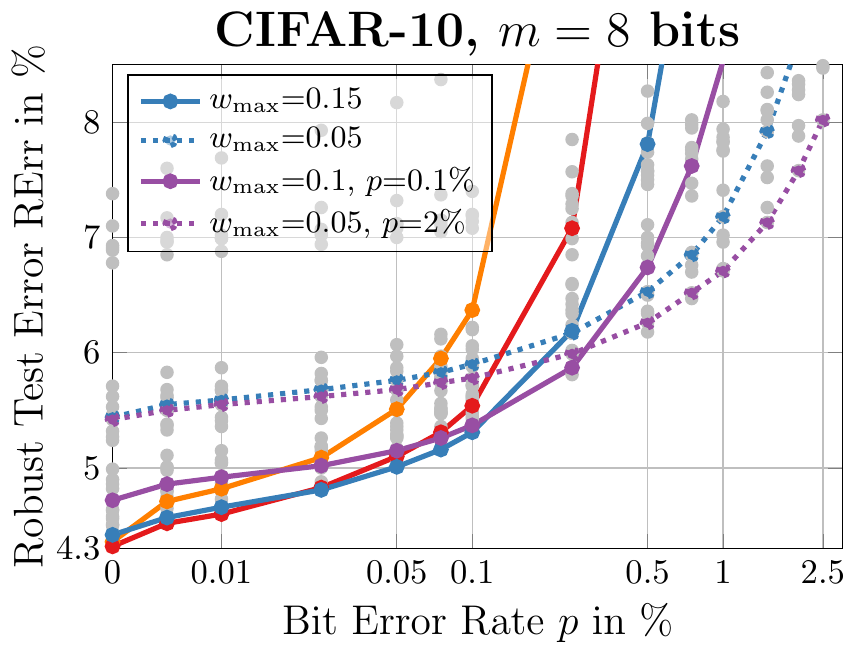}
	\end{subfigure}
	\begin{subfigure}{0.24\textwidth}
		\includegraphics[height=3.1cm]{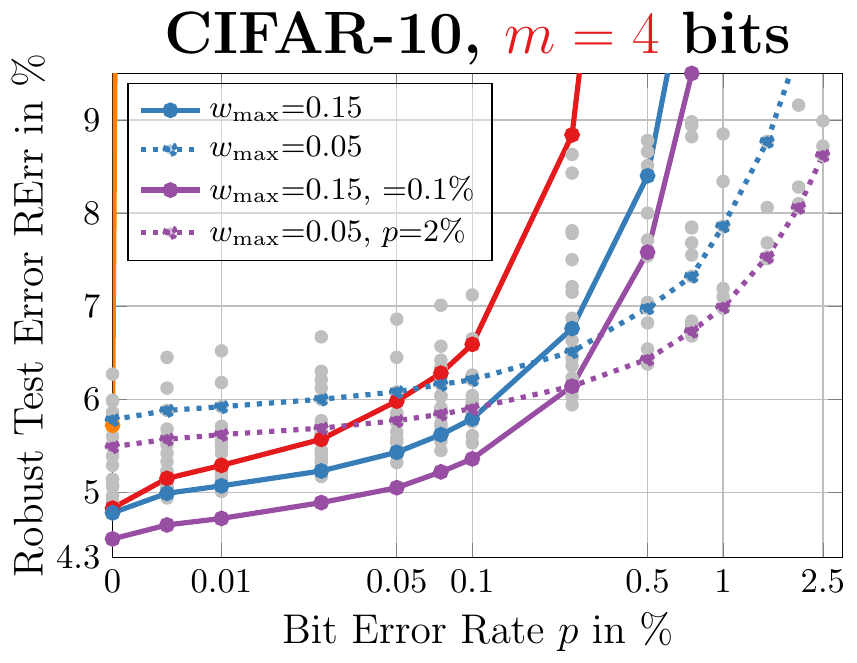}
	\end{subfigure}
	\begin{subfigure}{0.24\textwidth}
		\includegraphics[height=3.1cm]{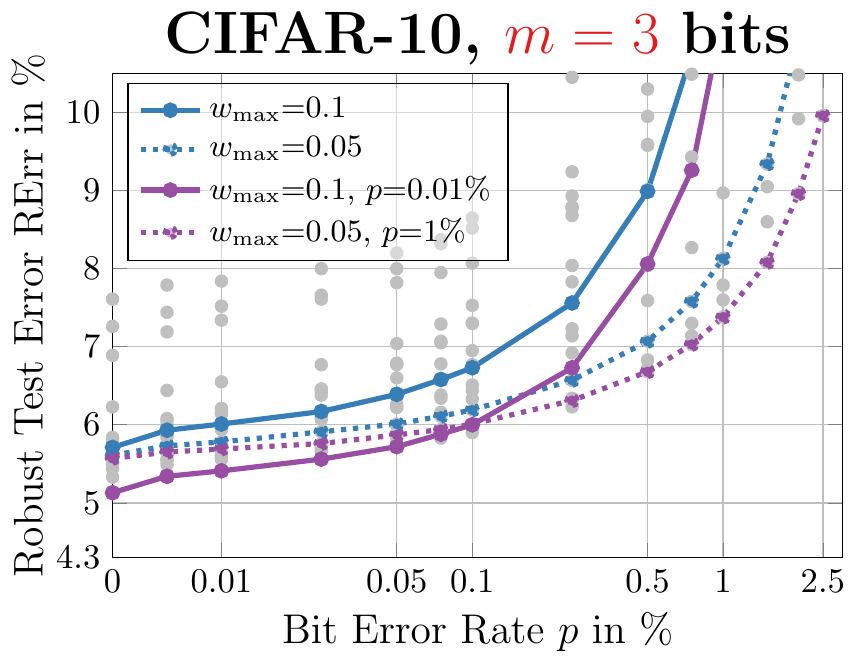}
	\end{subfigure}
	\begin{subfigure}{0.24\textwidth}
		\includegraphics[height=3.1cm]{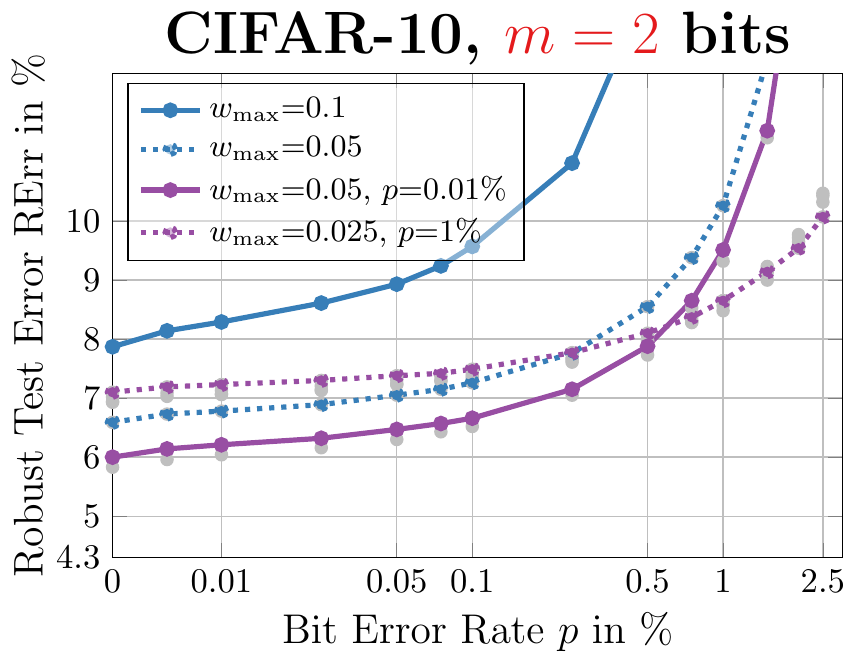}
	\end{subfigure}
	\\[2px]
	
	\begin{subfigure}{0.26\textwidth}
		\includegraphics[height=3.1cm]{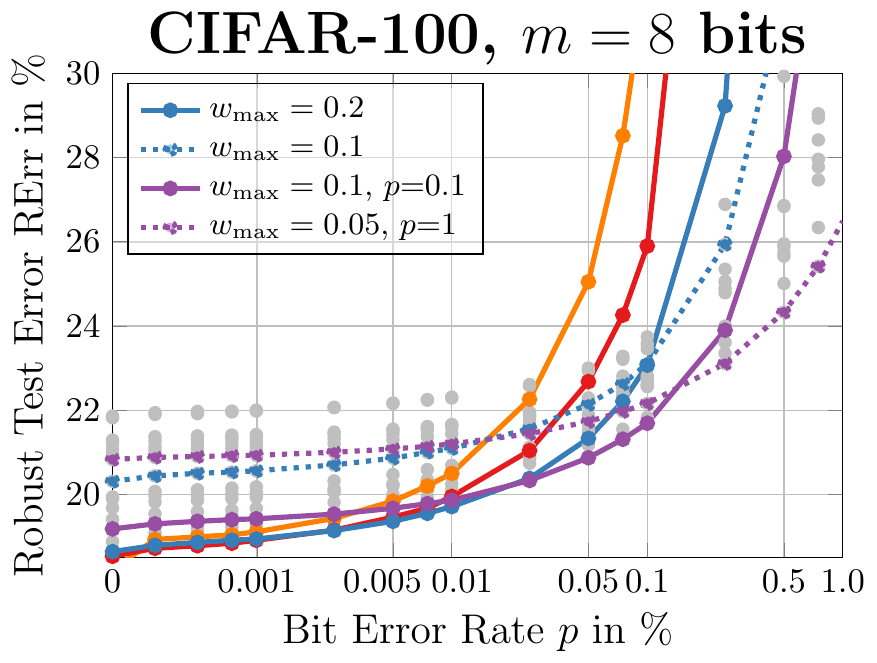}
	\end{subfigure}
	\begin{subfigure}{0.24\textwidth}
		\includegraphics[height=3.1cm]{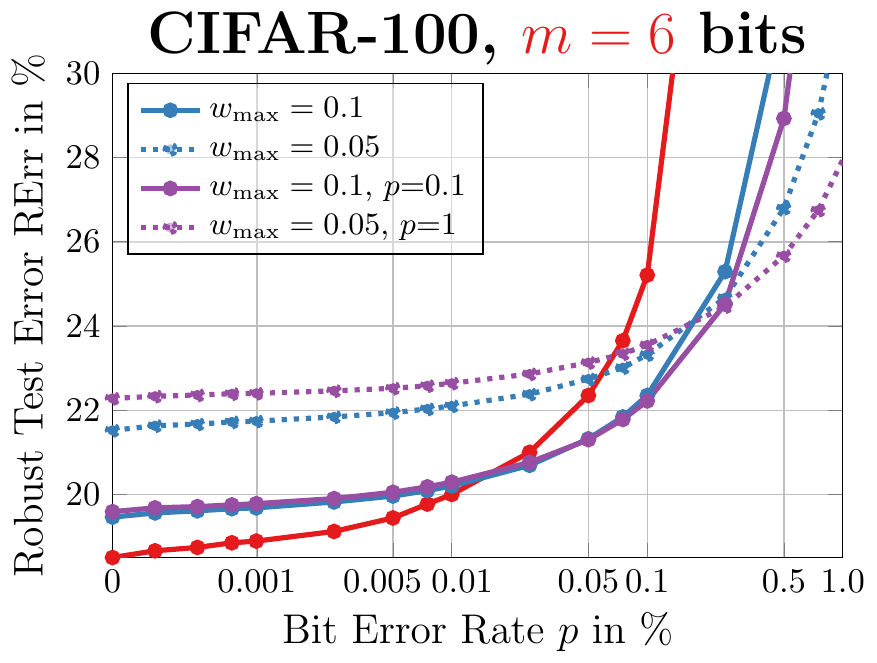}
	\end{subfigure}
	\begin{subfigure}{0.24\textwidth}
		\includegraphics[height=3.1cm]{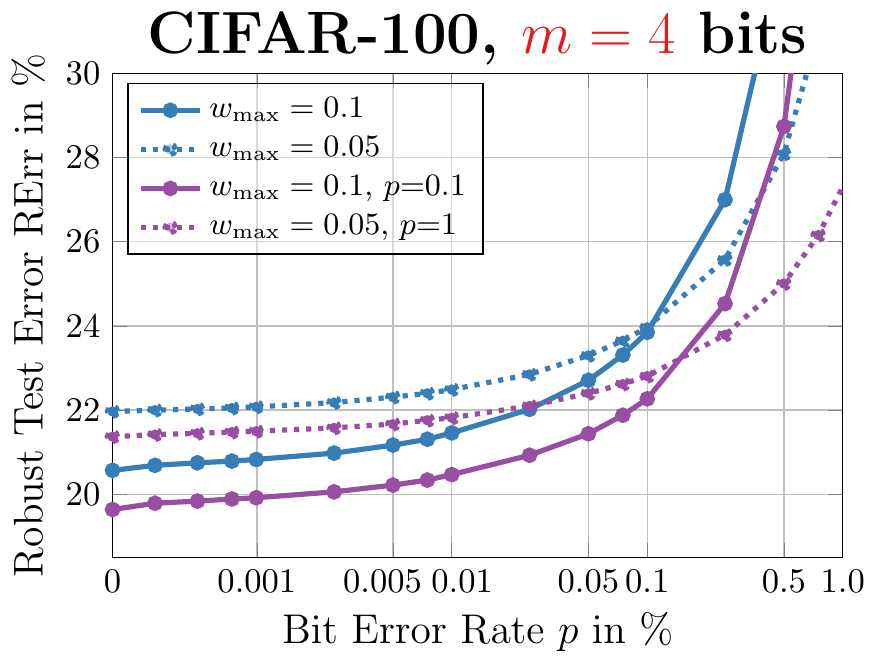} 
	\end{subfigure}
	\begin{subfigure}{0.24\textwidth}
		\hphantom{\includegraphics[height=3.45cm]{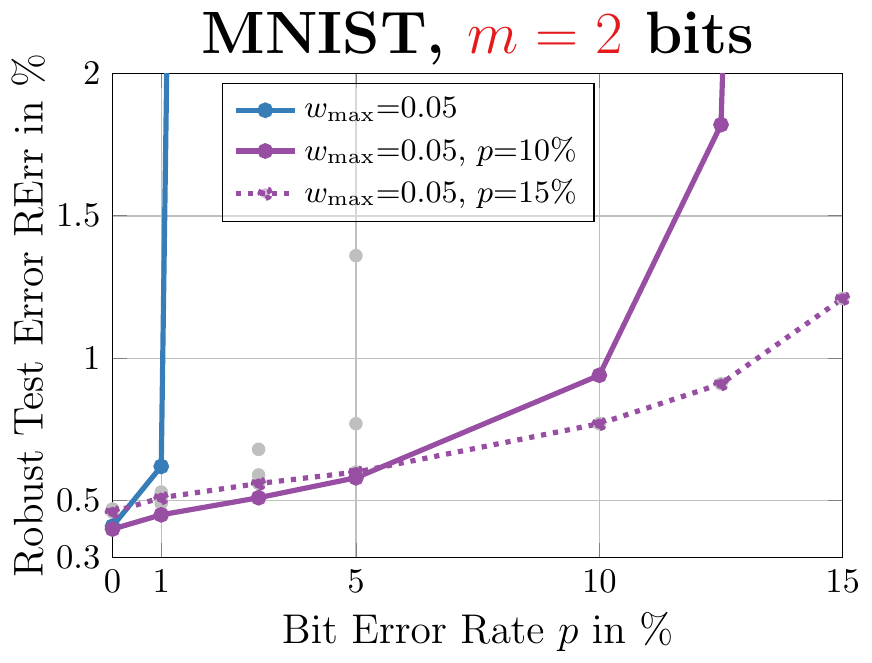}}
	\end{subfigure}
	\\[2px]
	
	\begin{subfigure}{0.26\textwidth}
		\includegraphics[height=3.1cm]{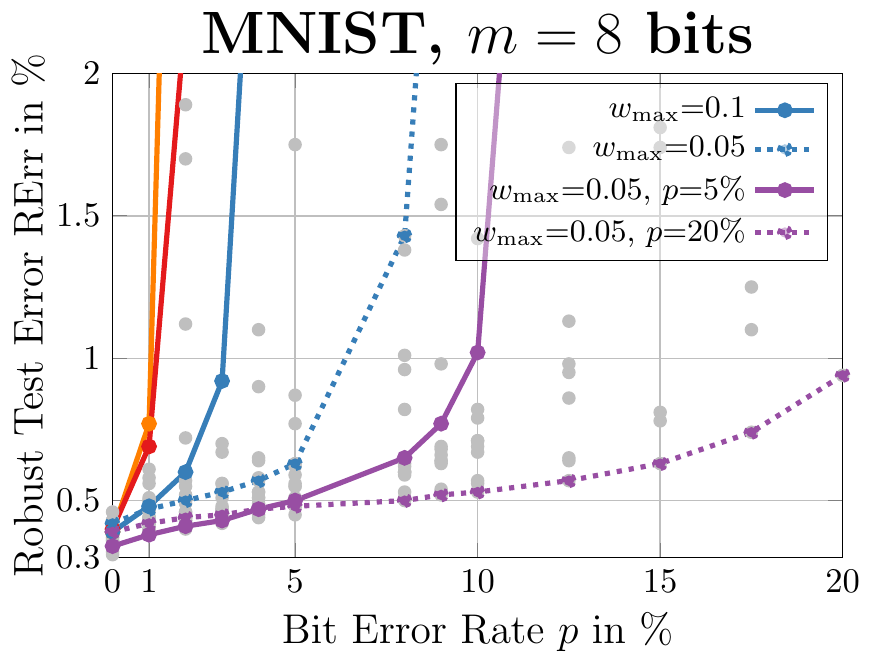}
	\end{subfigure}
	\begin{subfigure}{0.24\textwidth}
		\includegraphics[height=3.1cm]{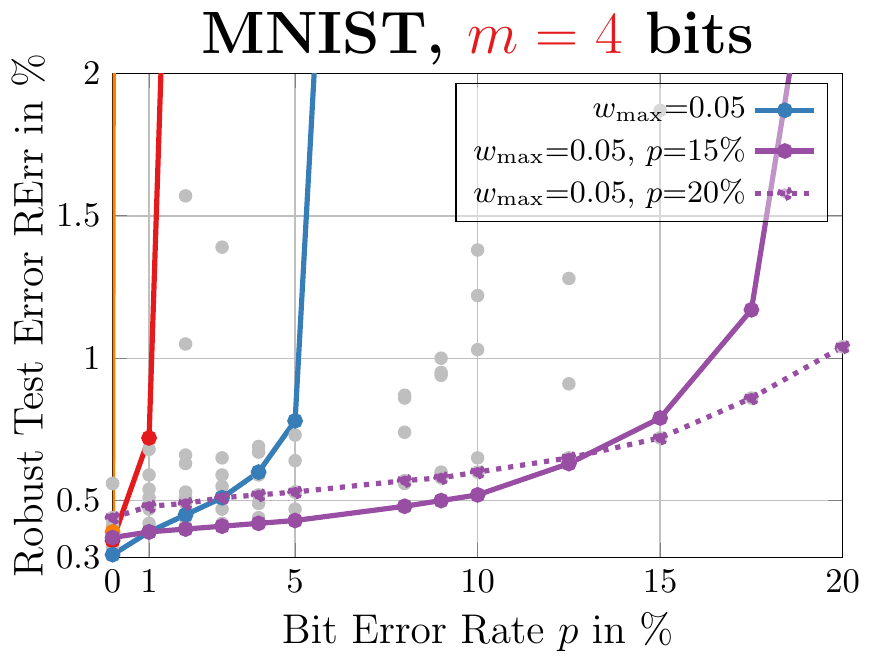}
	\end{subfigure}
	\begin{subfigure}{0.24\textwidth}
		\includegraphics[height=3.1cm]{m_summary_2bit.pdf}
	\end{subfigure}
	\begin{subfigure}{0.24\textwidth}
		\hphantom{\includegraphics[height=3.45cm]{m_summary_2bit.pdf}}
	\end{subfigure}

	\hspace*{-0.1cm}
	\fbox{
	\begin{subfigure}{0.98\textwidth}
		\centering 
		\includegraphics[width=0.8\textwidth]{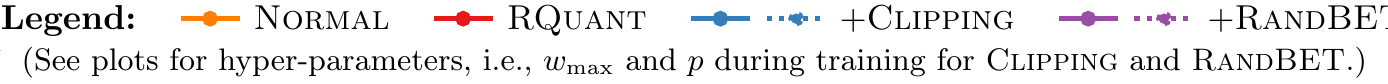}
	\end{subfigure}
	}
	\vspace*{-8px}
	\caption{\textbf{Summary Results on \CifarT, \CifarH and \MNIST.} Complementary to \figref{fig:summary}, we highlight individual \Clipping and \Random models. Note that \figref{fig:summary}, in contrast, presents the best, \ie, lowest \RTE, model for each bit error rate $p$ individually. Instead, individual models help to illustrate the involved trade-offs: \Clipping with small $\wmax$ or \Random with high bit error rate $p$ increases the clean \TE, thereby also increasing \RTE for very small bit error rates. However, \RTE against large bit error rates can be reduced significantly.}
	\label{fig:supp-summary}
\end{figure*}

In \tabref{tab:supp-randbet-baselines}, we follow the procedure of \appref{subsec:supp-errors-profiled} considering only persistent bit errors (\ie, where $p_{\text{1t0}}$ and $p_{\text{0t1}}$ are $1$). This is illustrated in \figref{fig:supp-errors} (right). Thus, the bit error rates deviate slightly from those reported in \tabref{tab:supp-randbet-generalization}, see the table in \appref{subsec:supp-errors-profiled} for details. Furthermore, We consider only one weight-to-SRAM mapping, \ie, without offset. \Pattern is trained and evaluated on the exact same bit error pattern, but potentially with different bit error rates $p$. Note that the bit errors for $p' < p$ are a subset of those for bit error rate $p$. Thus, it is surprising that, on both chips 1 and 2, \Pattern trained on higher bit error rates does not even generalize to lower bit error rates (\ie, higher voltage). This is problematic in practice as the DNN accelerator should not perform worse when increasing voltage.

\subsection{Guarantees from Prop. \ref{prop:bound}}
\label{subsec:experiments-stress}

Based on the bound derived in \secref{subsec:supp-bound}, we conduct experiments with $l = 1\text{Mio}$ random bit error patterns, such that $l \gg n$ where $n = \text{10k}$ is the number of test examples on \CifarT. Considering Prop. \ref{prop:bound}, this would guarantee a deviation in \RTE of at most $4.1\%$ with probability at least $99\%$. As shown in \tabref{tab:supp-stress}, the obtained \RTE with $1\text{Mio}$ random bit error patterns deviates insignificantly from the results in the main paper. Only standard deviation of \RTE increases slightly. These results emphasize that the results for \Clipping and \Random from the main paper generalize well.

\subsection{Other Architectures}
\label{subsec:supp-experiments-architectures}

\tabref{tab:supp-randbet-resnet} shows results on \CifarT using ResNet-20 and ResNet-50. We note that, in both cases, we use group normalization (GN) instead of batch normalization (BN) as outlined in \secref{subsec:supp-experiments-bn}. ResNet-50, in particular, suffers from using GN due to the significant depth: the clean \TE reduces from $3.67\%$ to $6.81\%$ in \tabref{tab:supp-accuracy}. Nevertheless, \Clipping and \Random remain effective against random bit errors, even for higher bit error rates of $p = 1.5\%$. This is striking as ResNet-50 consists of roughly $23.5\text{Mio}$ weights, compared to $5.5\text{Mio}$ of the used SimpleNet in the main paper.

\subsection{Summary Results}
\label{subsec:supp-experiments-summary}

\figref{fig:supp-summary} summarizes our results: In contrast to \figref{fig:summary}, we consider individual \Clipping and \Random models instead of focusing on the best results per bit error rate $p$. Additionally, we show our complete results for lower precisions, \ie, $m = 4,3,2$ on \CifarT and \MNIST. Note that these results, in tabular form, are included in \tabref{tab:supp-summary-cifar10} to \ref{tab:supp-summary-mnist}. Moderate \Clipping, \eg, using $\wmax = 0.15$ on \CifarT (in {\color{colorbrewer1}red}), has negligible impact on clean \TE (\ie, $p = 0$ on the x-axis) while improving robustness beyond $p = 0.1\%$ bit error rate. Generally, however, higher robustness is obtained at the cost of increased clean \TE, \eg, for $\wmax = 0.05$ (in {\color{colorbrewer2}blue}). Here, it is important to note that in low-voltage operation, only \RTE matters -- clean \TE is only relevant for voltages higher than \Vmin. \Random further improves robustness for high bit error rates, while continuing to increase clean \TE slightly. For example, \Random with $\wmax = 0.05$ and trained with $p = 2\%$ bit errors increases clean \TE to $5.42\%$ but is also able to keep \RTE below $7\%$ up to $p = 1\%$ bit error rate (in {\color{colorbrewer5}orange}). Reducing precision generally increases \TE and \RTE, especially for $m = 2$ bit. Here, our simple fixed-point quantization scheme is clearly limited compared to state-of-the-art. Nevertheless, even for $m = 2$ bits, \Random ({\color{colorbrewer4}violet} or {\color{colorbrewer5}orange}) is able to keep \RTE low until roughly $p = 0.1\%$ bit error rate. Note that for $m = 2$, more aggressive clipping generally helps during training and, thus, also reduces clean \TE (\cf $\wmax = 0.1$ and $\wmax = 0.05$ in {\color{colorbrewer1}red} and {\color{colorbrewer2}blue}).

Similar trade-offs can be observed on \CifarH and \MNIST. On \CifarH, we see that task difficulty also reduces the bit error rate that is tolerable without significant increase it \RTE. Here, $p = 0.1\%$ increases \RTE by more than $3\%$, even with \Random (and weight clipping). Furthermore, \CifarH demonstrates that \Clipping and \Random are applicable to significantly larger architectures such as Wide ResNets without problems. On \MNIST, in contrast, bit error rates of up to $p = 20\%$ are easily possible. At such bit error rates, the benefit of \Random is extremely significant as even \Clipping[$0.025$] exhibits very high \RTE of $32.68\%$ at $p = 20\%$, \cf \tabref{tab:supp-summary-mnist}.

\clearpage
\begin{table*}
	\centering
	\small
	\caption{\textbf{Overall Robustness Results on \CifarT.} Tabular results corresponding to \figref{fig:summary} and \ref{fig:supp-summary} for $m = 8$ and $m = 4$ bits. We show \RTE for \Normal, \Clipping and \Random with various $\wmax$ and $p$ across a subset of test bit error rates.}
	\label{tab:supp-summary-cifar10}
	\vspace*{-0.2cm}
	\begin{tabular}{| c | l | c | c | c | c | c | c | c | c | c |}
		\hline
		\multicolumn{11}{|c|}{\bfseries \CifarT: summary results for $\mathbf{m = 8}$ and  {\color{red}$\mathbf{m = 4}$} bit}\\
		\hline
		& Model & \multirow{2}{*}{\begin{tabular}{c}\TE\\in \%\end{tabular}} & \multicolumn{8}{c|}{\RTE in \%, $p$ in \% p=0.01}\\
		\cline{4-11}
		&&& $0.01$ & $0.05$ & $0.1$ & $0.5$ & $1$ & $1.5$ & $2$ & $2.5$\\
		\hline
		\hline
		\multirow{30}{*}{\rotatebox{90}{$m = 8$ bit}} & \Normal & 4.36 & 4.82 & 5.51 & 6.37 & 24.76 & 72.65 & 87.40 & 89.76 & 90.15\\
		& \Quant & 4.32 & 4.60 & 5.10 & 5.54 & 11.28 & 32.05 & 68.65 & 85.28 & 89.01\\
		& \Clipping[$0.25$] & 4.58 & 4.84 & 5.29 & 5.71 & 10.52 & 27.95 & 62.46 & 82.61 & 88.08\\
		& \Clipping[$0.2$] & 4.63 & 4.91 & 5.28 & 5.62 & 8.27 & 18.00 & 53.74 & 82.02 & 88.27\\
		& \Clipping[$0.15$] & 4.42 & 4.66 & 5.01 & 5.31 & 7.81 & 13.08 & 23.85 & 42.12 & 61.20\\
		& \Clipping[$0.1$] & 4.82 & 5.04 & 5.33 & 5.58 & 6.95 & 8.93 & 12.22 & 17.80 & 27.02\\
		& \Clipping[$0.05$] & 5.44 & 5.59 & 5.76 & 5.90 & 6.53 & 7.18 & 7.92 & 8.70 & 9.56\\
		& \Clipping[$0.025$] & 7.10 & 7.20 & 7.32 & 7.40 & 7.82 & 8.18 & 8.43 & 8.74 & --\\
		\cline{2-11}
		& \Random[$1$] $p{=}0.01$ & 4.56 & 4.93 & 5.50 & 6.06 & 14.14 & 66.07 & 86.86 & 89.80 & 90.35\\
		& \Random[$1$] $p{=}0.1$ & 4.50 & 4.80 & 5.27 & 5.72 & 10.33 & 41.10 & 75.90 & 86.52 & 89.03\\
		& \Random[$1$] $p{=}1$ & 7.38 & 7.69 & 8.17 & 8.58 & 11.10 & 14.90 & 21.08 & 41.11 & 71.09\\
		& \Random[$0.2$] $p{=}0.01$ & 4.44 & 4.67 & 5.09 & 5.48 & 8.64 & 17.97 & 41.53 & 68.95 & 82.48\\
		& \Random[$0.2$] $p{=}0.1$ & 4.51 & 4.73 & 5.07 & 5.39 & 7.99 & 19.21 & 54.94 & 80.12 & 86.55\\
		& \Random[$0.2$] $p{=}1$ & 5.46 & 5.68 & 5.97 & 6.20 & 7.63 & 9.47 & 12.38 & 21.47 & 50.86\\
		& \Random[$0.15$] $p{=}0.01$ & 4.64 & 4.87 & 5.17 & 5.45 & 7.54 & 15.83 & 54.07 & 81.41 & 86.75\\
		& \Random[$0.15$] $p{=}0.1$ & 4.86 & 5.07 & 5.36 & 5.64 & 7.74 & 12.33 & 22.38 & 40.09 & 60.78\\
		& \Random[$0.15$] $p{=}1$ & 5.27 & 5.44 & 5.68 & 5.88 & 7.11 & 8.63 & 11.13 & 27.74 & 64.97\\
		& \Random[$0.1$] $p{=}0.01$ & 4.99 & 5.15 & 5.39 & 5.62 & 6.93 & 9.01 & 12.83 & 22.81 & 41.04\\
		& \Random[$0.1$] $p{=}0.1$ & 4.72 & 4.92 & 5.15 & 5.37 & 6.74 & 8.53 & 11.40 & 15.97 & 23.59\\
		& \Random[$0.1$] $p{=}1$ & 4.90 & 5.05 & 5.26 & 5.43 & 6.36 & 7.41 & 8.65 & 12.25 & 27.21\\
		& \Random[$0.1$] $p{=}1.5$ & 5.53 & 5.67 & 5.87 & 6.03 & 6.84 & 7.76 & 8.80 & 10.03 & 11.68\\
		& \Random[$0.1$] $p{=}2$ & 5.71 & 5.87 & 6.07 & 6.22 & 7.00 & 7.83 & 8.69 & 9.70 & 10.91\\
		& \Random[$0.05$] $p{=}0.1$ & 5.32 & 5.41 & 5.59 & 5.72 & 6.34 & 6.96 & 7.62 & 8.28 & 9.13\\
		& \Random[$0.05$] $p{=}1$ & 5.24 & 5.36 & 5.50 & 5.60 & 6.18 & 6.73 & 7.26 & 7.88 & 8.49\\
		& \Random[$0.05$] $p{=}1.5$ & 5.62 & 5.71 & 5.84 & 5.95 & 6.50 & 7.02 & 7.52 & 7.97 & 8.51\\
		& \Random[$0.05$] $p{=}2$ & 5.42 & 5.55 & 5.68 & 5.78 & 6.26 & 6.71 & 7.13 & 7.58 & 8.02\\
		& \Random[$0.025$] $p{=}1$ & 6.78 & 6.88 & 7.00 & 7.08 & 7.46 & 7.75 & 8.02 & 8.24 & 8.47\\
		& \Random[$0.025$] $p{=}1.5$ & 6.89 & 6.99 & 7.11 & 7.19 & 7.58 & 7.94 & 8.26 & 8.52 & 8.77\\
		& \Random[$0.025$] $p{=}2$ & 6.93 & 7.02 & 7.12 & 7.20 & 7.57 & 7.87 & 8.11 & 8.33 & 8.58\\
		& \Random[$0.025$] $p{=}2.5$ & 6.91 & 6.99 & 7.08 & 7.14 & 7.50 & 7.83 & 8.10 & 8.36 & 8.63\\
		\hline
		\hline
		\multirow{19}{*}{\rotatebox{90}{$m = 4$ bit}} & \Quant & 4.83 & 5.29 & 5.98 & 6.59 & 15.72 & 50.45 & 79.86 & 87.17 & 89.47\\
		& \Clipping[$0.25$] & 4.78 & 5.16 & 5.75 & 6.26 & 12.08 & 30.62 & 60.52 & 80.07 & 87.01\\
		& \Clipping[$0.2$] & 4.90 & 5.20 & 5.65 & 6.04 & 9.67 & 27.24 & 63.96 & 82.63 & 87.21\\
		& \Clipping[$0.15$] & 4.78 & 5.07 & 5.43 & 5.79 & 8.40 & 14.61 & 28.53 & 50.83 & 70.32\\
		& \Clipping[$0.1$] & 5.29 & 5.49 & 5.75 & 5.99 & 7.71 & 10.62 & 15.79 & 24.97 & 37.94\\
		& \Clipping[$0.05$] & 5.78 & 5.92 & 6.08 & 6.21 & 6.98 & 7.86 & 8.77 & 9.76 & 11.04\\
		\cline{2-11}
		& \Random[$0.2$] $p{=}0.01$ & 5.14 & 5.42 & 5.85 & 6.23 & 10.44 & 23.84 & 49.25 & 73.35 & 83.16\\
		& \Random[$0.2$] $p{=}0.1$ & 4.77 & 5.01 & 5.41 & 5.76 & 8.66 & 16.06 & 32.40 & 56.69 & 75.21\\
		& \Random[$0.2$] $p{=}1$ & 6.27 & 6.52 & 6.86 & 7.12 & 8.78 & 11.33 & 15.17 & 21.43 & 32.19\\
		& \Random[$0.15$] $p{=}0.01$ & 4.88 & 5.13 & 5.54 & 5.92 & 8.51 & 14.21 & 26.26 & 46.02 & 66.13\\
		& \Random[$0.15$] $p{=}0.1$ & 4.50 & 4.72 & 5.05 & 5.36 & 7.58 & 14.12 & 43.00 & 76.28 & 85.54\\
		& \Random[$0.15$] $p{=}1$ & 5.99 & 6.18 & 6.45 & 6.65 & 8.00 & 9.74 & 12.50 & 16.73 & 24.09\\
		& \Random[$0.1$] $p{=}0.01$ & 5.07 & 5.29 & 5.58 & 5.83 & 7.54 & 10.46 & 15.34 & 24.63 & 39.76\\
		& \Random[$0.1$] $p{=}0.1$ & 4.82 & 5.04 & 5.32 & 5.53 & 6.82 & 8.85 & 12.48 & 21.36 & 40.03\\
		& \Random[$0.1$] $p{=}1$ & 5.39 & 5.55 & 5.77 & 5.96 & 7.04 & 8.34 & 9.77 & 11.85 & 14.91\\
		& \Random[$0.05$] $p{=}0.1$ & 5.14 & 5.26 & 5.46 & 5.61 & 6.38 & 7.19 & 8.06 & 9.16 & 10.46\\
		& \Random[$0.05$] $p{=}1$ & 5.60 & 5.71 & 5.85 & 5.97 & 6.54 & 7.10 & 7.68 & 8.28 & 8.99\\
		& \Random[$0.05$] $p{=}1.5$ & 5.51 & 5.64 & 5.77 & 5.87 & 6.38 & 6.98 & 7.51 & 8.10 & 8.72\\
		& \Random[$0.05$] $p{=}2$ & 5.49 & 5.62 & 5.77 & 5.90 & 6.43 & 6.99 & 7.53 & 8.06 & 8.62\\
		\hline
	\end{tabular}
	\vspace*{-0.2cm}
\end{table*}
\begin{table*}
	\centering
	\small
	\caption{\textbf{Overall Robustness Results on \CifarT.} \textbf{Overall Robustness Results on \CifarT.} Continued from \tabref{tab:supp-summary-cifar10}; tabular results corresponding to \figref{fig:summary} and \ref{fig:supp-summary} for $m = 3$ and $m = 2$ bits. We show \RTE for \Normal, \Clipping and \Random with various $\wmax$ and $p$ across a subset of test bit error rates.}
	\vspace*{-0.2cm}
	\begin{tabular}{| c | l | c | c | c | c | c | c | c | c | c |}
		\hline
		\multicolumn{11}{|c|}{\bfseries \CifarT: summary results for {\color{red}$\mathbf{m = 3}$} and {\color{red}$\mathbf{m = 2}$} bit}\\
		\hline
		& Model & \multirow{2}{*}{\begin{tabular}{c}\TE\\in \%\end{tabular}} & \multicolumn{8}{c|}{\RTE in \%, $p$ in \% p=0.01}\\
		\cline{4-11}
		&&& $0.01$ & $0.05$ & $0.1$ & $0.5$ & $1$ & $1.5$ & $2$ & $2.5$\\
		\hline
		\hline
		\multirow{20}{*}{\rotatebox{90}{$m = 3$ bit}} & \Quant & 79.59 & 83.95 & 88.57 & 91.07 & 96.15 & 97.81 & 98.20 & 98.60 & 99.07\\
		& \Clipping[$0.25$] & 6.89 & 7.34 & 8.00 & 8.65 & 14.46 & 28.70 & 53.64 & 75.51 & 85.13\\
		& \Clipping[$0.2$] & 5.82 & 6.21 & 6.79 & 7.30 & 11.90 & 23.31 & 43.00 & 65.68 & 78.79\\
		& \Clipping[$0.15$] & 5.84 & 6.16 & 6.60 & 6.95 & 9.95 & 15.92 & 27.84 & 47.54 & 67.08\\
		& \Clipping[$0.1$] & 5.71 & 6.01 & 6.39 & 6.73 & 8.99 & 13.06 & 20.88 & 35.13 & 51.76\\
		& \Clipping[$0.05$] & 5.61 & 5.78 & 6.01 & 6.19 & 7.07 & 8.13 & 9.34 & 10.95 & 13.16\\
		\cline{2-11}
		& \Random[$0.2$] $p{=}0.01$ & 5.72 & 6.14 & 6.77 & 7.30 & 12.84 & 26.46 & 50.52 & 72.46 & 83.09\\
		& \Random[$0.2$] $p{=}0.1$ & 6.23 & 6.55 & 7.04 & 7.53 & 11.38 & 21.36 & 41.93 & 65.54 & 79.94\\
		& \Random[$0.2$] $p{=}1$ & 7.61 & 7.84 & 8.20 & 8.52 & 10.30 & 12.82 & 16.65 & 21.81 & 29.64\\
		& \Random[$0.15$] $p{=}0.01$ & 5.61 & 5.94 & 6.40 & 6.77 & 9.59 & 15.72 & 28.06 & 46.88 & 64.39\\
		& \Random[$0.15$] $p{=}0.1$ & 5.33 & 5.56 & 5.99 & 6.33 & 9.01 & 14.06 & 23.44 & 40.36 & 59.92\\
		& \Random[$0.15$] $p{=}1$ & 7.26 & 7.52 & 7.82 & 8.07 & 9.58 & 11.47 & 13.87 & 17.58 & 23.01\\
		& \Random[$0.1$] $p{=}0.01$ & 5.13 & 5.41 & 5.72 & 6.00 & 8.06 & 11.25 & 17.22 & 26.96 & 42.72\\
		& \Random[$0.1$] $p{=}0.1$ & 5.69 & 5.96 & 6.26 & 6.51 & 8.04 & 10.81 & 15.51 & 23.88 & 37.52\\
		& \Random[$0.1$] $p{=}1$ & 5.76 & 5.95 & 6.22 & 6.44 & 7.59 & 8.97 & 10.76 & 13.21 & 16.95\\
		& \Random[$0.05$] $p{=}0.01$ & 5.50 & 5.62 & 5.83 & 5.99 & 6.83 & 7.79 & 9.05 & 10.48 & 12.32\\
		& \Random[$0.05$] $p{=}0.1$ & 5.44 & 5.58 & 5.76 & 5.90 & 6.72 & 7.60 & 8.60 & 9.92 & 11.70\\
		& \Random[$0.05$] $p{=}1$ & 5.57 & 5.69 & 5.87 & 6.01 & 6.68 & 7.38 & 8.08 & 8.96 & 9.96\\
		\hline
		\hline
		\multirow{11}{*}{\rotatebox{90}{$m = 2$ bit}} & \Quant & 88.68 & 89.53 & 91.62 & 93.23 & 97.74 & 98.40 & 97.85 & 99.20 & 98.74\\
		& \Clipping[$0.25$] & 90.14 & 90.54 & 91.13 & 91.82 & 95.96 & 96.90 & 97.21 & 96.66 & 97.12\\
		& \Clipping[$0.2$] & 82.00 & 84.86 & 90.79 & 94.17 & 97.25 & 96.69 & 97.16 & 97.73 & 97.01\\
		& \Clipping[$0.15$] & 14.62 & 15.29 & 16.30 & 17.16 & 22.88 & 33.18 & 50.86 & 71.17 & 84.30\\
		& \Clipping[$0.1$] & 7.87 & 8.29 & 8.93 & 9.57 & 13.95 & 23.65 & 42.43 & 64.65 & 80.89\\
		& \Clipping[$0.05$] & 6.59 & 6.78 & 7.05 & 7.26 & 8.55 & 10.26 & 12.73 & 15.99 & 20.51\\
		& \Clipping[$0.025$] & 6.94 & 7.06 & 7.23 & 7.34 & 7.96 & 8.57 & 9.16 & 9.77 & 10.47\\
		& \Random[$0.05$] $p{=}0.01$ & 6.00 & 6.21 & 6.47 & 6.66 & 7.88 & 9.51 & 11.53 & 14.99 & 19.60\\
		& \Random[$0.05$] $p{=}0.1$ & 5.83 & 6.04 & 6.30 & 6.52 & 7.73 & 9.32 & 11.41 & 14.49 & 19.77\\
		& \Random[$0.025$] $p{=}0.01$ & 6.93 & 7.07 & 7.24 & 7.37 & 8.05 & 8.65 & 9.23 & 9.72 & 10.43\\
		& \Random[$0.025$] $p{=}0.1$ & 7.02 & 7.13 & 7.31 & 7.41 & 7.98 & 8.48 & 9.00 & 9.65 & 10.32\\
		& \Random[$0.025$] $p{=}1$ & 7.10 & 7.23 & 7.38 & 7.49 & 8.10 & 8.65 & 9.14 & 9.54 & 10.07\\
		\hline
	\end{tabular}
\end{table*}
\begin{table*}
	\centering
	\small
	\caption{\textbf{Overall Robustness Results on \CifarH.} Tabular results corresponding to \figref{fig:summary} and \ref{fig:supp-summary} for $m = 8$. We show \RTE for \Normal, \Clipping and \Random with various $\wmax$ and $p$ across a subset of test bit error rates.}
	\label{tab:supp-summary-cifar100}
	\vspace*{-0.2cm}
	\begin{tabular}{|l | c | c | c | c | c | c | c |}
		\hline
		\multicolumn{8}{|c|}{\bfseries \CifarH: summary results for $\mathbf{m = 8}$ bit}\\
		\hline
		Model & \multirow{2}{*}{\begin{tabular}{c}\TE\\in \%\end{tabular}} & \multicolumn{6}{c|}{\RTE in \%, $p$ in \% p=0.01}\\
		\cline{3-8}
		&& $0.005$ & $0.01$ & $0.05$ & $0.1$ & $0.5$ & $1$\\
		\hline
		\hline
		\Normal & 18.21 & 19.84 {\color{gray}\scriptsize ${\pm}$0.16} & 20.50 {\color{gray}\scriptsize ${\pm}$0.25} & 25.05 {\color{gray}\scriptsize ${\pm}$0.94} & 32.39 {\color{gray}\scriptsize ${\pm}$1.89} & 97.49 {\color{gray}\scriptsize ${\pm}$0.95} & 99.10 {\color{gray}\scriptsize ${\pm}$0.19}\\
		\Quant & 18.53 & 19.46 {\color{gray}\scriptsize ${\pm}$0.13} & 19.95 {\color{gray}\scriptsize ${\pm}$0.16} & 22.68 {\color{gray}\scriptsize ${\pm}$0.63} & 25.90 {\color{gray}\scriptsize ${\pm}$1.01} & 87.24 {\color{gray}\scriptsize ${\pm}$3.99} & 98.77 {\color{gray}\scriptsize ${\pm}$0.31}\\
		\Clipping[$0.25$] & 18.88 & 19.76 {\color{gray}\scriptsize ${\pm}$0.1} & 20.11 {\color{gray}\scriptsize ${\pm}$0.11} & 21.89 {\color{gray}\scriptsize ${\pm}$0.18} & 23.74 {\color{gray}\scriptsize ${\pm}$0.35} & 62.25 {\color{gray}\scriptsize ${\pm}$4.51} & 96.62 {\color{gray}\scriptsize ${\pm}$1.22}\\
		\Clipping[$0.2$] & 18.64 & 19.36 {\color{gray}\scriptsize ${\pm}$0.09} & 19.71 {\color{gray}\scriptsize ${\pm}$0.1} & 21.33 {\color{gray}\scriptsize ${\pm}$0.23} & 23.07 {\color{gray}\scriptsize ${\pm}$0.38} & 49.79 {\color{gray}\scriptsize ${\pm}$4.21} & 94.02 {\color{gray}\scriptsize ${\pm}$2.38}\\
		\Clipping[$0.15$] & 19.41 & 20.00 {\color{gray}\scriptsize ${\pm}$0.08} & 20.24 {\color{gray}\scriptsize ${\pm}$0.09} & 21.68 {\color{gray}\scriptsize ${\pm}$0.17} & 23.02 {\color{gray}\scriptsize ${\pm}$0.3} & 37.85 {\color{gray}\scriptsize ${\pm}$2.03} & 79.45 {\color{gray}\scriptsize ${\pm}$5.08}\\
		\Clipping[$0.1$] & 20.31 & 20.86 {\color{gray}\scriptsize ${\pm}$0.07} & 21.09 {\color{gray}\scriptsize ${\pm}$0.09} & 22.14 {\color{gray}\scriptsize ${\pm}$0.17} & 23.10 {\color{gray}\scriptsize ${\pm}$0.21} & 31.78 {\color{gray}\scriptsize ${\pm}$1.15} & 51.71 {\color{gray}\scriptsize ${\pm}$3.47}\\
		\Clipping[$0.05$] & 21.82 & 22.16 {\color{gray}\scriptsize ${\pm}$0.05} & 22.29 {\color{gray}\scriptsize ${\pm}$0.06} & 22.94 {\color{gray}\scriptsize ${\pm}$0.13} & 23.46 {\color{gray}\scriptsize ${\pm}$0.18} & 26.86 {\color{gray}\scriptsize ${\pm}$0.46} & 31.47 {\color{gray}\scriptsize ${\pm}$0.79}\\
		\hline
		\Random[$0.1$] $p{=}0.01$ & 19.68 & 20.21 {\color{gray}\scriptsize ${\pm}$0.08} & 20.46 {\color{gray}\scriptsize ${\pm}$0.09} & 21.52 {\color{gray}\scriptsize ${\pm}$0.17} & 22.56 {\color{gray}\scriptsize ${\pm}$0.25} & 30.59 {\color{gray}\scriptsize ${\pm}$0.82} & 48.93 {\color{gray}\scriptsize ${\pm}$3.31}\\
		\Random[$0.1$] $p{=}0.05$ & 19.94 & 20.47 {\color{gray}\scriptsize ${\pm}$0.06} & 20.69 {\color{gray}\scriptsize ${\pm}$0.08} & 21.72 {\color{gray}\scriptsize ${\pm}$0.16} & 22.60 {\color{gray}\scriptsize ${\pm}$0.23} & 29.93 {\color{gray}\scriptsize ${\pm}$0.86} & 46.76 {\color{gray}\scriptsize ${\pm}$3.46}\\
		\Random[$0.1$] $p{=}0.1$ & 19.18 & 19.67 {\color{gray}\scriptsize ${\pm}$0.06} & 19.86 {\color{gray}\scriptsize ${\pm}$0.07} & 20.87 {\color{gray}\scriptsize ${\pm}$0.12} & 21.69 {\color{gray}\scriptsize ${\pm}$0.21} & 28.03 {\color{gray}\scriptsize ${\pm}$0.74} & 41.29 {\color{gray}\scriptsize ${\pm}$2.81}\\
		\Random[$0.1$] $p{=}0.5$ & 19.90 & 20.24 {\color{gray}\scriptsize ${\pm}$0.05} & 20.41 {\color{gray}\scriptsize ${\pm}$0.07} & 21.17 {\color{gray}\scriptsize ${\pm}$0.13} & 21.83 {\color{gray}\scriptsize ${\pm}$0.17} & 25.66 {\color{gray}\scriptsize ${\pm}$0.48} & 31.55 {\color{gray}\scriptsize ${\pm}$0.95}\\
		\Random[$0.1$] $p{=}1$ & 21.08 & 21.43 {\color{gray}\scriptsize ${\pm}$0.05} & 21.59 {\color{gray}\scriptsize ${\pm}$0.07} & 22.24 {\color{gray}\scriptsize ${\pm}$0.13} & 22.76 {\color{gray}\scriptsize ${\pm}$0.15} & 25.73 {\color{gray}\scriptsize ${\pm}$0.33} & 29.31 {\color{gray}\scriptsize ${\pm}$0.56}\\
		\Random[$0.05$] $p{=}0.01$ & 21.86 & 22.17 {\color{gray}\scriptsize ${\pm}$0.06} & 22.31 {\color{gray}\scriptsize ${\pm}$0.05} & 23.00 {\color{gray}\scriptsize ${\pm}$0.14} & 23.57 {\color{gray}\scriptsize ${\pm}$0.2} & 26.84 {\color{gray}\scriptsize ${\pm}$0.46} & 31.33 {\color{gray}\scriptsize ${\pm}$0.79}\\
		\Random[$0.05$] $p{=}0.05$ & 20.97 & 21.30 {\color{gray}\scriptsize ${\pm}$0.05} & 21.44 {\color{gray}\scriptsize ${\pm}$0.07} & 22.12 {\color{gray}\scriptsize ${\pm}$0.14} & 22.72 {\color{gray}\scriptsize ${\pm}$0.16} & 25.95 {\color{gray}\scriptsize ${\pm}$0.34} & 30.14 {\color{gray}\scriptsize ${\pm}$0.59}\\
		\Random[$0.05$] $p{=}0.1$ & 21.22 & 21.53 {\color{gray}\scriptsize ${\pm}$0.05} & 21.66 {\color{gray}\scriptsize ${\pm}$0.05} & 22.29 {\color{gray}\scriptsize ${\pm}$0.12} & 22.81 {\color{gray}\scriptsize ${\pm}$0.17} & 25.88 {\color{gray}\scriptsize ${\pm}$0.39} & 29.93 {\color{gray}\scriptsize ${\pm}$0.83}\\
		\Random[$0.05$] $p{=}0.5$ & 21.29 & 21.55 {\color{gray}\scriptsize ${\pm}$0.04} & 21.65 {\color{gray}\scriptsize ${\pm}$0.06} & 22.13 {\color{gray}\scriptsize ${\pm}$0.12} & 22.60 {\color{gray}\scriptsize ${\pm}$0.15} & 25.01 {\color{gray}\scriptsize ${\pm}$0.3} & 27.70 {\color{gray}\scriptsize ${\pm}$0.5}\\
		\Random[$0.05$] $p{=}1$ & 20.83 & 21.08 {\color{gray}\scriptsize ${\pm}$0.04} & 21.20 {\color{gray}\scriptsize ${\pm}$0.06} & 21.73 {\color{gray}\scriptsize ${\pm}$0.13} & 22.16 {\color{gray}\scriptsize ${\pm}$0.13} & 24.33 {\color{gray}\scriptsize ${\pm}$0.24} & 26.49 {\color{gray}\scriptsize ${\pm}$0.38}\\
		\hline
	\end{tabular}
\end{table*}
\begin{table*}
	\centering
	\small
	\caption{\textbf{Overall Robustness Results on \MNIST.} Tabular results corresponding to \figref{fig:summary} and \ref{fig:supp-summary} for $m = 8, 4, 2$ bits. We show \RTE for \Normal, \Clipping and \Random with various $\wmax$ and $p$ across a subset of test bit error rates.}
	\label{tab:supp-summary-mnist}
	\vspace*{-0.2cm}
	\begin{tabular}{| c | l | c | c | c | c | c | c | c | c |}
		\hline
		\multicolumn{10}{|c|}{\bfseries MNIST: summary results for $\mathbf{m = 8, 4, 3}$ bit}\\
		\hline
		& Model & \multirow{2}{*}{\begin{tabular}{c}\TE\\in \%\end{tabular}} & \multicolumn{7}{c|}{\RTE in \%, $p$ in \% p=0.01}\\
		\cline{3-10}
		&&& $1$ & $5$ & $10$ & $12.5$ & $15$ & $17.5$ & $20$\\
		\hline
		\hline
		\multirow{11}{*}{$m = 8$ bit} & \Normal & 0.39 & 0.77 & 86.37 & 89.92 & 89.82 & 89.81 & 90.09 & 90.03\\
		& \Quant & 0.40 & 0.69 & 85.96 & 90.20 & 89.86 & 90.10 & 89.72 & 89.83\\
		& \Clipping[$0.1$] & 0.39 & 0.48 & 18.21 & 88.93 & 90.35 & 90.06 & 90.56 & 90.18\\
		& \Clipping[$0.05$] & 0.42 & 0.47 & 0.63 & 8.67 & 51.38 & 80.64 & 87.79 & 89.57\\
		& \Clipping[$0.025$] & 0.43 & 0.47 & 0.56 & 0.71 & 0.95 & 1.81 & 7.22 & 32.68\\
		\cline{2-10}
		& \Random[$0.1$] $p{=}1$ & 0.36 & 0.44 & 3.41 & 86.29 & 89.05 & 89.85 & 90.10 & 89.93\\
		& \Random[$0.05$] $p{=}1$ & 0.34 & 0.39 & 0.59 & 8.92 & 51.32 & 79.35 & 87.63 & 89.15\\
		& \Random[$0.05$] $p{=}5$ & 0.34 & 0.38 & 0.50 & 1.02 & 5.12 & 41.31 & 79.19 & 87.88\\
		& \Random[$0.05$] $p{=}10$ & 0.40 & 0.43 & 0.51 & 0.67 & 0.86 & 1.74 & 9.77 & 47.58\\
		& \Random[$0.05$] $p{=}15$ & 0.39 & 0.40 & 0.45 & 0.56 & 0.64 & 0.78 & 1.10 & 2.72\\
		& \Random[$0.05$] $p{=}20$ & 0.39 & 0.42 & 0.48 & 0.53 & 0.57 & 0.63 & 0.74 & 0.94\\
		\hline
		\hline
		\multirow{14}{*}{$m = 4$ bit} & \Quant & 0.36 & 0.72 & 87.21 & 90.23 & 90.01 & 89.88 & 89.97 & 89.67\\
		& \Clipping[$0.1$] & 0.38 & 0.51 & 38.75 & 88.33 & 89.47 & 89.57 & 90.10 & 89.67\\
		& \Clipping[$0.05$] & 0.31 & 0.39 & 0.78 & 44.15 & 78.64 & 87.32 & 89.03 & 89.71\\
		& \Clipping[$0.025$] & 0.37 & 0.41 & 0.50 & 0.67 & 0.99 & 4.63 & 29.46 & 67.21\\
		\cline{2-10}
		& \Random[$0.1$] $p{=}1$ & 0.38 & 0.48 & 13.29 & 87.43 & 89.70 & 89.63 & 89.41 & 90.02\\
		& \Random[$0.1$] $p{=}5$ & 0.38 & 0.48 & 0.78 & 24.73 & 74.88 & 87.04 & 88.72 & 89.55\\
		& \Random[$0.1$] $p{=}10$ & 0.40 & 0.47 & 0.64 & 1.22 & 2.62 & 16.72 & 64.33 & 83.80\\
		& \Random[$0.1$] $p{=}15$ & 0.56 & 0.59 & 0.73 & 1.03 & 1.28 & 1.87 & 3.71 & 14.39\\
		& \Random[$0.1$] $p{=}20$ & 0.56 & 9.48 & 14.29 & 7.39 & 6.07 & 5.80 & 6.10 & 8.12\\
		& \Random[$0.05$] $p{=}1$ & 0.37 & 0.43 & 0.67 & 36.99 & 77.12 & 85.97 & 88.62 & 89.94\\
		& \Random[$0.05$] $p{=}5$ & 0.38 & 0.42 & 0.53 & 1.38 & 12.90 & 60.73 & 83.69 & 88.75\\
		& \Random[$0.05$] $p{=}10$ & 0.34 & 0.39 & 0.47 & 0.65 & 0.91 & 2.11 & 19.15 & 71.25\\
		& \Random[$0.05$] $p{=}15$ & 0.37 & 0.39 & 0.43 & 0.52 & 0.63 & 0.79 & 1.17 & 3.16\\
		& \Random[$0.05$] $p{=}20$ & 0.44 & 0.48 & 0.53 & 0.60 & 0.65 & 0.72 & 0.86 & 1.04\\
		\hline
		\hline
		\multirow{6}{*}{$m = 2$ bit} & \Clipping[$0.1$] & 0.47 & 3.82 & 89.19 & 89.92 & 90.22 & 90.14 & \multicolumn{2}{c}{\hphantom{c}}\\
		& \Clipping[$0.05$] & 0.41 & 0.62 & 77.19 & 89.47 & 90.40 & 90.06 & \multicolumn{2}{c}{\hphantom{c}}\\
		\cline{2-8}
		& \Random[$0.05$] $p{=}3$ & 0.47 & 0.53 & 1.36 & 82.71 & 88.66 & 90.28 & \multicolumn{2}{c}{\hphantom{c}}\\
		& \Random[$0.05$] $p{=}5$ & 0.40 & 0.49 & 0.77 & 25.72 & 78.71 & 88.22 & \multicolumn{2}{c}{\hphantom{c}}\\
		& \Random[$0.05$] $p{=}10$ & 0.40 & 0.45 & 0.58 & 0.94 & 1.82 & 15.70 & \multicolumn{2}{c}{\hphantom{c}}\\
		& \Random[$0.05$] $p{=}15$ & 0.46 & 0.51 & 0.60 & 0.77 & 0.91 & 1.21 & \multicolumn{2}{c}{\hphantom{c}}\\
		\cline{1-8}
	\end{tabular}
\end{table*}
\end{appendix}

\end{document}